\documentclass[10pt]{article}\synctex=1 
\usepackage{amsfonts}
\usepackage{amssymb}
\usepackage{amsmath}
\usepackage{amsthm}
\usepackage{geometry}
\usepackage{commath}
\usepackage{booktabs}
\usepackage{lscape}
\usepackage{xr}
\usepackage[authoryear]{natbib}
\renewcommand{\theequation}{\thesection.\arabic{equation}}
\newcommand{\myref}[2]{\hyperref[#1]{#2}}
\numberwithin{equation}{section}

\usepackage{latexsym}
\usepackage{graphicx} 
\usepackage{enumerate}
\usepackage{setspace}
\usepackage[toc,title,titletoc,header]{appendix}
\usepackage[bottom]{footmisc}   
\usepackage[pdfborder={0 0 0}]{hyperref}

\usepackage{lineno}
\setpagewiselinenumbers

\newtheorem{lemma}{Lemma}[section]
\newtheorem{corollary}{Corollary}

\theoremstyle{definition}

\theoremstyle{remark}

\newcounter{assumptionA}
\def\theassumptionA{\arabic{assumptionA}}
\newenvironment{assumptionA}[1][]{\refstepcounter{assumptionA}\medskip\noindent{\textbf{Assumption \theassumptionA. #1}}}{\medskip}

\newcounter{theorema}

\def\thetheorem{\arabic{theorema}}
\newenvironment{theorema}[1][]{\refstepcounter{theorema}\medskip\noindent{\textbf{Theorem \thetheorem. #1}}}{\medskip}

\newcounter{definitiona}

\def\thedefinition{\arabic{definitiona}}
\newenvironment{definitiona}[1][]{\refstepcounter{definitiona}\medskip\noindent{\textbf{Definition \thedefinition. #1}}}{\medskip}

\renewcommand{\thelemma}{\arabic{lemma}}

\hypersetup{
  colorlinks=true,linkcolor=blue, citecolor=blue, filecolor=blue, 
  linkbordercolor={0 1 1}, citebordercolor={1 0 0},
  pdftitle={Inference in partially identified models with many moment inequalities using Lasso},
  pdfauthor={Bugni, Caner, Kock, and Lahiri},
  pdfcreator={\LaTeX\ with package \flqq hyperref\frqq},
  pdfsubject={},
}

\begin{document}
\sloppy

\title{
Generalized Linear Models With Structured  Sparsity Estimators
}

\author{\textsc{Mehmet Caner\thanks{%
North Carolina State University, Nelson Hall,  Department of Economics, NC 27695. Email:mcaner@ncsu.edu.}} 
}

\date{\today}

\maketitle

\begin{abstract}

In this paper, we introduce structured sparsity estimators in Generalized Linear Models. Structured sparsity estimators in the least squares loss are introduced by Stucky and van de Geer (2018) recently for  fixed design and normal errors. We extend their results to debiased structured sparsity estimators with Generalized Linear Model based loss. Structured sparsity estimation means penalized loss functions with a possible sparsity structure used in the chosen norm. These include weighted group lasso, lasso and norms generated from convex cones. The significant difficulty is that it is not clear how to prove two  oracle inequalities. The first one is for the initial penalized Generalized Linear Model  estimator.  Since it is not clear how a particular feasible-weighted  nodewise regression may fit in an oracle inequality for  penalized Generalized Linear Model, we need a second oracle inequality to get oracle bounds for the approximate inverse for the sample estimate of second-order partial derivative of  Generalized Linear Model.

Our contributions are fivefold: 1. We generalize the existing oracle inequality results in penalized Generalized Linear Models by proving the underlying conditions rather than assuming them. One of the key issues is the proof of a sample one-point margin condition  and its use in an oracle inequality. 2. Our results cover even non sub-Gaussian errors and regressors. 3. We provide a feasible weighted nodewise regression proof which generalizes the results in the literature from a simple $l_1$ norm usage to norms generated from convex cones. 4. We realize that norms used in feasible nodewise regression proofs should be weaker or equal to the norms in penalized Generalized Linear Model loss. 5. We can debias the first step estimator via getting an approximate inverse of the singular-sample second order partial derivative of Generalized Linear Model loss. With this debiasing, we can  get uniformly consistent estimators and asymptotically honest confidence intervals for parameters of interest. Our simulations also show good power and excellent size of the tests based on structured sparsity estimation.

\end{abstract}



\newpage

\section{Introduction}

Generalized Linear Models (GLM) have been utilized in empirical work heavily both in econometrics and statistics. Recently, attention has been shifting to models when the number of parameters, $p$, exceeds the sample size, $n$. In a seminal paper, van de Geer et al. (2014) propose a debiased GLM with $l_1$ penalty. They were able to provide confidence intervals for parameters under high-level conditions. There were two significant issues that they solved in the literature with their article. First, they propose a formula for debiased GLM, and provide normal limits for the estimators of coefficients in  the model. Then they also solved how to estimate for the inverse of the second-order partial derivative of GLM loss. This estimate is used in  the formula for debiased GLM, and the standard plug-in estimators face the ill-posed inverse problem; hence they are not usable. They provide a non-standard solution based on a weighted version of nodewise regression, which was very difficult since the standard nodewise regression was not feasible. Recently Jankova and van de  Geer (2016) generalize the debiased GLM  with differentiable loss functions to possible non-differentiable GLM loss.  Ning and Liu (2017) consider decorrelated M estimators with convex penalty, in a similar vein and use a different technique to debias than previous two papers cited above.  They decorrelate one variable's effect on the other, and they use a Dantzig-based estimation of a specific moment function to get inference for target coefficients. Specifically, their inference centers on a low dimensional parameter, where the nuisance parameters are high dimensional. In their general theorem, there is the assumption of a consistency of moments with a specified rate. It is not also clear how this high dimensional moment estimation can have good power-size properties on inference.  

Shi et al. (2019) introduce general inference in lasso type penalties in GLM. They analyze constrained partial regularization to get likelihood ratio type tests. They do not cover 
debiased GLM. Some simulation problems in coverage probabilities of certain parameters for debiased lasso for GLM is analyzed in Xia et al. (2020). They provide a solution when $p<n$ case. 

One of the penalties that we  analyze, as a sub-case of structured sparsity-based penalties, is the group lasso by Yuan and Lin (2006). Also, a  weighted group lasso penalty for logistical regression is proposed by Meier et al. (2008). This last penalty is weighted by group size and combines $l_1$ and $l_2$ penalties.  Lounici et al. (2011) provide oracle inequalities, for the least-squares loss an $l_1$ error bound for group lasso. The $l_1$ error bound increases with the true number of groups. Recently, Mitra and Zhang  (2016) consider a debiased group lasso penalty in the least-squares loss. They use bounded regressors with subgaussian errors. They provide inference for structural parameters.  

In this paper, we contribute to the literature that is cited above in several ways. One essential contribution is  that GLM loss with structured sparsity estimators is amenable to inference in parameters of interest. We use a debiased GLM with penalties coming from structured sparsity-based norms. Our paper extends the least-squares loss 
with structured sparsity estimators as shown in Stucky and van de Geer (2018). Stucky and van de Geer (2018) use non-random covariates and normal errors, which is essential to their proof technique.

GLM case is not easy since fixed design with normal errors in the least-squares loss makes debiasing and inference  much easier to construct and handle. Note that  Stucky and van de Geer (2018)  proof in the case of least squares loss with structured sparsity estimators do not carry over to GLM loss with structured sparsity estimators with random covariates, and non-normal also non-sub Gaussian errors. So inference on debiased GLM coefficients with structured sparsity-based norms is not trivial to handle. To overcome the difficulties, we start with extending the existing oracle inequality results for GLM loss in chapters 7 and 12  of van de Geer (2016). 
Theorems 7.2 and 12.2 in  van de Geer (2016)  exist either under strong conditions that have to be verified or a sub-case of GLM loss in a simplified design. We realize that sample versions of these strong conditions hold with probability approaching one with our proofs. However, these conditions are not easy to verify. The  key to our proof is our introduction of a sample version of one-point margin condition (i.e. this is a condition that governs the loss function behavior in a neighborhood of true value of the parameters). We see that  one-point margin condition introduces additional terms in an oracle inequality proof, so we change the existing oracle inequality proofs to consider these difficulties. Next, to get an approximate inverse of the sample second order partial derivative of GLM loss we introduce a feasible weighted nodewise regression with a convex cone based norm.  In that sense, we extend the  results on $l_1$ norm of van de Geer et al. (2014) to structured sparsity based norms. To get sharper bounds on our intermediate results, we also realize that the nodewise regression norm has to be weaker than or equal to norm of the  penalized GLM loss.  This sequencing of norms is a new finding for debiased estimators in high dimensions and can be helpful in other contexts.

As an output of our approach, we can test many restrictions and also have uniform-honest confidence intervals for our parameters.
We also extend the previous literature to regressors with bounded moments and non-sub Gaussian errors. As a sub-case, we  also consider a debiased weighted group lasso estimator in GLM loss.

There have been papers in debiasing lasso type estimators and providing confidence intervals in the recent literature.  Starting with Belloni et al. (2014, 2016, 2017), Chernozhukov et al. (2018), Caner and Kock (2018, 2019), van de Geer et al. (2014) provide various ways debiasing in treatment effects, generalized linear models, least squares and GMM based models.  For panel data related debiasing, we see papers by Kock (2016), Kock and Tang (2019). In terms of quantile regression, we see contributions by  Chiang and Sasaki (2019).

Section 2 presents penalized general linear models with structured sparsity-based norm penalties. Section 3 provides a formula for how to debiasing in this new framework. Sections 4-5 offer a new oracle inequality for structured sparsity-based norm penalty and a feasible weighted nodewise regression technique. Section 6 provides a limit for increasing number of coefficients, and in Section 7, there is a sub-case of debiased weighted group lasso in GLM. Section 8 shows simulations that analyze test size, power, coverage in a limited exercise. 

\section{ Penalized Generalized Linear Models}\label{sec1}

In this section, we introduce penalized Generalized Linear Models (GLM). Our penalty  will extend $l_1$ penalty or elastic net penalty in GLM estimation. Our extension involves more general structured norms that will be tied to the sparsity properties of the parameter vector.  These types of estimators are analyzed in van de Geer (2016) formally in the least-squares case and also in more detail for least squares case in Stucky and van de Geer (2018). The assumptions in these studies for the least-squares case assume fixed design, and normal errors. The case for GLM with structured sparsity-inducing norms has not been studied. The techniques in the least-squares case  are not helpful in our case since we want to have random design with non-normal errors. GLM with structured sparsity-inducing norms in high dimensions will form the baseline estimator in our case, and we extend that to a debiased version where we can test restrictions and form confidence intervals.  We follow van de Geer et al. (2014),  where they study GLM with $l_1$ norm.  Consider the regressors $X_i \in {\cal X} \in R^p$, and the outcome
$y_i \in {\cal Y} \subseteq R$, for $i=1,\cdots, n$. The data is iid across $i=1,\cdots, n$. Regressor matrix $X$ is $n \times p$.  The loss function is:
\begin{equation}
 \rho_{\beta} (y_i, X_i):= \rho (y_i, X_i' \beta), \quad \beta \in {\cal B} \subset R^p,\label{rho}
 \end{equation}
which is convex in $\beta$.   The parameter space ${\cal B}$ is a convex subset of $R^p$.
So the loss function can be represented either as in the left of (\ref{rho}) or with the expression on the right of (\ref{rho}).
Define the first and second-order partial derivatives
\[ \dot{\rho}_{\beta}:= \frac{\partial}{\partial \beta} \rho_{\beta} (y_i, X_i), \quad \ddot{\rho}_{\beta}:= \frac{\partial}{\partial \beta \partial \beta'} \rho_{\beta} (y_i, X_i),\]
or in an alternative format
\begin{equation}
 \dot{\rho}_{\beta}:= X_i \dot{\rho} (y_i, X_i' \beta), \quad \ddot{\rho}_{\beta}:= X_i X_i' \ddot{\rho} (y_i, X_i'\beta),\label{pda}
\end{equation}
where $\dot{\rho}(.,.), \ddot{\rho}(.,.)$ are the partial derivative of function $\rho(.,.)$ with respect to second element.

 \noindent Let 
 \begin{equation}
 R (\beta):= E \rho_{\beta} (y_i, X_i' \beta),\label{pop1}
 \end{equation}
  and 
 \begin{equation}
  \beta_0 := argmin_{\beta \in {\cal B}} R (\beta).\label{pop2}
  \end{equation}
 
  So $\beta_0$ is defined as the optimizer of the expected loss function.
 Define ${\cal B}_{local}$  as  a convex subset of ${\cal B}$, which is in a local neighborhood of $\beta_0$. The local neighborhood will be defined in terms of the norm that we will use. This local set is needed  since one of the main proofs in the appendix depends on the estimator to be in this local neighborhood (i.e. one point margin condition, Lemma \ref{la2}).   

\noindent Let $\Omega(.)$ be a norm on $R^p$. We specify its properties immediately below, but first define the $\Omega$ structured sparsity GLM estimator as
\[ \hat{\beta}:= argmin_{\beta \in {\cal B} } \left[  \frac{1}{n} \sum_{i=1}^n \rho_{\beta} (y_i, X_i) +  \lambda \Omega (\beta)
\right],\]
with $\lambda > 0$ as a tuning parameter.
The norms that we analyze should have  weak decomposability property. Weak decomposability  will be a key requirement on $\Omega(.)$ and explained immediately below in Definition \ref{def1}.  We use definition 6.1 of van de Geer (2016), and this is also defining an allowed set $S$. To that effect, divide the set $J=\{1,2,\cdots, p\}$ into $S$ and its mutually exclusive complement $S^c$. In other words $J = S \cup S^c$. Let $|S|$ represent the cardinality of the index set $S$.
Now define another  norm $\Omega^{S^c}(.)$ on $R^{p - |S|}$. Also define $\beta_S$ as a vector with entries equal to zero for elements with indices when  $j \notin S$. Also define $\beta_{S^c}$ as the vector with  all elements with indices, $j$ inside the set $S$, set to zero, all elements with indices belonging to $S^c$ are kept, $S^c:=\{ j \in \{1,2,\cdots, p\}: j \notin S\}$.

\begin{definitiona}\label{def1}
(Definition 6.1, van de Geer (2016). Fix some set $S$.  We say that norm $\Omega$ is weakly decomposable for the set S if there exists a norm $\Omega^{S^c}$ on $R^{p- |S|}$ such that for all 
$\beta \in R^p$
\[ \Omega (\beta) \ge \Omega (\beta_S) + \Omega^{S^c} (\beta_{S^c}).\]
\end{definitiona}

\begin{definitiona}\label{def2}
 (Definition 6.1, van de Geer (2016)). We say that $S$ is an allowed set if $\Omega$ is weakly decomposable for the set S.
\end{definitiona}

 To give an example: for $l_1$ norm any subset $S$ of $J$ is an allowed set, and $\Omega^{S^c}(.)$ is again the $l_1$ norm. So $\|\beta\|_1= \|  \beta_{S} \|_1 + 
 \| \beta_{S^c} \|_1$. We will also give examples of weakly decomposable norms in this section. To give a broad example, all norms generated from convex cones are  weakly decomposable; see section 6.9 of van de Geer (2016).  Some of the specific examples of norms generated from convex cones are weighted group lasso norm, lasso, wedge norm, and concavity inducing norms. We give two examples of such norms. 
 
 Example 1. The first one is a weighted group lasso norm. The variables are grouped disjointly, and the penalty is designed accordingly. Let $\{G_j\}_{j=1}^m$ be a partition of $\{1, \cdots, p\}$ into disjoint $m$ groups.
For a parameter vector $\beta \in R^p$, the weighted group lasso norm is:

\[ \| \beta \|_{wgl}:= \sum_{j=1}^m \sqrt{|G_j|} \| \beta_{G_j} \|_2.\]
  So $\beta$ vector is grouped into m disjoint groups, and size of the group $G_j$ is: $| G_j|$.  Any union of groups can be an allowed set in weighted group norm.

Example 2. Another example is the wedge norm in section 6.9 of van de Geer (2016).  Consider the convex cone, ${\cal A}:= \{ a_1 \ge a_2 \ge a_3 ..\cdots a_p > 0\}$,
with $\| \beta \|_W:= \min_{a_j \in {\cal A}} \frac{1}{2} \sum_{j=1}^p \left( \frac{\beta_j^2}{a_j} + a_j 
\right).$ An allowed set is the first $s$ elements in $\beta$.

 For any norm, $\Omega$, not necessarily weakly decomposable we know that by triangle inequality
 \[ \Omega (\beta) \le \Omega (\beta_S) + \Omega (\beta^{S^c}),\]
 so clearly for weakly decomposable $\Omega(.)$, we have $\Omega (\beta^{S^c}) \ge \Omega^{S^c} (\beta_{S^c})$. By Chapter 6 of van de Geer (2016)
 dual norm of $\Omega(.)$ is defined as 
 \[ \Omega_* (w):= \max_{ \Omega (\beta) \le 1} |w'\beta|, \quad w \in R^p.\]
 
 We need few more concepts regarding norms. This is taken from Section 6.4 of van de Geer (2016). 
 
 \begin{definitiona}\label{sn} (Stronger norm). If $\Omega(.)$ and $\underline{\Omega}(.)$ are any two norms on $R^p$, and if we have 
 \[ \Omega (\beta) \ge \underline{\Omega} (\beta), \quad \forall \beta \in R^p,\]
 we say that $\Omega$ is a stronger norm than $\underline{\Omega}$.
\end{definitiona}
We also see that 
\begin{equation}
\Omega (\beta) \ge \underline{\Omega} (\beta) \quad {\mbox {implies}}  \quad \underline{\Omega}_* (\beta) \le \Omega_* (\beta),\label{dn}
\end{equation}
where $\underline{\Omega}_*$ is the dual norm of $\underline{\Omega}$. Stronger norm definition is applicable to all norms regardless of their weak decomposability or not.
As in section 6.4 of van de Geer (2016) we define the following lower bound norm for $\Omega(.)$. Formally define $\Omega^{S^c}(.)$, which is mentioned in Definition 1,  as the largest norm among the norms
$\underline{\Omega}^{S^c}(.)$ for which 
\[ \Omega(\beta) \ge \Omega(\beta_S) + \underline{\Omega}^{S^c} (\beta_{S^c}),\]
hence define
\begin{equation}
 \underline{\Omega}(\beta):= \Omega (\beta_S) + \Omega^{S^c} (\beta_{S^c}) \le \Omega (\beta).\label{unorm}
 \end{equation}

We define $S_0$ as the indices of the active set. This is defined with respect to a particular norm $\Omega(.)$. $S_0$ should be an allowed set and carry all the indices with nonzero elements in the model. To clarify the last statement, to give an example, these elements 
can be indices of individual non-zero true coefficients  in lasso via $l_1$ norm, $S_{0}:= \{ j: |\beta_{j0}| \neq 0 \}$, where $\beta_{j0}$ represents true value of $j$ th coefficient
where $j=1,\cdots, p$, where $p$ is the total number of coefficients. For the weighted group lasso norm, these indices with nonzero elements  are the indices of the active (non-zero) groups, so $S_0:=\{ j: \| \beta_{0,G_{j}} \|_2 \neq 0 \}$, where $\beta_{0,G_{j}}$ represents the true coefficients of $j$ th  group, where $j=1,\cdots, m$, where $m$ is the total number of groups in the model. We define the sparsity  as $s_0$, which is the cardinality of $S_0$,  $s_0:= |S_0|$. Let  $l_0$ ball ${\cal B}_{l_0} (s_0) := \{ \| \beta_0 \|_{l_0} \le s_0 \}$. Define the effective sparsity condition, or sometimes called $\Omega$ effective sparsity as follows.

\begin{definitiona}\label{es}
 Effective sparsity. (Definition 4.3 of van de Geer (2014)).  Suppose $S$ is an allowed set. Let $L>0$ be some constant. The effective sparsity 
is 
\[ \Gamma^2 (L,S) := [min \{ E  \| X \beta_S - X \beta_{S^c} \|_n^2 : \Omega (\beta_S) = 1, \Omega^{S^c} (\beta_{S^c}) \le L\}.]^{-1}.\]
This is the inverse of the more familiar $\Omega$-eigenvalue condition.
\end{definitiona}
This effective sparsity is defined as a population condition, compared to the sample version of van de Geer (2014), but Definition 7.5 of van de Geer (2016) has a general population version. The sample version of effective sparsity  is defined in Appendix, and also a variant of this population effective sparsity  is  given in Appendix.

\section{Debiased GLM Structured Sparsity Estimator}\label{sec2}

In this section we introduce a debiased version of GLM structured sparsity estimator. But first, we define by using differentiability of the objective function with (\ref{rho}), 
\[ \hat{\Sigma}_{\hat{\beta}}:= \frac{1}{n} \sum_{i=1}^n \ddot{\rho}_{\hat{\beta}}  (y_i, X_i)= \frac{1}{n} \sum_{i=1}^n X_{\hat{\beta},i} X_{\hat{\beta},i}',\]
 where $X_{\hat{\beta},i}:= X_i w_{\hat{\beta},i}$, in which 
 \begin{equation}
 w_{\hat{\beta},i}:=
\sqrt{\ddot{\rho} (y_i, X_i' \hat{\beta})}\label{weight}
\end{equation} by equation (\ref{pda}).  Also see that $w_{\beta_0,i}$ is defined in the same way as in (\ref{weight}), second order partial derivative depends on $\beta_0$ there.
This sample moment estimator plays a crucial role in our derivations. In the case of least-squares loss, this corresponds to the empirical Gram matrix. Note that in our GLM loss case when $p>n$, $\hat{\Sigma}_{\hat{\beta}}$ is singular.

Furthermore define a $p \times p$ matrix $\hat{\Theta}$ which will be defined later as output of a nodewise regression.  $\hat{\Theta}$ will be used as an approximate inverse for $\hat{\Sigma}_{\hat{\beta}}$. Section 5 considers the form and theory behind $\hat{\Theta}$.
Our debiased estimator is:
\begin{eqnarray}
\hat{b}&:=& \hat{\beta}- \hat{\Theta} \left[ \frac{1}{n} \sum_{i=1}^n \dot{\rho}_{\hat{\beta}} (y_i, X_i)\right] \nonumber \\
& := &\hat{\beta}- \hat{\Theta} \left[ \frac{1}{n} \sum_{i=1}^n X_i \dot{\rho} (y_i, X_i' \hat{\beta})\right]\label{dbe}
\end{eqnarray}
where $\dot{\rho}_{\hat{\beta}} (y_i, X_i)$ is the partial derivative of our GLM loss function with respect to $\beta$ and evaluated at $\hat{\beta}$, and we use (\ref{rho}) for the last equivalent definition.

We extend this debiased estimator to structured sparsity penalties. A slightly different formula is given in the previous literature, for the least-squares loss with structured sparsity penalty.  The previous literature uses nuclear norm regularized multi-nodewise regression in Definitions 3-4 of Stucky and van de Geer (2018). We realized that if the design is fixed and with normal errors in the least-squares context, their nuclear-norm-based debiased estimator is easy to come up with limits. That structure is not amenable in GLM, with random design.  

For testing in high dimensions, define a $p \times 1$ vector $\alpha$ such that $\| \alpha \|_2 =1$, and let ${\cal H}:=\{ j=1,\cdots, p: \alpha_j \neq 0\}$ with cardinality $|{\cal H}| = h$. $h$ will increase with sample size and we will precisely define this rate through our assumptions, and $h$ will be the number of restrictions that are tested and $h < p$.  Clearly 
\begin{equation}
\sum_{j \in {\cal H}} | \alpha_j | = O (h^{1/2}),\label{h1}
\end{equation}
since $\| \alpha \|_2 =1$, and ${\cal H}$ definition with using the norm inequality that puts an upper bound on $l_1$ norm in terms of $l_2$ norm. In the remaining sections we consider the following as the numerator of our test statistic:
\begin{equation}
n^{1/2} \alpha' (\hat{b} - \beta_0) = n^{1/2} \alpha' (\hat{\beta} - \beta_0) - n^{1/2} \alpha'  
\hat{\Theta} \left[ \frac{1}{n} \sum_{i=1}^n  	X_i \dot{\rho} (y_i, X_i' \hat{\beta})\right].\label{s2.1}
\end{equation}
The denominator of our test statistic will be 
\begin{equation}
\hat{V}_{\alpha} := \sqrt{\alpha' \hat{\Theta}  \left[ \frac{1}{n} \sum_{i=1}^n X_i X_i' \dot{\rho} (y_i, X_i' \hat{\beta})^2
\right] \hat{\Theta}' \alpha}.\label{s2.2}
\end{equation}

\section{ Assumptions} \label{sec3}

We provide the main assumptions used throughout the paper, and  needed for oracle inequality in Theorem \ref{thm1}. Another set of assumptions will be provided in their sections related to nodewise regression and limit theorem.

\begin{assumptionA}\label{as1}
The data $X_i,y_i$ are iid across $i=1,\cdots,n$.  Furthermore $\max_{1 \le j \le p} E | X_{1j}|^{r_x} \le C < \infty$, where $r_x  \ge 4$ and $C>0$ is a positive constant.
Also the effective sparsity is bounded away from infinity: $0< c \le \Gamma^2(2, S_0) \le C < \infty$ with $c>0$ which is a positive constant.
\end{assumptionA}

\begin{assumptionA}\label{as2}

(i). Define
\[M_1:= \max_{1 \le i \le n} \max_{1 \le j \le p} \max_{1 \le l \le p} | X_{ij} X_{il} - E X_{ij} X_{il}|,\]
and 
\[ M_2:= \max_{1 \le i \le n} \max_{1 \le j \le p} | X_{ij}|,\]
Then we assume
\[ \max \left( \frac{\sqrt{E M_1^2} \sqrt{lnp}}{\sqrt{n}}, \frac{\sqrt{E M_2^2} \sqrt{lnp}}{\sqrt{n}}\right) = O(1).\]

(ii). \[ s_0  \sqrt{lnp/n} \to 0.\]

\end{assumptionA}

\begin{assumptionA}\label{as3}
There exists a positive constant $C_{\rho}$, which depends on the shape of the second order partial derivative $\ddot{\rho}(.)$, and $\kappa$ all positive constants such that 
\[\ddot{\rho} (y_i, X_i' \beta)\ge 1 /C_{\rho}^2,\]
for all $ \sup_{\beta_0 \in {\cal B}_{l_0} (s_0)} | X_i' (\beta - \beta_0)| \le \kappa.$
\end{assumptionA}

\begin{assumptionA}\label{as4}
The derivatives $\dot{\rho} (y,a) = \frac{\partial \rho(y,a)}{ \partial a }$, $\ddot{\rho}(y,a) = \frac{\partial \rho(y,a)}{ \partial a^2 }$ exist for all $y,a$ and for some $\delta$ neighborhood of $X_i' \beta_0$, $\delta>0$ 

(i). \[ \sup_{\beta_0 \in {\cal B}_{l_0} (s_0)} \max_{a_0 \in \{X_i' \beta_0 \}} \sup_{y_i} | \dot{\rho} (y_i,a_0)| = O(1).\]

(ii). \[ \sup_{\beta_0 \in {\cal B}_{l_0} (s_0)} \max_{a_0 \in \{X_i' \beta_0 \}} \sup_{ | a - a_0 | \le \delta} \sup_{y_i} | \ddot{\rho}(y_i,a) | = O(1).\]

 (iii). Also $\ddot{\rho}(y,a)$ is Lipschitz
 \[ \sup_{\beta_0 \in {\cal B}_{l_0} (s_0)}
  \max_{ a_0 \in \{ x_i' \beta_0\} } \sup_{ | a - a_0| \cup | \hat{a} - a_0| \le \delta}
 \sup_{y_i}  \frac{ | \ddot{\rho} (y_i, a) - \ddot{\rho} (y_i, \hat{a}) | }{|a- \hat{a} | } \le 1.\]

\end{assumptionA}

We discuss the assumptions here. Assumption \ref{as1} is standard, and effective sparsity is used to control certain  matrix's singularity. Our proofs and remarks after Theorem \ref{thm1} will show how it is related to compatibility condition, which is more familiar in high dimensional statistics.  Assumption \ref{as2} is needed for concentration inequalities that we use, and these inequalities are from Chernozhukov et al. (2017).  Assumption \ref{as3} provides a lower bound on the second-order partial derivative of the loss function and is needed to control the one-point margin condition, which will be explained in the Appendix. This condition is also used in Chapter 12 of van de Geer (2016). It is possible to relax this condition, but this will lengthen the proofs immensely, so we avoided that. Assumption \ref{as4} puts structure on GLM loss, and this is used as Assumption C.1 in van de Geer et al. (2014). We strengthened this to uniform over ball ${\cal B}_{l_0} (s_0)$. We also think that it is possible to get rid of bounded first-order partial derivative at $\beta_0$, and bounded second-order partial derivative bound in a uniform neighborhood of $\beta_0$ by using Assumption \ref{as2} type  of moments of these derivatives.

\section{$\underline{\Omega}$ bound}\label{sec4}

One of the crucial elements in the paper is the $\underline{\Omega}$ bound for our estimator. This bound will be used in de-biasing, and the former literature takes this type of result given or allows fixed-random regressors in restrictive data setups.  Theorems 9.19 and Corollary 9.20 of Wainwright (2019) provide oracle bounds under very restrictive conditions on the data and eigenvalue type conditions. 
Another paper related to our paper   is the logistic result in $l_1$ norm in Theorem 12.2 of van de Geer (2016), which provides sample eigenvalue conditions with restrictive conditions on the data set.  Compared to these results before us, we provide a different proof based on primitive assumptions in a general norm-setting. Even though the estimator is obtained by using $\Omega$ bound, and we are interested in 
$\underline{\Omega}$ bound, which is a weaker norm than $\Omega$ ($\underline{\Omega} \le \Omega$). Also, some of the key difficulties in obtaining such a result are that an empirical process result has to  be established, sample one-point margin condition has to be proved, and since this is new, it has to be shown that sample one point margin condition does not impediment the oracle inequality proof. Details are in the Appendix.

We define a positive sequence $t_1$, which is defined in (\ref{e3.4}), and $t_1= O(\sqrt{lnp/n})$.  As in p.107 of van de Geer (2016) we take $\beta_{S_0}$ as "relevant coefficients" in $\beta_0$, and treat $\beta_{S_0^c}$ as "irrelevant smallish-like" part of $\beta_0$. This can be thought of nonzero coefficients as $\beta_{S_0}$, and local-to-zero and zero coefficients in $\beta_{S_0^c}$. Specifically, we formalize a condition in Theorem 
\ref{thm1}(ii) below for $\beta_{S_0^c}$ in terms of the weakly-decomposable norm that we use. A form of weak-sparsity will be imposed for asymptotic results. Define  $l_0$ ball ${\cal B}_{l_0} (s_0):= \{ \| \beta_0 \|_{l_0} \le s_0 \}$.

\begin{theorema}\label{thm1}

(i). Under Assumptions \ref{as1}-\ref{as4}, with sufficiently large $n$

\[ \underline{\Omega} (\hat{\beta} - \beta_0) \le (18 \lambda) C_{\rho}^2 \Gamma^2 (2,S_0) + 32 \Omega(\beta_{S_0^c}),\]
with probability at least $1- \frac{3}{p^{2c}} - \frac{1}{p^c} - \frac{7 C }{4 (lnp)^2}= 1- o(1)$. 

(ii). Also our Remark 2 below will show that,  with assuming $\sup_{\beta_0 \in {\cal B }_{l_0} (s_0)} \Omega (\beta_{S_0^c}) \to 0$, then  
\[ \sup_{\beta_0 \in {\cal B}_{l_0} (s_0)}
 \underline{\Omega} ( \hat{\beta} - \beta_0) = O_p ( s_0 \sqrt{\frac{lnp}{n}}) = o_p (1).\]

\end{theorema}

Remarks. 1. First, we want to rewrite the upper bound in terms of the population version of the compatibility constant, where the literature is familiar with. Define the compatibility constant as  in Definition 6.2 of van de Geer (2016) as 
\[ \phi^2 (L,S) := min\{ |S| E [\| X \beta_S - X \beta_{S_c} \|_n^2]: \Omega(\beta_S) =1, \Omega^{S^c} (\beta_{S^c}) \le L\},\]
where $S, S^c$ are any allowed set and its complement respectively. Next by Definition \ref{es} and the above expression 
\begin{equation}
\Gamma^2 (L,S) = \frac{|S|}{\phi^2 (L,S)}.\label{cc}
\end{equation}

Also, the empirical version of the equality in (\ref{cc}) is on p.81 of van de Geer (2016), just before section 6.6 there. 
Using (\ref{cc}) we can write the upper bound in terms of  sparsity of the coefficients explicitly. In that respect by at $S= S_0$, with $L=2$
\[ \Gamma^2 (2, S_0)  = \frac{|S_0|}{\phi^2 (2, S_0)}.\]

2. First, impose the weak-sparsity assumption, uniformly over ${\cal B}_{l_0} (s_0)$,  we impose $\Omega (\beta_{S_0^c}) \to 0$.
Next to get an asymptotic sense from our bound, $\lambda_e$ is a positive sequence defined in (\ref{pep8}), 
 we can have $\lambda = 16 \lambda_e = O (\sqrt{lnp/n})$ as shown in Lemma \ref{la4} in Appendix, and in Lemma \ref {la3} of Appendix we also have $t_1= O (\sqrt{lnp/n})$, then the upper bound
\begin{equation}
( 18 \lambda)  \Gamma^2 (2,S_0)= (18 \lambda) \frac{|S_0|}{\phi^2 (2, S_0)}
=o(1),\label{42a}
\end{equation}
when we replace $\Gamma^2 (2, S_0) \le C < \infty$, with $\phi^2 (2, S_0) \ge c > 0$  in Assumption \ref{as1} and since $\lambda |S_0| = \lambda s_0 = o(1)$ by Assumption \ref{as2}.

3. Even though the estimator optimizes over $\Omega$ norm, the bound is in weaker $\underline{\Omega}$ norm, which is needed for the debiased estimator.

These results imply 

\begin{equation}
\underline{\Omega} (\hat{\beta} - \beta_0) = O_p ( s_0 \lambda) = O_p (s_0 \lambda_e)=
 O_p ( s_0 \sqrt{lnp/n}) = o_p (1).\label{asymt1}
\end{equation}

4. One issue is the cost of the generality of the results.  An alternative technique could have used a different approach and  it may have been possible to get a better rate than in (\ref{asymt1}). 
 Our proof technique, on the other hand, is very general and uses the ranking of norms in Lemma \ref{la1}. 

\section{Nodewise Regression in Structured Sparsity Estimators}\label{sec5}

We start with definitions of several matrices used in nodewise regression with norm $\Omega(.)$. So we generalize the results in van de Geer et al. (2014) from $l_1$ norm to a more general norm structure designated by $\Omega(.)$. Next, we show that how nodewise regression be carried, and last we show that nodewise regression provides an approximate inverse of singular sample moment matrix, $\hat{\Sigma}_{\hat{\beta}}$ which is defined in section \ref{sec2} in GLM structure.

We extend the definitions in section \ref{sec2}.
Define $X_{\hat{\beta}}:= W_{\hat{\beta}}X$, where $W_{\hat{\beta}}:= diag (w_{\hat{\beta},1},\cdots , w_{\hat{\beta},n})'$ which is a $n \times n$ diagonal matrix. Note that $w_{\hat{\beta},i}:= \sqrt{\ddot{\rho} (y_i, X_i' \hat{\beta})}$ for $i=1,\cdots, n$. See  that $j$ th column of $X_{\hat{\beta}}$ is denoted as $X_{\hat{\beta},j}: n \times 1$, and 
$X_{\hat{\beta}, -j}:n \times p-1$ is defined as all columns of $X_{\hat{\beta}}$ except $j$ th one. We define $ W_{\beta_0}:= diag (w_{\beta_0,1},\cdots, w_{\beta_0,i}, \cdots, w_{\beta_0,n})$ which is $n \times n$ diagonal matrix, with  $w_{\beta_0,i}:= \sqrt{\ddot{\rho} (y_i, X_i' \beta_0)}$, for $i=1,\cdots,n$.
Define $n \times p$ matrix: $X_{\beta_0} := W_{\beta_0} X$, and $j$ th column  of that matrix as $X_{\beta_0,j}$, and all the columns except j th one as: $X_{\beta_0,-j}$,
Define $\Theta:= \Sigma_{\beta_0}^{-1}$, where $\Sigma_{\beta_0}:= E X_{\beta_0, i} X_{\beta_0,i}'$, where $X_{\beta_0,i}'$ is the $i$ th row of $n \times p$ matrix $X_{\beta_0}$. So $X_{\beta_0,i}':= X_i' w_{\beta_0,i}$. $X_{\beta_0, i}$ is the column version of the row $X_{\beta_0, i}'$.
Define  $\gamma_{\beta_0,j}$ as $\gamma_j$ that minimizes $E [ X_{\beta_0,j} - X_{\beta_0, -j} \gamma_j]^2$.

We  can write the following from p.3 of the supplement of van de Geer et al. (2014)
\begin{equation}
X_{\beta_0, j} = X_{\beta_0, -j}  \gamma_{\beta_0,j} + \eta_{\beta_0,j},\label{nw1}
\end{equation}
where 
\begin{equation}
E X_{\beta_0, -j}' \eta_{\beta_0,j} = 0.\label{nw1a}
\end{equation}

 By  the analysis in p.157 of Caner and Kock (2018) and (\ref{nw1}) we get the relation between $\Theta$ and regression coefficient $\gamma_{\beta_0,j}$, and the scalar $\tau_j^2$. Note that $\tau_{j}^2:= \frac{1}{\Theta_{j,j}}$, where $\Theta_{j,j}$ is the $j$ th main diagonal element of $\Theta$. With that analysis we get $\Theta= C_j/ \tau_{j}^2$, where $C_j$ is a $ p \times 1$ vector, with 1 in $j$ th cell, and the rest of $C_j$ is defined as $-\gamma_{\beta_0,j}$, $j=1,\cdots, p$,  so at $j=1$ for example,  $C_1:= (1, - \gamma_{\beta_0,1}')'$.
 

Hence, as shown in the  proof of Theorem 3.2 in van de Geer et al. (2014), premultiply (\ref{nw1}) by $W_{\hat{\beta}} W_{\beta_0}^{-1}$ to have 
\begin{equation}
X_{\hat{\beta},j} = X_{\hat{\beta},-j} \gamma_{\beta_0,j} + W_{\hat{\beta}} W_{\beta_0}^{-1} \eta_{\beta_0,j}.\label{nw2}
\end{equation}
We have the definition, for $j=1,\cdots,p$ 
\begin{equation}
\hat{\gamma}_{{\beta},j} =arg min_{\gamma_j \in R^{p-1}} [ \| X_{\hat{\beta},j} - X_{\hat{\beta},-j} \gamma_j \|_n^2 + 2 \lambda_j \underline{\Omega} (\gamma_j)]
,\label{nw3}
\end{equation}
where we will impose $\lambda_{nw}=\lambda_j$ for each $j=1,\cdots,p$, and $\lambda_{nw}$ is a positive sequence and its rate will be determined in the proofs. Define the nodewise regression estimates in the same form as in van de Geer et al. (2014).
Define $\hat{\Theta}_j:= \hat{C}_j/\hat{\tau}_j^2$, with $\hat{C}_j$ defined as a vector with $1$ in $j$ th cell, and all other $p-1$ cells are $-\hat{\gamma}_{\hat{\beta},j}$ vector from the nodewise regression, and $\hat{\tau}_j^2:=\frac{X_{\hat{\beta},j}' (X_{\hat{\beta},j} - X_{\hat{\beta},-j} \hat{\gamma}_{\hat{\beta},j})}{n}$.

One word of caution is that recently van de Geer (2016) and Stucky and van de Geer (2018) use nuclear norm loss with  a sum of norms over the restrictions tested instead of nodewise regression. This type of analysis works well due to the fixed design  nature of regressors and normal errors via the least-squares loss in the main structural parameter estimation. The technique did not carry out to our more general random design with non-normal errors and generalized linear model.

One of the key issues is the penalty in the nodewise regression. We propose $\underline{\Omega} (.)$ norm instead of $\Omega (.)$ norm. The main reason is that dual norms have the following inequality  $\underline{\Omega}_{*} \ge \Omega_{*}$ when $\underline{\Omega} \le \Omega$  by definitions of these norms. The proofs use dual norm inequality, and due to Theorem \ref{thm1} result, they use $\underline{\Omega}$ norm bounds. If we had operated with $\Omega$ based bounds in nodewise regression we had to convert them still to $\underline{\Omega}$ results which can be done via large upper bounds as shown in Lemma 3 of Stucky and van de Geer (2018) since this results in a larger bound, which the bounds depend on sparsity, so usage of $\Omega$ is not advised. In that sense, our proposal is new and will result in better- smaller bounds and faster convergence rates of the error to zero in certain proofs regarding central limit theorem type result. Specifically, this can be seen in Step 2 of proof of Theorem \ref{thm2}, the equation before (\ref{nwt}).
In summary, we provide a new approach to debiasing. If the main penalty in the loss function of the interest is $\Omega$ as in section 2, then the nodewise regression has to run with a weaker norm: $\underline{\Omega}: \Omega \ge \underline{\Omega}$. A similar approach is suggested by Stucky and van de Geer (2018) by using gauge functions as norms (weakest possible decomposable norm) in forming precision matrix estimate, with fixed design and the least squares loss. Their setup is different  and does not overlap with us, since we only analyze norms generated from cones in our main equation, and use their properties to our advantage in the proofs such as Lemma \ref{la1}.

Now we form an inequality that will help us in the proofs of debiased GLM with structured sparsity. Start with $\hat{\tau}_j^2$ definition
 and divide each side by $\hat{\tau}_j^2$, and also using $X_{\hat{\beta},j}, X_{\hat{\beta},-j}$ definitions in $X_{\hat{\beta}}$
 \begin{equation}
 1 = \frac{X_{\hat{\beta},j}' (X_{\hat{\beta},j} - X_{\hat{\beta},-j} \hat{\gamma}_{\hat{\beta},j})}{n \hat{\tau}_j^2}
 = \frac{(X_{\hat{\beta}} \hat{\Theta}_j)' X_{\hat{\beta},j}}{n} = \frac{X_{\hat{\beta},j}' X_{\hat{\beta}} \Theta_j}{n},\label{that1}
  \end{equation}
 where we use the definition $\hat{\Theta}_j:= \hat{C}_j/\hat{\tau}_j^2 $, with $(X_{\hat{\beta},j} - X_{\hat{\beta},-j} \hat{\gamma}_{\hat{\beta},j}) = X_{\hat{\beta}}
 \hat{C}_j$ by using $X_{\hat{\beta}}, \hat{C}_j$ definitions. Denoting $\hat{z}$ as the sub-differential, and getting the KKT conditions from (\ref{nw3}) 
 \[ \hat{z}_j  =  \frac{X_{\hat{\beta},-j}' (X_{\hat{\beta},j}- X_{\hat{\beta},-j} \hat{\gamma}_{\hat{\beta},j})}{ n \lambda_{nw}}.\]
 and $\underline{\Omega}_* (\hat{z}_j) \le 1$ for $j=1,\cdots,p$, $\underline{\Omega}_*(.)$ is the dual norm for $\underline{\Omega}(.)$. From KKT conditions
 \[  \frac{ \underline{\Omega}_* \left( X_{\hat{\beta},-j}' (X_{\hat{\beta},j}- X_{\hat{\beta},-j} \hat{\gamma}_{\hat{\beta},j})\right)}{n } \le \lambda_{nw},\]
 which implies by dividing each side by $\hat{\tau}_j^2$, and using $(X_{\hat{\beta},j}- X_{\hat{\beta},-j} \hat{\gamma}_{\hat{\beta},j}) = X_{\hat{\beta}} \hat{C}_j$,
 and by $\hat{\Theta}_j$ above
 \begin{equation}
 \frac{ \underline{\Omega}_* ( X_{\hat{\beta},-j}' X_{\hat{\beta}} \hat{\Theta}_j)}{n} \le \frac{\lambda_{nw}}{\hat{\tau}_j^2}.\label{that2}
\end{equation}
Combine (\ref{that1})(\ref{that2}), and $\hat{\Sigma}_{\hat{\beta}}:= X_{\hat{\beta}}' X_{\hat{\beta}}/n$, for $j=1,\cdots, p$
\begin{equation}
 \underline{\Omega}_* ( \hat{\Theta}_j' \hat{\Sigma}_{\hat{\beta}} - e_j' ) \le \lambda_{nw}/\hat{\tau}_j^2.\label{that3}
 \end{equation}

 So we show that in dual norm $\hat{\Sigma}_{\hat{\beta}}$ has an approximate inverse $\hat{\Theta}$.  Equation (\ref{that3}) is a new result  and can be used in other contexts.
 Next we put forward our assumptions for this nodewise regression result. Before the next assumption we define the following terms. Let the number of restrictions in $\beta_0$ vector as 
 $h$, and with indices of ${\cal H}:= \{ j: H_0: \beta_j = \beta_{0j} \}$.  We can let $h$ grow with $n$, but $h < p$. Let $X_{\beta_0,-j,i,k}$ represent 
 $X_{\beta_0,-j}$ ($n \times (p-1$)) matrix $(i,k)$ th element, and $\eta_{\beta_0, j,i}$ is the ($n \times 1$) $\eta_{\beta_0,j}$ vector's  $i$ th element
 
 \[ M_3:= \sup_{\beta_0 \in {\cal B}_{l_0} (s_0)} \max_{1 \le i \le n} \max_{j \in  {\cal H}} \max_{1 \le k \le p-1} | X_{\beta_0,-j, i,k} \eta_{\beta_0,j,i} |.\]
 
 \[ M_4:= \sup_{\beta_0 \in {\cal B}_{l_0} (s_0)} \max_{1 \le i \le n} \max_{j \in  {\cal H}} | \eta_{\beta_0, j,i}^2 - E \eta_{\beta_0, j,i}^2|.\]
 
 Since $h < p$, so $ln ph < ln p^2 = 2 ln p$. 
 
 We index $\gamma_{\beta_0,j}$ into $\gamma_{\beta_0, S_j}$, where $S_j$ represents all the indices with nonzero components in $\gamma_{\beta_0,j}$, and $\gamma_{\beta_0, S_j^c}$ where $S_j^c$ represents the indices with all the local to zero, and zero coefficients.
We also provide a compatibility condition, for $j=1,\cdots, p$, 
\begin{equation}
 \phi^2 (L, S_j):= \min \{ | S_j | E \| X_{\beta_0, -j} \gamma_{\beta_0,S_j} - X_{\beta_0, -j} \gamma_{\beta_0,S_j^c} \|_n^2:  \Omega (\gamma_{\beta_0,S_j}) =1, \Omega^{S^c} (\gamma_{\beta_0,S_j^c}) \le L \}
.\label{evl1}
\end{equation}

\begin{assumptionA}\label{as5}

 (i). 
 $inf_{\beta_0 \in {\cal B}_{l_0} (s_0)} Eigmin (\Sigma_{\beta_0}) \ge c > 0$.
 Also 
$\sup_{\beta_0 \in {\cal B}_{l_0} (s_0)}\max_{1 \le i \le n} \max_{1 \le j \le p}  E | \eta_{\beta_0,i,j}|^r \le C < \infty$, for $r > 8$.

(ii). \[ \frac{\sqrt{E M_3^2} \sqrt{lnp}}{\sqrt{n}} = O (1).\]
\[ \frac{\sqrt{E M_4^2} \sqrt{ln h}}{\sqrt{n}} = O (1).\]

\end{assumptionA}


Define $\bar{s}:= \max_{1 \le j \le p} |S_j|$. Define two positive sequences, $H_n:= O (h^{2/r} n^{2/r})$, and  $K_n := O (p^{2/r_x} n^{2/r_x})$. Define a known sequence $g_n $ which depends on the norm that is analyzed and the sample size, and $g_n$ is a nondecreasing function in $n$. To give an example, if  $l_1$ norm  is used  for $\underline{\Omega}(.)$ then $g_n=\bar{s}^{1/2}$ by (B.55) of Caner and Kock (2018).
Formally, $g_n$ is defined in Assumption \ref{as6}(iii).

\begin{assumptionA}\label{as6}

(i). \[ g_n\bar{s}^{1/2} \sqrt{\frac{lnp}{n}} (\max(\bar{s}, H_n^2 s_0^2) = o(1).\]

(ii).\[ K_n \bar{s} s_0 \sqrt{\frac{lnp}{n}} = o(1).\]


(iii). \[ \sup_{\beta_0 \in {\cal B}_{l_0} (s_0)}
\max_{1 \le j \le p} \underline{\Omega} (\gamma_{\beta_0,j}) = O ( g_n)
.\]

\end{assumptionA}

Assumptions \ref{as5}, \ref{as6} only relate to nodewise regression. Assumption \ref{as5} uses an eigenvalue condition and implies compatibility condition  (\ref{evl1}) via Lemma 4.1 of van de Geer (2014), and it is different in form and elements from the effective sparsity condition in Definition 4. This difference stems from the nature of nodewise regression, which is described above and our oracle inequality proof in Lemma 1 below. Our Assumption \ref{as5} is a strengthened version of eigenvalue assumption for nodewise regression in van de Geer et al. (2014) due to uniformity in ${\cal B}_{l_0} (s_0)$ in our case.
Cross product of moments rate assumption can be relaxed at the expense of  lengthening the proofs via marginal moment conditions.
Assumption \ref{as6} is a sparsity type assumption that also replaces Assumption \ref{as2}(ii). 

We take a specific example to show that Assumption \ref{as6}(i)-(ii) is holding, without some of the constants to simplify the issue.  Let $\bar{s}=  ln n, s_0 = ln n, p=2n, H_n = n^{2/r},
K_n =   (2 ln n) ^{2/r_x} n^{4/r_x}$. Then with $l_1$ norm,$g_n = \sqrt{\bar{s}}$, and 
 \[\bar{s} \sqrt{ln p/n} \max (\bar{s}, H_n^2 s_0^2)= O (\frac{(ln n)^{3/2}}{n^{1/2}} max(ln n, n^{4/r} (ln n)^2))
=O \left(  (ln n)^{7/2} n^{4/r - 1/2}\right) = o(1) \] with $r > 8$, and $K_n \bar{s} s_0 \sqrt{lnp/n} = O ( (ln n)^{\frac{5}{2} + \frac{2}{r_x}} n^{4/r_x - 1/2}) = o(1)$ with $r_x > 8$.  Assumption \ref{as6}(iii), and $g_n$ can be shown in other contexts than $l_1$ norm, for example in weighted group lasso norm, $g_n =  m \sqrt{| G|}$, where $m$ is the number of groups, and $|G|$ is the largest group size.

The following lemma is an essential result and shows the estimation of the rows of the precision matrix with nodewise regression, and the estimators are consistent. Define 
\begin{equation}
 d_n:= g_n \sqrt{\bar{s}} \sqrt{\frac{lnp}{n}} \max(\bar{s}, H_n^2 s_0^2).\label{eqdn}
 \end{equation}

\begin{lemma}\label{la10}
Under Assumptions \ref{as1} with $r_x > 8$,  \ref{as2}(i),\ref{as3}-\ref{as6}, with the following weak sparsity condition (\ref{wsc1}), uniformly over ${\cal B}_{l_0} (s_0)$
 \begin{equation}
 \max_{j \in {\cal H}} \Omega^{S^c} (\gamma_{\beta_0, S_j^c})=o(d_n)=o(1),\label{wsc1}
 \end{equation}
 then 
\[ \max_{j \in {\cal H}}
\underline{\Omega} (\hat{\Theta}_j - \Theta_j) = O_p \left( g_n \sqrt{\bar{s}} \sqrt{\frac{lnp}{n}} \max(\bar{s}, H_n^2 s_0^2)\right) =o_p (1).\]

This result is also valid uniformly over $l_0$ ball ${\cal B}_{l_0} (s_0)$.

\end{lemma}

Remarks. 1. This is a new lemma in the literature and establishes  general norm bounds on nodewise regression estimates for GLM based estimators. In this sense, this provides a general result for estimating the inverse of the second-order partial derivative of GLM objective function. The usage of nodewise regression is necessitated by the singularity of the sample second-order partial derivative of GLM objective function. The closest to this result is in Theorem 3.2 of van de Geer et al. (2014) with $l_1$ norm bounds in GLM. Our proof also extends Theorem 6.1 of van de Geer (2016) proof for linear loss, with high-level conditions to generalized linear models with primitive conditions, and for a weaker norm, $\underline{\Omega}$.

2. The limit on van de Geer et al. (2014) depends on strong assumptions such as uniformly bounded regressors, uniformly bounded product of nodewise regression coefficient with regressor, and the knowledge of oracle inequalities in GLM in prediction norm as well as $l_1$ norm. Our result generalizes their results to regressors with moment bounds, and also there is no need for the product of regressors and the nodewise coefficient  to be uniformly bounded. Also we obtain oracle inequalities in our Theorem \ref{thm1}.

3. The cost to a more general proof will be slightly different rates compared to $l_1$ norm. We have a different proof technique than van de Geer et al. (2014), and benefiting from a maximal  inequality that is due to  Chernozhukov et al. (2017). In case of $l_1$ norm Theorem 3.2 in van de Geer et al. (2014) under the strong assumptions provide a rate of 
$\max(K \sqrt{\bar{s} lnp/n}, K^4 s_0 \sqrt{lnp/n})$, where they need a $\lambda = O ( K \sqrt{lnp/n})$, where $K$ is the rate for the uniformly bounded regressors in their case, i.e. $\max_{i,j}
| X_{i,j}| = O( K)$, which is their Assumption D.1.  In our Lemma \ref{la10} above, our rate is $\max( \bar{s}^2 \sqrt{lnp/n}, \bar{s} s_0^2 H_n^2 \sqrt{lnp/n})$, since in $l_1$ case $g(\bar{s}) = \bar{s}^{1/2}$ as can be shown via analysis in p.159 of  Caner and Kock (2018).  In $l_1$ case it is not clear which proof technique will provide a sharper rate, since assumptions are different, and our proof is geared toward a general norm result, the bounds/proofs are different, hence not resulting in the same rate for $l_1$ in both cases.


4. Also, an interesting point is that whether a different weaker norm can also be useful in this lemma. In other words if we have $\bar{\Omega}$ such that $l_1 \le \bar{\Omega}
\le \underline{\Omega}$. Proof of this lemma clarifies that such a $\bar{\Omega}$ proof will go through. Essentially, a very good choice can be $l_1$ norm
which provides a sharper bound,  unless there is a specially structured sparsity for nodewise regression. This norm choice also can be seen  by Assumption \ref{as6}(iii).

\section{Limit}

In this section we provide a limit result. But before that, for variance-covariance estimation we need the following Assumption which is a stricter version of Assumption \ref{as2}(i). Define 
\[ M_5:= \max_{1 \le i \le n} \max_{1 \le k \le p } \max_{1 \le l \le p} | X_{ik}^2 X_{il}^2 - E X_{ik}^2 X_{il}^2|.\]

\begin{assumptionA}\label{as7}

(i). \[ \frac{\sqrt{E M_5^2} \sqrt{lnp}}{\sqrt{n}} = O(1).\]

(ii). Set $r_x>8$ in Assumption \ref{as1}, and let $a \wedge b = min (a,b)$

 \[ \frac{(h\bar{s})^{(r_x/4)+1} \wedge (h \bar{s})^{r_x/4} p }{n^{(r_x/4)-1}} = o(1).\]

(iii). \[  h g_n \bar{s} \frac{lnp}{\sqrt{n}} \max(\bar{s}, H_n^2 s_0^2) = o(1).\]

(iv). \[ (h \bar{s})^{1/2} K_n s_0^2 \frac{lnp}{n^{1/2}} = o(1).\]

\end{assumptionA}

\begin{assumptionA}\label{as9}

\[\inf_{\beta_0 \in {\cal B}_{l_0} (s_0)}
 Eigmin (E X_i X_i' \dot{\rho} (y_i, X_i' \beta_0)^2 ) \ge c > 0,\]

\[ \sup_{\beta_0 \in {\cal B}_{l_0} (s_0)} Eigmax ( \Sigma_{\beta_0}) \le C < \infty,\]
where $c, C$ are positive constants.

\end{assumptionA}

Assumptions \ref{as7}, \ref{as9} are needed for central limit theorem result. Specifically, we use $r_x>8$.  Assumption \ref{as9} is a  standard assumption on population moments by taking into account GLM nature of our problem. Assumption \ref{as9} can be weakened easily by use of uniformly bounded weights and moment conditions on regressors. To see that Assumption \ref{as7} is feasible, we set up the following example. Let $h= ln (n), \bar{s}=ln(n), s_0 = ln (n), p=2n, r_x=9$, then Assumption \ref{as7}(ii) holds since $\frac{(ln n)^{13/2}}{n^{5 /4}} \to 0$.  For Assumption \ref{as7}(iii), with $l_1$ norm for 
$g_n= \sqrt{\bar{s}}$, with $\bar{s} = ln (n), s_0 = ln (n)$ and $r=9$, $h = ln (n)$, $p =2n$, we have $H_n = O ([ln (n)]^{2/9} n^{2/9})$, so 
$\max( \bar{s}, H_n^2 s_0^2) = O \left( [ln (n)]^{4/9} n^{4/9} [ln (n)]^2\right)$, then 
\[ h \bar{s}^{3/2} \frac{lnp}{\sqrt{n}} H_n^2 s_0^2 = O \left( \frac{[ln (n)]^{(7/2)+ (22/9)} n^{4/9}}{n^{1/2}}\right) = o(1).\]
To show Assumption \ref{as7}(iv), with the setup in (iii), $r_x = 9$, $K_n = O ( n^{4/9})$, then $[ln (n)]^4 n^{4/9}/n^{1/2} \to 0$ provides (iv).

We provide our main result, which is  a central limit theorem for debiased GLM structured sparsity estimators. As far as we know, this is a new result in the literature where we have general weakly decomposable norms.  We want to test the null of $\beta_j= \beta_{j0}$ for $j \in {\cal H}$.

\begin{theorema} \label{thm2}
Under Assumptions \ref{as1}-\ref{as2}(i), \ref{as3}-\ref{as9},  with $\sup_{\beta_0 \in {\cal B}_{l_0}  (s_0)} \max_{j \in {\cal H}} \Omega^{S_c} (\gamma_{\beta_0, S_j^c})=o(d_n)=o(1)$

(i).  Uniformly over $l_0$ ball ${\cal B}_{l_0} (s_0)$
\[\frac{n^{1/2}  \alpha' (\hat{b} - \beta_{0})}{\hat{V}_{\alpha}} \stackrel{d}{\to} N(0,1),\]
where $\hat{V}_{\alpha}^2:= \alpha' \hat{\Theta} [\frac{1}{n} \sum_{i=1}^n X_i X_i' \dot{\rho} (y_i, X_i' \hat{\beta})^2] \hat{\Theta}' \alpha.$

(ii). \[ \sup_{\beta_0 \in {\cal B}_{l_0} (s_0)}  | \hat{V}_{\alpha}^2 - V_{\alpha}^2| = o_p (1),\]
with $V_{\alpha}^2:= \alpha' \Theta [ E X_i X_i' \dot{\rho} (y_i, X_i' \beta_0)^2] \Theta \alpha$.

\end{theorema}

Remarks. 1. This theorem extends Theorem 3.3  in van de Geer et al.et al. (2014) from $l_1$ norm to $\Omega (.)$ norm under weaker conditions.  The main issues are: 
a). We need to show a new  oracle inequality for $\underline{\Omega}$ based norm as in Theorem \ref{thm1}.  
b). Then we need a feasible nodewise regression via the  proof in Lemma \ref{la10}. The reason that Lemma \ref{la10} is needed rather than simple usage of proof of  Theorem \ref{thm1} is that weights and feasible nodewise regression introduce a technical issue so that we need to extend least squares loss proof in Theorem 6.1 of van de Geer (2016) to GLM in our Lemma \ref{la10}.

2.  Nonlinear restrictions may be another topic, but we think  it will take more space for this paper, so we did not cover it.


3. A good choice to get better approximation rates can be the usage of $l_1$ norm in nodewise regression, regardless of the penalty norm for the initial estimator $\hat{\beta}$ in section 2, this point can be seen in the proofs by (\ref{pt2.10})
(\ref{pt2.12}) and the step 2 of proof of Theorem \ref{thm2} by Assumption \ref{as6}(iii) with Lemma A1(ii), in terms of vector norms: $l_1 \le \underline{\Omega} \le \Omega$.

We provide a theorem  that provides uniform confidence intervals for our parameters. The proof uses Theorem \ref{thm2} and follows the proof of Theorem 3 in Caner and Kock (2018). So no proof will be given. Let $\Phi (t)$ be the cdf of a standard normal distribution, and $z_{1 - \delta/2}$ is the $1 - \delta/2$ percentile of the standard normal distribution, and let $diam ([a,b])=b-a$ be the length of the interval $[a,b]$ in the real line, for all $j=1,\cdots, p$ we have the following Theorem.

\begin{theorema}\label{thm3}
Under Assumption \ref{as1} with $r_x>8$, Assumptions \ref{as2}(i), \ref{as3}-\ref{as9},  with $\sup_{\beta_0 \in {\cal B}_{l_0} (s_0)}
\max_{j \in {\cal H}} \Omega (\gamma_{\beta_0, j})=o(d_n)=o(1)$

(i). \[ \sup_{t \in R} \sup_{\beta_0 \in {\cal B}_{l_0} (s_0)} \left| P \left(  \frac{n^{1/2} \alpha' (\hat{b}- \beta_{0})}{\sqrt{ \alpha' \hat{\Theta} [\frac{1}{n} \sum_{i=1}^n X_i X_i' \dot{\rho} (y_i, X_i' \hat{\beta})^2] \hat{\Theta}' \alpha}} \le t \right) - \Phi (t)
\right| \to 0.\]

(ii).  For each $j=1,\cdots, p$
\[ \lim_{ n \to \infty} \inf_{\beta_0 \in {\cal B}_{l_0} (s_0)} P \left( 
\beta_{j0} \in [ \hat{b}_j - z_{1 - \delta/2} \frac{\hat{\sigma}_j}{n^{1/2}}, \hat{b}_j + z_{1 - \delta/2} \frac{\hat{\sigma}_j}{n^{1/2}}]
\right) = 1 - \delta.\]

(iii). \[ \sup_{\beta_0 \in {\cal B}_{l_0} (s_0) }diam \left( [ \hat{b}_j - z_{1 - \delta/2} \frac{\hat{\sigma}_j}{n^{1/2}}, \hat{b}_j + z_{1 - \delta/2} \frac{\hat{\sigma}_j}{n^{1/2}}]\right)
= O_p (\frac{1}{n^{1/2}}).\]

\end{theorema}

\section{Example: Logistic Loss with Debiased Weighted Group Lasso}

In this part of the paper, we follow our theorems with an example. This sub-case of our main theorems will be an analysis of the logistical loss function with a weighted group lasso norm. Then this estimator will be debiased, and we form confidence intervals for the coefficients that are of interest. Yuan and Lin (2006) introduces group lasso  to capture the relation between the outcome variable and group of variables, rather than the individual variables. Oracle inequalities are proved by Lounici et al. (2011), and the debiased weighted group lasso estimator is analyzed by Mitra and Zhang (2016). These three papers handled the linear loss function with group structure.  Meier et al. (2008) analyze weighted group lasso for logistic regression and provide maximal inequalities. So we extend this literature by providing a debiased weighted group lasso in a logistical loss context. First, we start with the penalty function and then show the more familiar logistical loss. Our penalty is the weighted group lasso norm in p.89 of van de Geer (2016).  We follow the description and the properties of this norm from p.89-90 in van de Geer (2016).
Let $\beta_{G_j}$ represent the $\beta$ vector that correspond to group $G_j$ entries, and there are $m$ groups in total. We assume groups are disjoint $G_j \cap G_k = \emptyset$. So $ \cup_{j=1}^m G_j = \{1, \cdots, p\}$. Group size of group $G_j$ is the cardinality of the group $|G_j|$. Let us denote the maximum group size by $g:= \max_{1 \le j \le m} |G_j |$.  The weighted group lasso norm is defined as
\begin{equation}
\Omega (\beta):= \sum_{j=1}^m \sqrt{ | G_j |} \| \beta_{G_j} \|_2,\label{wgln}
\end{equation}
where  each group is weighted by the square root of its cardinality, in this way large groups are penalized proportionately to their size. Penalization occurs for each group, hence a group with its all members are included or excluded from the regression. The dual norm for weighted group lasso norm is, for a general vector $\omega$ 
\begin{equation}
\Omega_* (\omega):= \max_{1 \le j \le m} \frac{ \| \omega_{G_j} \|_2}{\sqrt{| G_j|}},\label{dnorm}
\end{equation}
where $\omega_{G_j}$ represent all elements correspond to $G_j$. Also any group is an allowed set, as well as their unions, since the weighted group lasso is weakly decomposable.  Furthermore this norm is also decomposable. This means
\[ \sum_{j=1}^m \sqrt{|G_j |} \| \beta_{G_j }\|_2 = \sum_{j \in S} \sqrt{|G_j|} \| \beta_{G_j} \|_2 + \sum_{j \in S^c} \sqrt{|G_j|} \|\beta_{G_j} \|_2,\]
and $S$ can be a subset of all groups, say  $S:= \{ 2, 4, 5\}$, and the remainder is $S_c:= \{ 1, 3, 6\}$, if $m=6$. So groups \{2,4,5\} and \{1,3,6\} can be decomposed 
into two separate sets $S$, $S^c$. Denote the active set with $S_0$, where this carries the indices  of the active (relevant-nonzero groups), whereas $s_0$ is the cardinality, which is the number of relevant-active groups.  We setup a logistic loss function with group structure as in Meier et al. (2008). Let $y_i$  be iid across i and take values of 0 or 1 
, and $X_i$ is also iid across i and is a $p \times 1 $ vector. Denote $X_{i, G_j}$ as the predictors in $G_j$ th group at $i$ th observation, and $\beta_{G_j}$ represent the parameter vector corresponding to th $G_j$ th group, here $p_j:=| G_j|$ are the number of elements in $j$ th group, i.e. number of parameters in $\beta_{G_j}$.\[ \rho_{\beta} (y_i, X_i):= \rho (y_i, X_i' \beta) = - y_i \sum_{j=1}^m X_{i,G_j}' \beta_{G_j}  + ln (1 + exp (\sum_{j=1}^m X_{i,G_j}' \beta_{G_j} )),\]
where $X_{i,G_j}: p_j \times 1$, and $\sum_{j=1}^m p_j = p$. Let $\beta:=( \beta_{G_1}, \cdots, \beta_{G_j} , \cdots, \beta_{G_m})'$ which is a $p \times 1$ vector 
with $\beta_{G_j}: p_j \times 1$, for $j=1,\cdots, m$. The logistic loss with weighted group norm estimator is:

 \begin{equation}
  \hat{\beta}_{LL}:= argmin_{\beta \in R^p} \left[ \frac{1}{n} \sum_{i=1}^n \{ -y_i \sum_{j=1}^m X_{i,G_j}'  \beta_{G_j} + ln (1 + exp (\sum_{j=1}^m X_{i,G_j}' \beta_{G_j}))\}
 + 2 \lambda \sum_{j=1}^m \sqrt{ |G_j|} \| \beta_{G_j} \|_2 \right].\label{wgl}
 \end{equation}

 We want to carefully analyze whether Assumptions \ref{as1}-\ref{as4} are verified. First we see that Assumptions \ref{as1}-\ref{as2} are still needed, and $s_0$ is the number of relevant-active groups, which is the cardinality of $S_0:\{ j: \| \beta_{G_{j0}}\|_2 \neq 0 \}$. Assumption \ref{as3} is also needed with $C_{\rho} \ge 1$ since 
 \[ \ddot{\rho} (y_i, X_i' \beta ) = \frac{exp (X_i' \beta)}{[1 + exp (X_i'\beta)]^2},\]
 where $X_i' \beta = \sum_{j=1}^m X_{i, G_j}' \beta_{G_j}$. Assumption \ref{as4} holds in this case of logistic loss. We start with Assumption \ref{as4}(i).  See that 
 \[ \dot{\rho} (y_i, X_i' \beta_0) = - y_i + \frac{exp(X_i' \beta_0)}{1 + exp (X_i'\beta_0)},\]
 and clearly since $y_i$ is binary, with zero or one value, $| \dot{\rho} (y_i, X_i' \beta_0)| \le 2$ by triangle inequality. So (i) is satisfied. Next, for Assumption \ref{as4}(ii), we have, $a:= X_i' \beta= \sum_{j=1}^m X_{i, G_{j}}' \beta_{G_j}$ 
 \begin{equation}
  \ddot{\rho} (y_i, a) = \frac{exp (a)}{(1+ exp (a))^2},\label{ew1}
  \end{equation}
 hence $|\ddot{\rho} (y_i, a)| \le 1$, so (ii) is verified. Then Assumption \ref{as4}(iii) is clearly satisfied since we have third degree differentiable in $a= X_i' \beta$ for $\rho(.)$ in logistical loss. Third order partial derivative is bounded, and with mean value theorem we get Lipschitz continuity in $a$.

 \begin{corollary}\label{c1}
 Under Assumptions \ref{as1}-\ref{as3} 
 \[ \| \hat{\beta}_{LL} - \beta_0 \|_{wgl}:= \sum_{j=1}^m \sqrt{|G_j|} \| \hat{\beta}_{G_j} - \beta_{0, G_j} \|_2 = 
 O_p ( s_0 \frac{\sqrt{lnp}}{\sqrt{n}}),\]
 where $s_0$ is the number of active-relevant (nonzero in $l_2$ norm) groups, which is the cardinality of $S_0=\{ j: \| \beta_{0,G_j} \|_2 \neq 0 \}$. The result is uniform over $l_0$ ball ${\cal B}_{l_0} (s_0) = \{ | \{ j: \| \beta_{0, G_j}\|_2 \neq 0 | \le s_0 \}$, where $|.|$ represents the cardinality of an index set.
 \end{corollary}
 
 Remark. Note that we have $\sqrt{lnp}$, where $p$ is the dimension of regressors. This rate may be the cost that we incur with our proof. 
 So our general proof technique may have  a cost in the rate, albeit a mild one.  We now compare our results with the ones we can find in the literature.  Note that we cite the two examples   using the least-squares loss unlike our GLM loss. Our example and two comparison examples use the group lasso norm. 
 For group norm, under uniformly bounded empirical Gram matrix, with non-normal errors,  Theorem 8.1 of Lounici et al. (2011) has
 \[ s_0 \frac{ \sqrt{ (ln m)^{3/2 + \delta}}}{\sqrt{n}},\]
 where $\delta >0$ is a positive constant. This result is (8.3) of Lounici et al. (2011) with one task (T=1) there which provides the group structure equivalent to us. So main difference between the rates is comparing $lnp$  with $ (ln m)^{3/2 + \delta}$, so with large number of groups $m$, our and their result will be similar or we do better, otherwise when $m$ is small, the estimation  error may be smaller than our result. Mitra and Zhang (2016) on the other hand find the rate
 \[ \| \hat{\beta} - \beta_0 \|_{wgl} = O_p ( \frac{l + s_0 ln m}{n}),\]
 where $l$ is the cardinality of the largest group. This last result is derived under uniformly bounded regressor assumption. So if $l$ is close to $n$ our rate seems better, otherwise, their rate is very good.

 All the other assumptions are not tied to penalties.  Now we define the debiased logistic estimator with a weighted group lasso penalty. Formally
 \begin{equation}
 \hat{b}_{LL}:= \hat{\beta}_{LL} - \hat{\Theta} \left[ 
 \frac{1}{n} \sum_{i=1}^n X_i \left( -y_i + \frac{exp(X_i' \hat{\beta}_{LL})}{1 + exp (X_i' \hat{\beta}_{LL})}
 \right)
 \right] .\label{llwgl}
  \end{equation} 
 As mentioned above to get $\hat{\Theta}$ we follow (\ref{nw3}) and the paragraph below that, we can  use $l_1$ norm for nodewise regression. The term, which is summed in the last parenthesis, is the partial derivative of logistic loss with respect to $\beta$, and this point is made in (\ref{dbe}). Set sparsity in the precision matrix  for all $j \in S_j^c: \gamma_{\beta_0, S_j^c} = 0$ to simplify the expressions in the corollary below, although weak sparsity is allowed as shown in Theorems above. We provide the limit 
 for a debiased logistic estimator with weighted group norm.As far as we know, this is a new result in the literature.

 \begin{corollary}\label{cor2}
 Under Assumptions \ref{as1}-\ref{as2}(i), \ref{as3}, \ref{as5}--\ref{as9},  uniformly over $l_0$ ball ${\cal B}_{l_0} (s_0)$, 
\[\frac{n^{1/2}  \alpha' (\hat{b}_{LL} - \beta_{0})}{\hat{V}_{\alpha}} \stackrel{d}{\to} N(0,1),\]
where $\hat{V}_{\alpha}^2:= \alpha' \hat{\Theta} \left[\frac{1}{n} \sum_{i=1}^n X_i X_i' \left( -y_i+ \frac{exp (X_i' \hat{\beta}_{LL})}{(1 + exp (X_i' \hat{\beta}_{LL}))}
\right)^2 \right] \hat{\Theta}' \alpha.$
 \end{corollary} 
 
 \section{Monte Carlo}

 In this section, we consider the performance of the debiased weighted group lasso with logistical loss that is described in the previous section. 
 We consider two main setups. They will differ in terms of   number of groups. Setup 1 will have 5 groups, and Setup 2 will have 10 groups. In each setup, we want to see the size and power of the test, and coverage of zero and nonzero coefficients.  Since the computations are time-consuming, we use 100 iterations for each exercise.

 Setup1: There are 5 groups and one intercept, which is not included in the groups (the intercept is not penalized). 
 Let $p+1$ represent the total number of parameters fitted. Also, just to give an example, for $l$ th group, the row vector can be represented as: $\beta_{0,g_l}' = 0_{p/5}'$, which is all zeros, with dimension of the $l$ th group as $p/5$. Given that our parameter set is:
 \[ \beta_0:=(\beta_{0,1}=0, \beta_{0,g_1}' = 0_2', \beta_{0,g_2}'= 0_{p/5+10}', \beta_{0,g_3}'=1_{p/5}', \beta_{0, g_4}'= 0_{p/5}',
 \beta_{0,g_5}'=0_{2*p/5-12}')'.\]
 
 For each $i_l=1,\cdots, n_l$, where $n_l$ is the observations in $l$ th group. Across $i_l$, the data is iid, with $X_{i_l}$ (which is $p$ is multivariate normal and $X_{i_l} \sim N(0, \Sigma)$ with $k,j$ th element in $\Sigma$)
\[ \Sigma_{k,j} = \rho^{|k-j|},\]
with $\rho=0.5, 0.75$. Inside the groups, the regressors are correlated, but outside there is independence. Also  each group has the same multivariate normal distribution, but as described independent from other groups. 

In setup 2, we deviate from setup 1. Here we just cover the differences between two setups. In this design, we have 10 groups to measure the effect of number of groups in our analysis.
\begin{eqnarray*}
 \beta_0:&=&(\beta_{0,1}=0, \beta_{0,g_1}' = 0_2', \beta_{0,g_2}'= 0_{p/10+10}', \beta_{0,g_3}'=1_{p/10}', \beta_{0, g_4}'= 0_{p/10}',
 \beta_{0,g_5}'=0_{2*p/10-12}',\\
 & &\beta_{0,g_6}'=0_{p/10}', \beta_{0,g_7} = 0_{p/10}',\beta_{0,g_8}'= 2_{p/10}', \beta_{0,g_9}'=0.5_{p/10}', \beta_{0,g_{10}}'=0_{p/10}'
 )'.
 \end{eqnarray*}

Tuning parameter choice is essential, and for the weighted group lasso estimation, we use the procedure outlined by Meier et al. (2008). First, we set up a grid of $\lambda$ choices, and let $\lambda_{max}$ represent the $\lambda$ when used in weighted group lasso (logistical loss), will provide all zero parameter estimates. The "grplasso-R" program by Meier (2020)
computes both $\lambda_{max}$ and the weighted group lasso in the logistical loss. Our grid of $\lambda$ choices are 
\[ \Lambda:= \{ \lambda_{max}*0.3, \lambda_{max}*(0.3)^2, \cdots, \lambda_{max}*(0.3)^{25} \}.\]
These are  $25$ possibilities in total. This type of grid is very similar to the one used in p.66 of Meier et al. (2008), and the idea is taken from that paper. For the weighted group lasso estimator with logistical loss, the  tuning parameter choice is given by p.66 of Meier et al. (2008). This is similar to a two-fold cross-validation exercise. We give a broad outline of the procedure to choose $\lambda$ in Section 7 above, and form our estimator in Monte Carlo.

1. From the first half of the data $(i=1,\cdots, n/2))$ pick coefficient estimates by applying the weighted group lasso in (\ref{wgl}) for each $\lambda \in \Lambda$ above.

2. Use these estimates in the second half of the sample $i=n/2+1,\cdots, n$ in the unpenalized logistical loss, (i.e. the  term (\ref{wgl}) without the weighted group lasso penalty).

3. Pick the $\lambda$ that provides the minimum in step 2 above. Denote this $\lambda$ as $\lambda_o$.

4. Run (\ref{wgl}) with full sample with $\lambda_o$ and get coefficient estimates $\hat{\beta}_{	LL}$.

5. We describe now the nodewise regression to get $\hat{\Theta}$. For this purpose we form weighted regressors 
$X_{\hat{\beta}_{LL}, j}:= W_{\hat{\beta}_{LL}} X_j$, and $X_{\hat{\beta}_{LL},-j}:= W_{\hat{\beta}_{LL}} X_{-j}$ with 
$W_{\hat{\beta}_{LL}}:= diag (\hat{w}_{\hat{\beta}_{LL},1}, \cdots, \hat{w}_{\hat{\beta}_{LL},i}, \cdots, \hat{w}_{\hat{\beta}_{LL},n})$ with 
$\hat{w}_{{\beta}_{LL},i}:=\sqrt{\ddot{\rho} (y_i, X_i' \hat{\beta}_{LL})}$ where we use (\ref{ew1}).

6. Then we run (\ref{nw3}) with $l_1$ penalty, and to choose tuning parameters, we use five-fold cross-validation. 

7. We can then form $\hat{\Theta}$ as described in (\ref{nw3}) and below that equation.

8. 
 Use  $\hat{\beta}_{LL}, \hat{\Theta}$ in the formula for the debiased weighted group lasso estimator  in (\ref{llwgl}) and get $\hat{b}_{LL}$.

9. After getting $\hat{b}_{LL}$ the test formation and coverage can be seen in Sections 6-7.

We consider four different targets. First, we report the size of the test at $5\%$ with $h=2$ restrictions, and we test $H_0: \beta_2 =0, \beta_3 =0$, $\beta_{1}$ is the intercept, so we test basically whether group 1 is significant or not. For the power exercise, we test $H_0: \beta_2 = 0.5, \beta_3 =0.5$, which is a mild deviation from the true parameters. We also check the coverage of nonzero and nonzero parameter by checking $\beta_{0, g3,3} =1$ which is the third group's third coefficient for the nonzero parameter, and $\beta_{0,2}=0$ for the zero parameter, which is the first coefficient of group 2.

 All else is the same for setup 2, except for the coverage of nonzero parameter exercise, which we check $\beta_{0, j}=1$, $j=p/10+16$. Tables 1-2 report the results. All cells in tables report 
 percentages.
 
 In both setups, we cover five combinations of sample size with number of parameters: $(n=100, p=100,) (n=150,p=100), (n=150,p=200), (n=300, p=200), (n=300,p=400)$. This type of setup is chosen since it can analyze three issues: 1. at fixed $n$, what will be the role of increase in $p$ on our metrics?, 2. at fixed $p$ what will be the role of increase in $n$ on our metrics? 3. when we increase $p,n$ simultaneously what will be the effect on our metrics?

  \begin{table}[h]
  \centering
Table 1: Design 1: Five Groups
\begin{tabular}{|c||cccc|cccc|}
\toprule 
 & \multicolumn{4}{|c|}{$\rho=0.5$}& \multicolumn{4}{|c|}{$\rho=0.75$}\\ \hline
\hline &Size & Power& Cov. Zero & Cov. Nonzero & Size & Power & Cov.Zero & Cov. Nonzero 
 \\ \hline
$n=100, p=100$ &3 & 76 & 98  & 90& 0 & 30 & 100 & 100\\ 
$n=150, p=100$ &  5 & 97 &92& 91& 0 & 39 & 100 & 100\\ 
 $n=150,p=200$ & 4 & 99 &97& 62& 0 & 43& 100 & 100\\ 
  $n=300,p=200$& 6 & 99 &98&89& 0 & 60 & 100 & 100 \\ 
 $n=300,p=400$ & 3&100 &100& 45& 0 & 70 & 100 & 100 
 \\  \hline   
\end{tabular}
Note that all numbers are in percentages. Also Cov.Zero, Cov. Nonzero reflects Coverage (average across iterations) for a  zero parameter, and nonzero parameter respectively. 
\end{table}

  \begin{table}[h]
  \centering
Table 2: Design 1: Ten  Groups
\begin{tabular}{|c||cccc|cccc|}
\toprule 
 & \multicolumn{4}{|c|}{$\rho=0.5$}& \multicolumn{4}{|c|}{$\rho=0.75$}\\ \hline
\hline &Size & Power& Cov. Zero & Cov. Nonzero & Size & Power & Cov.Zero & Cov. Nonzero 
 \\ \hline
$n=100, p=100$ &9 & 86 & 95  & 81& 4 & 35 & 100 & 100\\ 
$n=150, p=100$ &  5 & 85 &99& 88& 0 & 38 & 100 & 100\\ 
 $n=150,p=200$ & 5 & 91 &97& 67& 2 & 45& 100 & 100\\ 
  $n=300,p=200$& 6 & 100 &100&76& 0 & 59 & 100 & 100 \\ 
 $n=300,p=400$ & 2&100 &99& 41& 1 & 70 & 100 & 100 
 \\  \hline   
\end{tabular}
Note that all numbers are in percentages. Also Cov.Zero, Cov. Nonzero reflects Coverage (average across iterations) for a  zero parameter, and nonzero parameter respectively. 
\end{table}
 
 Tables show that our test has a very good size across two different group sizes with different correlation structures for regressors. To give an example with five groups in Table 1, with $p=400, n=300$, the size of the tests are 3\% and 0\% at 5\% levels at $\rho=0.5, \rho=0.75$ respectively. We see more varying power results. The test has good power with $\rho=0.50$ structure in both Tables 1-2. The power is between 76-100\%. However, with $\rho=0.75$, a larger correlation among regressors, we see that the power declines. At $p=400, n=300$, in Table 2, with 10 groups and $\rho=0.75$, the power is 70\%; however with $\rho=0.5$ in Table 2, the power is at 100\%. At 95\% ideal coverage level, we see that in Table 1, for zero parameter, the coverage is very good at $p=200, n=150$, and they are at 97\%, and 100\% level with $\rho=0.5, \rho=0.75$ respectively. For the nonzero parameter, the coverage is good at $p=100$, but deteriorates at $p=400$.
 
 To answer the  questions about increasing sample size-parameter dimension, Tables 1-2 show that when we keep $n=300$ and increase the number of parameters  $p$ from 200 to 400, the size improves at $\rho=0.5$ from 6\% to 2-3\%, the power is stable at 99-100\%, coverage of zero parameter is stable at 98-100\%, however, the coverage of nonzero parameter deteriorates from 76-89\% to 41-45\%. The same type of results is more stable at $\rho=0.75$. Then we also see, if we fix $p=200$ and increase $n$ from 150 to 300 in Tables 1-2, at $\rho=0.50$, the size is stable at 4-6\%, the power improves from 91-99\% to 99-100\%, the coverage of zero parameter is stable at 97-100\%, and the coverage of the nonzero parameter improves from 62-67\% to 76-89\%. In the case of $\rho=0.75$, with the same question of increasing $n$ with fixed $p$, the power improves, and the other metrics are stable. The last question is, what may happen when we jointly increase $n,p$ from $n=150,p=200$ to $n=300, p=400$? The answer is very similar to the first question, for example, at 	Tables 1-2, with $\rho=0.50$, we see size decline from 4-5\% to 2-3\%, and the power improves from 91-99\% to 99-100\%, the coverage of zero parameter increases from 97\% to 99-100\%, however, the coverage of nonzero parameter declines from 62-67\% to 41-45\%.

 \section{Conclusion}
 
 In this paper, we propose structured penalty functions in generalized linear models. Using a feasible-weighted nodewise regression with the same or weaker penalty norm than the original problem, we estimate the inverse of the second order partial derivative of the loss  function. Using this approximate inverse of the second-order partial-derivative we get a debiased GLM -structured sparsity estimator. We build uniformly valid confidence intervals around the parameters using the debiased estimate. A sub-case of debiased logistical loss with weighted group lasso penalty is analyzed. For future work, M-estimation in sparse structured framework can be considered.

\setcounter{equation}{0}\setcounter{lemma}{0}
\renewcommand{\theequation}{A.%
\arabic{equation}}\renewcommand{\thelemma}{A.\arabic{lemma}}%
\renewcommand{\baselinestretch}{1}\baselineskip=15pt

\section*{Appendix}



The Appendix has two parts. Part A covers oracle inequality  for GLM structured sparsity estimator, and then we have a limit theorem for the debiased version of GLM structured sparsity estimator.
Oracle inequality that we provide extends Theorem 7.2 of van de Geer (2016) to the case of random-non sub gaussian dat. We provide proofs for noise reduction, sample effective sparsity-eigenvalue, and a new one-point margin condition that takes into account the sample  that is used. The limit theorem extends $l_1$ norm result, Theorem 3.1-3.3 of van de Geer et al. (2014) to weakly decomposable norms and under weaker assumptions by proving norm bounds rather than assuming them. 

 Part B covers nodewise regression with structured sparsity for the first time in the literature. We extend the least-squares result  Theorem  6.1 of  van de Geer (2016) to GLM and extend $l_1$ norm result of van de Geer et al. (2014)  to weakly decomposable norms.\\

{\bf PART A}:\\

Here we start with some of the results that will be repeated in the proofs.  Let $\underline{\Omega}$ be a norm and its dual is represented as $\underline{\Omega}_*$.

\begin{lemma}\label{la1}
Dual Norm Inequality-Generalized Cauchy-Schwartz Inequality. (p.2179, Stucky and  van de Geer (2018)). For any vectors $x,y$, we have 

(i).\begin{equation}
 | x'y| \le \underline{\Omega}_*(x)  \underline{\Omega} (y).\label{sa01}
\end{equation}

(ii). If we have  $\underline{\Omega}$ norm generated from cones which implies weak decomposability, then  as in Lemma 6.9 of van de Geer (2016) or 
paragraph after Lemma 6.5 of van de Geer (2016)
\begin{equation}
 l_1(x)  \le \underline{\Omega} (x) .\label{sa02}
  \end{equation}

(iii).  Using (ii), we can obtain the following from p.79, or Lemma 6.2 of van de Geer (2016).
  \begin{equation}
  \underline{\Omega}_* (x) \le l_{\infty} (x).\label{sa03}
    \end{equation}

(iv). For a matrix $A$, and vector $x$ we have 
\begin{equation}
\| A x \|_{\infty} \le \| A \|_{\infty} \| x\|_1.\label{sa04}
\end{equation}
\end{lemma}

So (\ref{sa01}) applies to all norms, but (\ref{sa02})(\ref{sa03}) applies only to norms that are generated from cones.

\subsection*{Maximal Inequalities}

In this section, we show some extra results. These will help us in proofs but
they are already used in other papers, for completeness we provide them here.
Define iid random variables across $i=1,2\cdots,n $, $F_{i} = ( F_{i1},
\cdots, F_{ij}, \cdots, F_{id})^{\prime }$. Also define $\sigma_{F}^{2} = n
max_{1 \le j \le d} var F_{ij}$, and $M_{F} = \max_{1 \le i \le n} \max_{1 \le
j \le d} |F_{ij} - E F_{ij} |$. Note that $\hat{\mu}_{j} = n^{-1} \sum
_{i=1}^{n} F_{ij}$, and $\mu_{j} = E F_{ij}$.

Now we have two assumptions that will provide us maximal inequalities. The random variables $F_i$ are iid random vectors, and our Assumptions A.1-A.2 will be holding through our main Assumptions \ref{as1}-\ref{as2}  in the main text.

{\it Assumption A.1. Assume $F_{i}$ are iid random vectors across $i=1,2,\cdots, n$
with $\max_{1 \le j \le d} var F_{ij}$ bounded away from infinity uniformly in $n$.\newline}

{\it 
Assumption A.2. Assume
\[
\frac{\sqrt{E M_{F}^{2}} \sqrt{ln d}}{\sqrt{n}} = O(1).
\]}

Note that Assumptions A.1-A.2 can be extended to independent data with slightly more restrictive conditions with a cost of  tedious notation.

Maximal inequality we benefit derived from the following one. With Assumption
A.1, Lemma E.2(ii) of Chernozhukov et al.. (2017) is: (set $\eta=1, s=2$ in their Lemma)
\begin{equation}
P \left[ \max_{1 \le j \le d} | \hat{\mu}_{j} - \mu_{j} | \ge2 E \max_{1 \le j \le
d} |\hat{\mu}_{j} - \mu_{j} | + \frac{t}{n} \right] \le exp(-t^{2}/3 \sigma_{F}^{2})
+ K_1 \frac{ E M_{F}^2}{t^2},\label{sa1}%
\end{equation}
for a constant $K_1 >0$. With Assumptions A.1-A.2 here,
 Lemma E.1 of  Chernozhukov et al.. (2017) provides, with $K_2>0$ a positive constant,
\begin{align}
E \max_{1 \le j \le d} | \hat{\mu}_{j} - \mu_{j} |  & \le K_2 [ \frac{\sqrt
{lnd}}{\sqrt{n}} + \frac{ \sqrt{E M_{F}^{2}} ln d}{n}]\nonumber\\
&  =  O ( \frac{\sqrt{ln d}}{\sqrt{n}}).\label{sa2}%
\end{align}

Define the function $\kappa_{n} = ln d$. Set $t = t_{n} = ( n\kappa_{n})^{1/2}$ to have (\ref{sa1}) as
\begin{eqnarray}
P \left[ \max_{1 \le j \le d} | \hat{\mu}_{j} - \mu_{j} | \ge 2 E \max_{1 \le j \le d} | \hat{\mu}_{j} - \mu_{j} | + \frac{\sqrt{\kappa_{n}}}{\sqrt{n}} \right] &\le&
 exp(-c \kappa_{n} ) + K_1 \frac{ E M_{F}^2}{n \kappa_{n}} \nonumber \\
& \le & \frac{1}{d^c} + \frac{C}{(lnd)^2},
\label{sa3}%
\end{eqnarray}
where $C, c>0$, are   positive constants. To get the last inequality we use
\begin{eqnarray}
\frac{ E M_F^2}{n ln d} & = & \left[ \frac{(E M_F^2)^{1/2}}{n^{1/2} (ln d)^{1/2}}
\right]^2 \nonumber \\
& = & \left[ \frac{\sqrt{E M_F^2} \sqrt{lnd}}{\sqrt{n}}
\right]^2 \left( \frac{1}{lnd} \right)^2 \nonumber \\
& \le & \frac{C}{(lnd)^2},\label{sa3a}
\end{eqnarray}
where we use multiplication and division by $lnd$ to get the second equality, and Assumption A.2 to get the inequality.

Now combine (\ref{sa2}) with (\ref{sa3a}) to have

\begin{eqnarray}
P \biggl( \max_{1 \le j \le d} | \hat{\mu}_{j} - \mu_{j} |  & \ge& 2K_2 [ \frac
{\sqrt{lnd}}{\sqrt{n}} + \frac{ (E M_{F}^{2})^{1/2} ln d}{n}] + \frac
{\sqrt{lnd}}{\sqrt{n}} \biggr) \nonumber \\
&  \le &\frac{1}{d^{c}} + \frac{ C}{(lnd)^2}= o(1),\label{sa4}%
\end{eqnarray}
by Assumptions A1-A.2 and $d \to \infty$ as $n \to \infty$.
 This shows also that
\begin{equation}
\max_{1 \le j \le d } | \hat{\mu}_{j} - \mu_{j} | = O_{p} ( \sqrt{lnd}%
/\sqrt{n}).\label{sa5}%
\end{equation}

\noindent We refer to  the results (\ref{sa2})(\ref{sa4})(\ref{sa5}) in this  paper for our proofs.

\subsection*{Events}

We define three events that we use in the proof of upper bound on our norm. The probabilities of these events and  specifically the case that they are holding with wpa1 will be shown in the next three lemmata. Note that these events are assumed as known in Theorem 7.2 of van de Geer (2016).

First a sample version of one point margin condition in Chapter 7 of van de Geer (2016), with $t_1>0$ (a positive sequence that converges to zero, and defined in (\ref{e3.4})), and for all $\tilde{\beta} \in {\cal B}_{local}$, and $C_p$ is a constant that is defined in Assumption \ref{as3}, and for $M >0$
\begin{equation}
 E_1:= \left\{ R (\tilde{\beta}) - R (\beta_0) \ge \frac{ \| X (\tilde{\beta} - \beta_0) \|_n^2}{2 C_p^2} - \frac{M^2 t_1}{2 C_p^2}
\right\}.\label{e1}
\end{equation} 
Define the sample effective sparsity as Definition 4.3 in van de Geer (2014)

\[\hat{\Gamma}^2 (L, S):= [min \{ \| X \beta_S - X \beta_{S_c} \|_n^2 : \Omega (\beta_S)=1, \Omega^{S^c} (\beta_{S_c}) \le L \}]^{-1}
.\]

\noindent Then an inequality tying the population to sample effective sparsity condition, for sufficiently large $n$,
\begin{equation}
 E_2:= \{ 2 \, \Gamma^{2} (2,S_0) \ge \hat{\Gamma}^2 (2,S_0) \} .
\label{e2}
\end{equation}
Next, we have the empirical process condition, or noise reduction, with $\lambda_e>0$ a positive sequence that is defined in (\ref{pep8}) below, for $M>0$, that will be defined in (\ref{po21}) 

\begin{equation}
 E_3:= \{ \sup_{ \check{\beta} \in {\cal B}: \, \underline{\Omega} ( \check{\beta} - \beta_0) \le M }
| [ R_n (\check{\beta}) - R (\check{\beta})] - [ R_n (\beta_0) - R (\beta_0)]| \le \lambda_e M \} 
.\label{e3}
\end{equation}

\subsection*{Proofs}

The following lemma is a one-point margin condition. This condition is used in oracle inequality proofs  when the proofs involve population effective sparsity combined with population one-point margin condition. It is not clear this type of condition holds in generalized linear models with structured sparsity estimators.
We extend that to sample one-point margin condition since it will be combined with sample effective sparsity condition.  We show that it holds for our estimators 
in the following lemma. 
${\cal B}_{local}$ is a convex subset of the collection $\{ \check{\beta}: \underline{\Omega} ( \check{\beta} - \beta_0) \le M \}.$
We define $\tilde{\beta} \in {\cal B}_{local}$.

\begin{lemma}\label{la2}
Under Assumptions \ref{as1}, \ref{as2}(i), \ref{as3} for $M>0, t_1>0$,

 \[ R (\tilde{\beta}) - R (\beta_0) \ge \frac{ \| X (\tilde{\beta} - \beta_0) \|_n^2}{2 C_p^2} - \frac{M^2 t_1}{2 C_p^2},\]
with probability at least $1 - \frac{1}{p^{2c}} - \frac{C}{4 ( ln p)^2}= 1- o(1)$, with $c, C$ are positive constants, and $t_1 = O (\sqrt{lnp/n}).$

\end{lemma}

Remark. We extend the population margin condition to a  sample one and show that we need to subtract $M^2 t_1/2 C_p^2$ on the right side compared with the population one in Condition 7.6.1 of van de Geer (2016). Also, we prove that indeed this condition holds  with probability approaching one for structured sparsity estimators.

{\bf Proof of Lemma \ref{la2}}.  Define $\Sigma:= E X_i X_i'$, and $\hat{\Sigma}:= n^{-1} \sum_{i=1}^n X_i X_i'$.
First by a second degree Taylor expansion like Lemma 11.1 in van de Geer (2016) simplified through $\beta_0$ definition (\ref{pop1})(\ref{pop2}) with the partial derivative  $\dot{R} (\beta_0)=0$  and our Assumption 
\ref{as3}
\begin{equation}
R (\tilde{\beta}) - R (\beta_0) \ge \frac{ (\tilde{\beta} - \beta_0)' \Sigma (\tilde{\beta} - \beta_0)}{2 C_p^2}.\label{e3.1}
\end{equation}
Now add and subtract $\frac{(\tilde{\beta} - \beta_0)' \hat{\Sigma} (\tilde{\beta} - \beta_0)}{2 C_p^2}$ to the right side of (\ref{e3.1})
\begin{eqnarray}
R (\tilde{\beta})& -& R (\beta_0) \ge \frac{\| X (\tilde{\beta} - \beta_0) \|_n^2}{2 C_p^2} +
\frac{ (\tilde{\beta} - \beta_0)'(\Sigma-  \hat{\Sigma}) (\tilde{\beta} - \beta_0)}{2 C_p^2} \nonumber \\
& \ge & \frac{\| X (\tilde{\beta} - \beta_0) \|_n^2}{2 C_p^2} - \left|\frac{ (\tilde{\beta} - \beta_0)'(\Sigma-  \hat{\Sigma}) (\tilde{\beta} - \beta_0)}{2 C_p^2}\right|
.\label{e3.2}
\end{eqnarray}
Next, we consider the numerator of the last term on the right side of (\ref{e3.2}) 
\begin{eqnarray}
| (\tilde{\beta} - \beta_0)' (\Sigma - \hat{\Sigma}) ( \tilde{\beta} - \beta_0)| & \le & 
\underline{\Omega} (\tilde{\beta} - \beta_0) \underline{\Omega}_* ( [ \hat{\Sigma} - \Sigma] [\tilde{\beta} - \beta_0]) \nonumber \\ 
& \le & \underline{\Omega} (\tilde{\beta} - \beta_0)  \|   [ \hat{\Sigma} - \Sigma] [\tilde{\beta} - \beta_0] \|_{\infty} \nonumber \\ 
& \le &  \underline{\Omega} (\tilde{\beta} - \beta_0)  \|   \hat{\Sigma} - \Sigma \|_{\infty} \| \tilde{\beta} - \beta_0 \|_1 \nonumber \\ 
& \le & [\underline{\Omega} (\tilde{\beta} - \beta_0)]^2 \| \hat{\Sigma} - \Sigma \|_{\infty} \nonumber \\
& \le & M^2 t_1,\label{e3.3}
\end{eqnarray}
where we use Lemma \ref{la1} (i) for the first inequality, Lemma \ref{la1} (iii) for the second inequality, Lemma \ref{la1}(iv) for the third inequality, and Lemma \ref{la1}(ii) for the fourth inequality, $\tilde{\beta} \in {\cal B}_{local}$ for the fifth inequality, and  (\ref{sa2})(\ref{sa4})(\ref{sa5})  setting $d=p^2$ there, with finding 
\begin{equation}
\| \hat{\Sigma} - \Sigma \|_{\infty} \le t_1,\label{e3.3a}
\end{equation}
with 
\begin{equation}
t_1 =  2 K [ \frac{ C \sqrt{lnp^2}}{n} +  \frac{ \sqrt{E M_1^2} ln p^2}{n}] + \frac{\sqrt{ln p^2}}{\sqrt{n}}.\label{e3.4}
\end{equation}
and probability at least $ 1 - \frac{1}{p^{2c}} - \frac{C}{4 (lnp)^2}=1-o(1)$ under Assumption \ref{as1}-\ref{as2}(i), and 
\begin{equation}
t_1 = O (\sqrt{\frac{lnp}{n}}).\label{e3.4a}
\end{equation}

{\bf Q.E.D.}

Now we tie our sample and population effective sparsity conditions. Define a positive constant $L>0$. Note that the following lemma is new and this result is assumed in Theorem 7.2 of van de Geer (2016), and  in  section 11.6 of van de Geer (2016).

\begin{lemma}\label{la3}
Under Assumptions \ref{as1}, \ref{as2}

 \[ \hat{\Gamma}^{-2} (L,S) \ge \Gamma^{-2} (L,S) - (L+1)^2 t_1,\]
with probability at least $1 - \frac{1}{p^{2c}} - \frac{C}{4 (ln p)^2}= 1- o(1)$, and $t_1= O (\sqrt{lnp/n})=o(1)$.

\end{lemma}

{\bf Proof of Lemma \ref{la3}}.

The proof has two parts. The first part considers the definition of effective sparsity. The second part relates empirical effective sparsity to the population one.

Part 1.

We start with $\Omega$ effective sparsity definition in Definition 4.3 of van de Geer (2014).  To do that, let $S$ be an allowed set.  The definition is, for an allowed set $S$
\begin{equation}
\hat{\Gamma}^2 (L, S):= [min \{ \| X \beta_S - X \beta_{S_c} \|_n^2 : \Omega (\beta_S)=1, \Omega^{S^c} (\beta_{S_c}) \le L \}]^{-1}.\label{sevd}
\end{equation}

Note  that Lemma 4.1 of van de Geer (2014) shows that for an allowed set $S$
\begin{equation}
\Omega (\beta_S) \le \hat{\Gamma} (L, S) \| X \beta \|_n,\label{ev00}
\end{equation}
and 
\begin{equation}
\hat{\Gamma} (L,S) = \left[ \min \{ \frac{ \| X \beta \|_n}{\Omega (\beta_S)}: \Omega^{S_c} ( \beta_{S_c}) \le L \Omega (\beta_S)\}\right]^{-1}.\label{ev01}
\end{equation}

The population version of the same condition is:
\begin{equation}
\Gamma (L,S) = \left[ \min \{ \frac{ \beta'  \Sigma \beta }{\Omega (\beta_S)}: \Omega^{S_c} ( \beta_{S_c}) \le L \Omega (\beta_S)\}\right]^{-1}.\label{ev02}
\end{equation}

Part 2. 

Start with 
\begin{equation}
| \beta' (\hat{\Sigma} - \Sigma ) \beta | \le [ \underline{\Omega} ( \beta)]^2 \| \| \hat{\Sigma} - \Sigma \|_{\infty} ,\label{ev03}
\end{equation}
where the proof is exactly as in (\ref{e3.3}).  Next we want to bound $\underline{\Omega} ( \beta)$ on the right side of (\ref{ev03}). To that effect start with the cone condition and add $\Omega ( \beta_S)$ to both sides of the cone condition
\begin{equation}
\underline{\Omega}(\beta):=\Omega (\beta_S) + \Omega^{S_c} ( \beta_{S_c}) \le L \Omega (\beta_S) + \Omega (\beta_S) = (L+1) \Omega (\beta_S),\label{ev04}
\end{equation}
where we use the definition of $\underline{\Omega} (\beta)$. 

Then use (\ref{ev04}) with (\ref{ev03}) to have 
\begin{eqnarray}
| \beta' \hat{\Sigma} \beta |  & \ge & | \beta' \Sigma \beta | - | \beta' ( \hat{\Sigma} - \Sigma ) \beta | \nonumber \\
& \ge & | \beta' \Sigma \beta| - (L+1)^2 [\Omega (\beta_S)]^2 \| \hat{\Sigma} - \Sigma \|_{\infty} \label{ev05}
\end{eqnarray}
Divide each side by $\Omega^2 (\beta_S)>0$ 
\begin{equation}
\frac{ | \beta' \hat{\Sigma} \beta |}{[\Omega (\beta_S)]^2} \ge \frac{| \beta' \Sigma \beta |}{[\Omega (\beta_S)]^2} - (L+1)^2 \| \hat{\Sigma} - \Sigma \|_{\infty}\label{ev06}
\end{equation}
Take into account (\ref{e3.3a})(\ref{e3.4})(\ref{ev01})(\ref{ev02}), and minimize both left and right sides of (\ref{ev06})  with respect to $\beta$
\begin{equation}
\hat{\Gamma}^{-2} ( L,S) \ge \Gamma^{-2} (L,S) - (L+1)^2 t_1,
\end{equation}
with probability at least $ 1 - \frac{1}{p^{2c}} - \frac{C}{4 (lnp)^2}=1-o(1)$ with $t_1=o (1)$ under Assumption \ref{as1}-\ref{as2}.{\bf Q.E.D.}

Lemma \ref{la3} has implications for the effective sparsity.  To see that with probability at least $ 1 - \frac{1}{p^{2c}} - \frac{C}{4 (lnp)^2}$
\begin{equation}
\frac{\Gamma^2 (L,S)}{1 - (L+1)^2 t_1 \Gamma^2 (L,S)} \ge \hat{\Gamma}^2 (L,S),\label{ev07}
\end{equation}

\noindent At $S= S_0$, $L =2$ we have, with sufficiently large $n$ 
\[ 9 t_1 \Gamma^2 (2, S_0) \le 1/2,\]
since $t_1 \Gamma^2 (2, S_0) = o(1)$ and this is due to $t_1 = o(1)$, $\Gamma^2 (2, S_0)$ being finite by Assumption \ref{as1}. So 
\[ 1 - 9 t_1 \Gamma^2 (2, S_0) \le 1 - 1/2.\]
Then with probability at least $1 - \frac{1}{p^{2c}} - \frac{C}{4 (ln p)^2}$ with sufficiently large $n$
\begin{equation}
E_2 = \{ 2 \, \Gamma^2 (2, S_0) \ge \hat{\Gamma}^2 (2, S_0) \}.\label{ev09}
\end{equation}


The following lemma is stated as a condition in Theorem 7.2 of van de Geer (2016). Section 10.5 of van de Geer (2016) provides high level conditions to get that lemma. We show below that this condition can be proven, wpa1, under weaker conditions for Generalized Linear Models.

\begin{lemma} \label{la4}
Under Assumptions \ref{as1}, \ref{as2}, \ref{as4}, and for a given $\lambda_e >0$ which is defined in (\ref{pep8}) in the proof below

(i). \[ P \left[ \sup_{\check{\beta} \in {\cal B}: \, \underline{\Omega} ( \check{\beta} - \beta_0) \le M }
\left| [ R_n (\check{\beta}) - R (\check{\beta})] - [ R_n (\beta_0) - R (\beta_0)] \right| \le \lambda_e M 
\right]\ge 1 - \frac{1}{p^{2c}} - \frac{1}{p^c} - \frac{ 5 C}{4 (ln p)^2} = 1 - o(1),
\]
and $\lambda_e = O( \sqrt{lnp/n})$.

\end{lemma}

Remark. In Lemma \ref{la2}, we use $\tilde{\beta} \in {\cal B}_{local}$, and in Lemma A.4 we use $\check{\beta} \in {\cal B}$. Note that ${\cal B}_{local}$ is a convex subset of ${\cal B}$.

{\bf Proof of  Lemma \ref{la4}}.  
Using the definitions, and Assumption \ref{as1}, data being iid
\[ R_n (\check{\beta}) =\frac{1}{n} \sum_{i=1}^n \rho (y_i, X_i' \check{\beta}), \quad R (\check{\beta}) = E \rho( y_i, X_i' \check{\beta}),\]
and
\[ R_n (\beta_0) =\frac{1}{n} \sum_{i=1}^n \rho (y_i, X_i' \beta_0), \quad R (\beta_0) = E \rho( y_i, X_i' \beta_0),\]
Then by second order Taylor series expansion
\begin{eqnarray}
R_n (\check{\beta}) - R_n (\beta_0) & = & \frac{1}{n} \sum_{i=1}^n \dot{\rho} (y_i, X_i' \beta_0) X_i' (\check{\beta} - \beta_0) \nonumber  \\
& + & \frac{1}{2n} \sum_{i=1}^n \ddot{\rho} (y_i, X_i' \bar{\beta}) (\check{\beta} - \beta_0)' X_i X_i' (\check{\beta} - \beta_0),\label{pep1}
\end{eqnarray}
where $\bar{\beta} \in (\beta_0, \check{\beta})$. Also

\begin{equation}
R (\check{\beta}) - R (\beta_0) = \frac{1}{2} E [ \ddot{\rho} (y_i, X_i' \bar{\beta}) (\check{\beta} - \beta_0)' X_i X_i' (\check{\beta} - \beta_0)],\label{pep2}
\end{equation}
by iid data, and  by $\beta_0$ definition in (\ref{pop2}) we have $E X_i \dot{\rho} ( y_i, X_i' \beta_0) =0$.

Subtract (\ref{pep2}) from (\ref{pep1})
\begin{eqnarray}
[ R_n (\check{\beta} ) - R ( \check{\beta})]& - & [ R_n (\beta_0) - R (\beta_0)] \nonumber \\
& = & \frac{1}{n} \sum_{i=1}^n \dot{\rho} ( y_i, X_i' \beta_0) X_i' (\check{\beta} - \beta_0) 
\nonumber \\
& + & \frac{1}{2n} (\check{\beta} - \beta_0)' [ \frac{1}{n} \sum_{i=1}^n \ddot{\rho} (y_i, X_i' \bar{\beta}) X_i X_i' ] (\check{\beta} - \beta_0) \nonumber \\
& - & \frac{1}{2n} (\check{\beta}- \beta_0)' [ \frac{1}{n} \sum_{i=1}^n E [ \ddot{\rho} (y_i, X_i' \bar{\beta}) X_i X_i'] (\check{\beta} - \beta_0).\label{pep3}
 \end{eqnarray}

Define $\Xi_i:= \dot{\rho} ( y_i, X_i' \beta_0)$ then rewrite the first term on the right side of (\ref{pep3})
\begin{eqnarray}
\left| \frac{1}{n} \sum_{i=1}^n \Xi_i X_i' (\check{\beta} - \beta_0) 
\right| & \le & \underline{\Omega}_* ( \frac{1}{n} \sum_{i=1}^n X_i \Xi_i) \underline{\Omega} (\check{\beta} - \beta_0) \nonumber \\
& \le & \| \frac{1}{n} \sum_{i=1}^n X_i \Xi_i  \|_{\infty} \underline{\Omega} (\check{\beta} - \beta_0) \nonumber \\
& \le & \| \frac{1}{n} \sum_{i=1}^n X_i \Xi_i  \|_{\infty} M,\label{pep4}
\end{eqnarray}
by Lemma \ref{la1}(i), \ref{la1}(iii), and using the statement in the lemma. So combine second and third terms on the right side of (\ref{pep3})
\begin{eqnarray}
\frac{1}{2} (\check{\beta} - \beta_0)' [ \frac{1}{n} \sum_{i=1}^n [ \ddot{\rho}  ( y_i, X_i' \bar{\beta}) - E\ddot{\rho}  ( y_i, X_i' \bar{\beta})]  X_i X_i']
(\check{\beta} - \beta_0)  & \le & \frac{1}{2} v (\check{\beta}- \beta_0)' (\hat{\Sigma} - \Sigma) (\check{\beta} - \beta_0) \nonumber \\
& \le & \frac{v}{2} \underline{\Omega} ( \check{\beta} - \beta_0) \underline{\Omega}_* ( (\hat{\Sigma} - \Sigma) ( \check{\beta} - \beta_0)) \nonumber \\
& \le & \frac{v}{2}   \underline{\Omega} ( \check{\beta} - \beta_0) 
\| (\hat{\Sigma} - \Sigma) ( \check{\beta} - \beta_0) \|_{\infty} \nonumber \\
& \le & \frac{v}{2}  \underline{\Omega} ( \check{\beta} - \beta_0) 
\| \hat{\Sigma} - \Sigma \|_{\infty} \| \check{\beta} - \beta_0 \|_1 \nonumber \\
& \le & \frac{v}{2}  [\underline{\Omega} ( \check{\beta} - \beta_0)]^2 \| \hat{\Sigma} - \Sigma \|_{\infty} \nonumber \\
& \le & \frac{v}{2} M^2 \|\hat{\Sigma} - \Sigma \|_{\infty},\label{pep5}
\end{eqnarray}
where we use Assumption \ref{as4}, to bound $ \ddot{\rho}  ( y_i, X_i' \bar{\beta}) - E\ddot{\rho}  ( y_i, X_i' \bar{\beta})$ uniformly by a positive constant $v$ for the first inequality,
\begin{equation}
| \ddot{\rho} (y_i, X_i' \bar{\beta}) - E \ddot{\rho} (y_i, X_i' \bar{\beta}) | \le 2 | \ddot{\rho} (y_, X_i' \bar{\beta}| \le v ,\label{a.35a}
 \end{equation}
 and for the second inequality we use Lemma \ref{la1}(i), and for the third inequality we use Lemma \ref{la1}(iii), and for the fourth inequality  we use Lemma \ref{la1}(iv), and for the fifth inequality we use Lemma \ref{la1}(ii), and the last inequality is by the norm bound in the statement of the Lemma.  Set $\lambda_e$ by using (\ref{pep3})(\ref{pep4})(\ref{pep5})
\begin{equation}
M \lambda_e \ge \frac{v M^2}{2} \| \hat{\Sigma} - \Sigma \|_{\infty} + \| \frac{1}{n} \sum_{i=1}^n X_i \Xi_i \|_{\infty} M.\label{pep5a}
\end{equation}
We want to show that (\ref{pep5a}) holds with wpa1.
To that end, note that by (\ref{sa2}), (\ref{sa4}), (\ref{sa5}) under Assumptions \ref{as1}, \ref{as2}, and $\Xi$ being uniformly bounded by Assumption 
\ref{as4} 
\begin{equation}
P [ \| \frac{1}{n} \sum_{i=1}^n X_i \Xi_i \|_{\infty} \le t_2 ] \ge 1 - \frac{1}{p^c} - \frac{C}{(ln p)^2},\label{pep6}
\end{equation}
where 
\begin{equation}
t_2 = 2 K [ \sqrt{lnp/n} + \sqrt{E M_2^2} lnp/n] + \sqrt{lnp/n}.\label{pep7}
\end{equation}

We can now define $\lambda_e$ as 
\begin{equation}
\lambda_e := t_1 \frac{v M}{2} + t_2,\label{pep8}
\end{equation}

Then clearly by (\ref{e3.3a})(\ref{e3.4}) (\ref{pep6}) in (\ref{pep5a})  to have 

\[ P \left[ \frac{v M }{2} \| \hat{\Sigma} - \Sigma \|_{\infty} + \| \frac{1}{n} X_i \Xi_i \|_{\infty} \le \lambda_e
\right]
\ge 1 - \frac{1}{p^{2c}} - \frac{1}{p^c} - \frac{5 C}{4 (ln p)^2}.\] 

Combine (\ref{pep3})-(\ref{pep8}) to have the first result.

To have the asymptotics, 
we can show $t_2 = O (\sqrt{lnp/n})$ by Assumptions \ref{as1}-\ref{as2}, and by (\ref{pep6})(\ref{pep7}), and by (\ref{e3.4a})  $t_1 = O_p (\sqrt{lnp/n})$, hence $\lambda_e = O( \sqrt{lnp/n})$.

 {\bf Q.E.D.} 

{\bf Proof of Theorem \ref{thm1}}. 

We start with a definition of a positive sequence, that will go  to zero.
Next define $M$, (multiplied by $\lambda$) with sufficiently large $n$
\begin{equation}
\lambda M : = 8 \left[  \frac{18 \lambda^2 C_{\rho}^2 }{8} ( \Gamma^2 (2, S_0)
)
\right] + 32 \lambda \Omega (\beta_{S_0^c}).\label{po21}
\end{equation}

(i). The proof is divided into several parts. Our proof follows some of the proof of Theorem 7.2 in van de Geer (2016) but differs in the way that, sample effective sparsity condition is used with one margin condition, and we are interested in a norm bound only here rather than both a norm bound and prediction bound in van de Geer (2016). Also, our proof involves specifically Generalized Linear Models; hence a specific effective sparsity is imposed here with a specific convex conjugate condition-inequality. Using sample conditions compared to population conditions complicates our proof compared with van de Geer (2016), especially in using  one-point margin condition. We start by conditioning on the events $E_1, E_2, E_3$, and then we relax this at the end of the proof. Note that proof of Theorem 7.2 of van de Geer (2016) works under high-level conditions; specifically it assumes events $E_1, E_2, E_3$ exist. Our  Lemmata  \ref{la2}-\ref{la4} about $E_1, E_2, E_3$ before our proof here proves them.

Part 1. In this part we define the following $\tilde{\beta}:=  d \hat{\beta} + (1-d) \beta_0$,
where we have 
\[ d:= \frac{M}{M + \underline{\Omega} ( \hat{\beta} - \beta_0)}.\]
So 
\begin{eqnarray}
\underline{\Omega} (\tilde{\beta} - \beta_0) & = & \underline{\Omega} ( d \hat{\beta} + (1 - d) \beta_0 - \beta_0) \nonumber \\
& = & d \underline{\Omega} ( \hat{\beta} - \beta_0) \nonumber \\
& = & \frac{M \underline{\Omega} (\hat{\beta} - \beta_0)}{M + \underline{\Omega} ( \hat{\beta} - \beta_0)} \le M,\label{po1}
 \end{eqnarray}
 
 So clearly we show that $\tilde{\beta} \in {\cal B}_{local}$ which is a convex subset of ${\cal B}$. This helps us in the coming steps, since we need this result so that we can use one point margin condition in Lemma \ref{la2}. Note that $R_n (.) + \lambda \Omega (.)$ is convex and $\hat{\beta} $ is the minimizer of the sample objective function we get 
 \begin{eqnarray}
 R_n (\tilde{\beta}) + \lambda \Omega (\tilde{\beta})& \le &d R_n (\hat{\beta}) + d \lambda \Omega (\hat{\beta}) 
 + (1-d) R_n (\beta_0) + (1-d) \lambda \Omega (\beta_0) \nonumber \\
 & \le & R_n (\beta_0) + \lambda \Omega (\beta_0).\label{po2}
  \end{eqnarray}
  Add to both sides of (\ref{po2}) $R (\tilde{\beta}) - R ( \beta_0)$ to have 
  \begin{equation}
  R (\tilde{\beta}) - R ( \beta_0) \le - [ (R_n (\tilde{\beta}) - R (\tilde{\beta})) - (R_n (\beta_0) - R (\beta_0))] 
  + \lambda \Omega (\beta_0) - \lambda \Omega (\tilde{\beta}).\label{po3}
      \end{equation}
Then since we use the event $E_3$ in (\ref{e3}), we can rewrite (\ref{po3}) as, with $\tilde{\beta} \in {cal B}_{local} \subset {\cal B}$ 
\begin{equation}
R (\tilde{\beta}) - R (\beta_0) \le \lambda_e M + \lambda \Omega (\beta_0) - \lambda \Omega (\tilde{\beta}).\label{po4}
\end{equation} 
Then we simplify the last two terms on the right side of (\ref{po4}).
Next, by definition 7.6 or 6.1 of van de Geer (2016), with $\beta_{S_0}$ as an allowed vector, and since  weak decomposability holds, Lemma 7.3 of van de Geer (2016) triangle property holds at $\beta_{S_0}$(definition 7.4 of van de Geer (2016)). 

Then by Lemma 7.2 of van de Geer (2016) provides  Corollary 7.1 of van de Geer (2016) which is the following inequality 
\[ \Omega (\beta_0) - \Omega (\tilde{\beta}) \le \Omega (\tilde{\beta}_{S_0} - \beta_{S_0}) - \Omega^{S_0^c} (\tilde{\beta}_{S_0^c} - \beta_{S_0^c})
+ 2 \Omega (\beta_{S_0}^c)\]
  Use this last inequality in (\ref{po4}) to have 
\begin{equation}
R (\tilde{\beta}) - R (\beta_0) \le \lambda_e M +\lambda \Omega (\tilde{\beta}_{S_0} - \beta_{S_0}) - \lambda \Omega^{S_0^c} (\tilde{\beta}_{S_0^c} - \beta_{S_0^c}) + 2 \lambda \Omega (\beta_{S_0^c}) .\label{po5}
\end{equation}
This last inequality will be utilized a lot in the subsequent proof and can also be seen in p.115 of van de Geer (2016).

Part 2. This part has two sections, a and b. We provide a norm inequality depending on two conditions. Then in the next part, we merge results under  two conditions to get the result.

Part 2a. Part 2a will impose the following condition:
\begin{equation}
\lambda \Omega (\tilde{\beta}_{S_0} - \beta_{S_0}) \le \lambda_e M + 2 \lambda \Omega (\beta_{S_0}^c).\label{po6}
\end{equation}
Now by (\ref{po5}) and since $\beta_0$ minimizes $R(\beta)$ we have $R (\tilde{\beta}) - R (\beta_0) \ge 0$, to have 
\begin{equation}
0 \le \lambda_e M + \lambda \Omega (\tilde{\beta}_{S_0} - \beta_{S_0}) - \lambda \Omega^{S^c} ( \tilde{\beta}_{S_0^c}- \beta_{S_0^c}) + 2 \lambda \Omega (\beta_{S_0^c})
.\label{po8}
\end{equation}
Then 
\begin{equation}
\lambda \Omega^{S^c} ( \tilde{\beta}_{S_0^c} - \beta_{S_0^c}) \le \lambda_e M + \lambda \Omega (\tilde{\beta}_{S_0} - \beta_{S_0}) + 2 \lambda \Omega (\beta_{S_0^c})
.\label{po9}
\end{equation}
Use (\ref{po6}) on the second term on the right side of (\ref{po9})
\begin{equation}
\lambda \Omega^{S^c} ( \tilde{\beta}_{S_0^c} - \beta_{S_0^c}) \le \lambda_e M + \lambda_e M+2 \lambda \Omega (\beta_{S_0^c})+2 \lambda \Omega (\beta_{S_0^c})=  2 \lambda_e M+ 4 \lambda \Omega (\beta_{S_0^c}).\label{po10}
\end{equation}
Add (\ref{po6}) to both sides of (\ref{po10}) to have 
\begin{equation}
\lambda [ \Omega (\tilde{\beta}_{S_0}  - \beta_{S_0}) + \Omega^{S^c} (\tilde{\beta}_{S_0^c} - \beta_{S_0^c})] \le 3 \lambda_e M + 6 \lambda \Omega (\beta_{S_0^c}).\label{po11}
\end{equation}
Next since $\underline{\Omega}( \tilde{\beta} - \beta_0) := \Omega (\tilde{\beta}_{S_0} - \beta_{S_0}) + \Omega^{S^c} (\tilde{\beta}_{S_0^c} -\beta_{S_0^c})$
\begin{equation}
\lambda \underline{\Omega} (\tilde{\beta} - \beta_0) \le 3 \lambda_e M + 6 \lambda \Omega (\beta_{S_0^c}).\label{po12}
\end{equation}

Now we go back and simplify (\ref{po12}). We use definition of M in (\ref{po21}), since it holds regardless of conditions (\ref{po6})or (\ref{po13}). 

(\ref{po12}) can be rewritten as, with $\lambda \ge 16 \lambda_e$, $\frac{M}{32} \ge \Omega (\beta_{S_0^c})$ by (\ref{p021}) with sufficiently large $n$
\begin{equation}
\lambda \underline{\Omega} (\tilde{\beta} - \beta_0) \le 3 \lambda_e M + 6 \lambda \Omega (\beta_{S_0^c}) \le
\frac{3}{16} M \lambda + \frac{3}{16} M \lambda = \frac{6}{16} M \lambda \le M \lambda/2.\label{po24a}
\end{equation}
This ends part 2a.

Part 2b.  This part differs from proof of Theorem 7.2 of van de Geer (2016), and considers sample one-point margin which introduces additional difficulty.
Now we reverse the condition in part 2a and try to get a similar bound as in part 2a. We impose only in this subsection
\begin{equation}
\lambda \Omega (\tilde{\beta}_{S_0} - \beta_{S_0} ) \ge \lambda_e M + 2 \lambda \Omega (\beta_{S_0^c}).\label{po13}
\end{equation}
Now, use (\ref{po5}), by seeing $R (\tilde{\beta}) - R (\beta_0) \ge 0$, and the condition in (\ref{po13})
\begin{equation}
\lambda \Omega^{S^c} (\tilde{\beta}_{S_0^c} - \beta_{S_0^c}) \le 2\lambda \Omega (\tilde{\beta}_{S_0} - \beta_{S_0}).\label{po14}
\end{equation}
This clearly shows that $\Omega^{S_0^c} ( \tilde{\beta}_{S_0^c} - \beta_{S_0^c}) \le 2 \Omega (\tilde{\beta}_{S_0} - \beta_{S_0})$ and satisfy the cone condition in effective sparsity definition with $L=2$. Then add $(\frac{\lambda}{2}) \Omega ( \tilde{\beta}_{S_0} - \beta_{S_0})$ to both sides of (\ref{po5}) to get
\begin{equation}
R (\tilde{\beta} ) - R (\beta_0) + \lambda \Omega^{S_0^c} (\tilde{\beta}_{S_0^c} - \beta_{S_0^c}) + \frac{\lambda}{2} \Omega (\tilde{\beta}_{S_0} - \beta_{S_0})
\le \frac{3 \lambda}{2} \Omega (\tilde{\beta}_{S_0} - \beta_{S_0}) + \lambda_e M  + 2 \lambda \Omega (\beta_{S_0^c}).\label{po15} 
\end{equation}

Since the cone condition is satisfied, we can use the effective sparsity on the first right-side term in (\ref{po15}) via (\ref{ev00})(\ref{ev01}), and divide and multiply the first term on the right side by $0< C_{\rho} < \infty$, which is a positive constant that depends on the shape of $\rho(.)$
\begin{equation}
R (\tilde{\beta} ) - R (\beta_0) + \lambda \Omega^{S^c} (\tilde{\beta}_{S_0^c} - \beta_{S_0^c}) + \frac{\lambda}{2} \Omega (\tilde{\beta}_{S_0} - \beta_{S_0})
\le \frac{3 \lambda C_{\rho}}{2}  \left[ \frac{\| X ( \tilde{\beta} - \beta_0 ) \|_n}{C_{\rho}}
\right] \hat{\Gamma} (2, S_0) + \lambda_e M + 2 \lambda \Omega (\beta_{S_0^c}).\label{po16} 
\end{equation}
After that use the convex conjugate condition for generalized linear models $u v \le u^2/2 + v^2/2$, by taking 
$u := \| X (\tilde{\beta} - \beta_0) \|_n/ C_{\rho}, v := \frac{3 \lambda C_{\rho}}{2} \hat{\Gamma} (2, S_0)$ 
\begin{equation}
R (\tilde{\beta} ) - R (\beta_0) + \lambda \Omega^{S^c} (\tilde{\beta}_{S_0^c} - \beta_{S_0^c}) + \frac{\lambda}{2} \Omega (\tilde{\beta}_{S_0} - \beta_{S_0})
\le \frac{9 \lambda^2 C_{\rho}^2}{8} [\hat{\Gamma} (2, S_0)]^2+   \left[ \frac{\| X ( \tilde{\beta} - \beta_0) \|_n^2}{2 C_{\rho}^2}
\right]  + \lambda_e M + 2 \lambda \Omega (\beta_{S_0^c}).\label{po17} 
\end{equation}
Now we use $E_1$ on the second right side term in (\ref{po17}) 
\begin{equation}
R (\tilde{\beta} ) - R (\beta_0) + \lambda \Omega^{S^c} (\tilde{\beta}_{S_0^c} - \beta_{S_0^c}) + \frac{\lambda}{2} \Omega (\tilde{\beta}_{S_0} - \beta_{S_0})
\le \frac{9 \lambda^2 C_{\rho}^2}{8} [\hat{\Gamma} (2, S_0)]^2+   \left[  (R (\tilde{\beta}) - R (\beta_0)) + \frac{M^2 t_1}{2 C_{\rho}^2}
\right] + \lambda_e M + 2 \lambda \Omega (\beta_{S_0^c}).\label{po18} 
\end{equation}
Note that cancelling the first term on the left side and second term on the right side provides
\begin{equation}
 \lambda \Omega^{S^c} (\tilde{\beta}_{S_0^c} - \beta_{S_0^c}) + \frac{\lambda}{2} \Omega (\tilde{\beta}_{S_0} - \beta_{S_0})
\le \frac{9 \lambda^2 C_{\rho}^2}{8} [\hat{\Gamma} (2, S_0)]^2+   \left[   \frac{M^2 t_1}{2 C_{\rho}^2}
\right] + \lambda_e M + 2 \lambda \Omega (\beta_{S_0^c}).\label{po19} 
\end{equation}
Then use $E_2$ for the link between effective sparsity in sample and in population to have, by (\ref{ev07}) 
\begin{equation}
 \lambda \Omega^{S^c} (\tilde{\beta}_{S_0^c} - \beta_{S_0^c}) + \frac{\lambda}{2} \Omega (\tilde{\beta}_{S_0} - \beta_{S_0})
\le \frac{9 \lambda^2 C_{\rho}^2}{8} \left[ \frac{\Gamma^2 (2, S_0)}{ 1 - 9 t_1 \Gamma^2 (2, S_0)}
\right]+   \left[   \frac{M^2 t_1}{2 C_{\rho}^2}
\right] + \lambda_e M + 2 \lambda \Omega (\beta_{S_0^c}).\label{po20} 
\end{equation}
After this inequality note that by Assumptions \ref{as3}- \ref{as4}, we have  $v \ge 1/C_{\rho}^2$ via (\ref{a.35a}) and by definition $t_2 \ge 0$, so 
we have 
\begin{equation}
\lambda_e \ge \frac{M t_1}{2 C_{\rho}^2},\label{eqle}
\end{equation}
 by $\lambda_e$ definition in (\ref{pep8}). 
Note that with sufficiently large $n$ since $t_1=o(1)$, and $\Gamma^2 (2, S_0)$ is a constant we have 
\begin{equation}
9 t_1 \Gamma^2 (2, S_0) \le 1/2.\label{star}
\end{equation}

Using  (\ref{po21}) for the first and fourth term on the right side of (\ref{po20}) and the inequality (\ref{eqle}) for $\lambda_e$ in  the second right side term in (\ref{po20}), and dividing the first term by 2, with sufficiently large $n$
\begin{equation}
 \frac{\lambda}{2} \Omega^{S^c} (\tilde{\beta}_{S_0^c} - \beta_{S_0^c}) + \frac{\lambda}{2} \Omega (\tilde{\beta}_{S_0} - \beta_{S_0})
\le \frac{\lambda M}{ 8} +  2 \lambda_e M.\label{po22} 
\end{equation}
Then use $\lambda \ge 16 \lambda_e$ on the right side of (\ref{po22}) to have 
\begin{equation}
\frac{\lambda}{2} \underline{\Omega} (\tilde{\beta} - \beta_0) \le \lambda M /4,\label{po23}
\end{equation}
by definition of $\underline{\Omega} ( \tilde{\beta} - \beta_0)$. By simple algebra in (\ref{po23}) we get 
\begin{equation}
\underline{\Omega} (\tilde{\beta} - \beta_0) \le M /2.\label{po24}
\end{equation}
So under condition (\ref{po13}) we have (\ref{po24}).

Part 3. Here we merge two cases in part 2. See that  under both conditions (\ref{po6}) and (\ref{po13}) via (\ref{po24a})(\ref{po24})
\[ \underline{\Omega} (\tilde{\beta} - \beta_0) \le M/2.\]

By (\ref{po1})
\[ \underline{\Omega} ( \hat{\beta} - \beta_0) = \frac{\underline{\Omega}( \tilde{\beta} - \beta_0)}{d}
= \frac{ \underline{\Omega}( \tilde{\beta} - \beta_0)}{M} [ M + \underline{\Omega}(\hat{\beta} - \beta_0)],\]
Arrange the last expression to have 

\[ \underline{\Omega} ( \hat{\beta} - \beta_0)  = \frac{\underline{\Omega} (\tilde{\beta} - \beta_0)}{1 - \underline{\Omega} (\tilde{\beta} - \beta_0)/M}
\le \frac{M/2}{1 - (M/2)/M} = M.\]\\

Proof follows by event definitions, $E_1, E_2, E_3$ and Lemma \ref{la2}, \ref{la3}, \ref{la4} under Assumptions \ref{as1}-\ref{as4}.\\

(ii). To get asymptotics  set $\lambda = 16 \lambda_e = O (\sqrt{\frac{lnp}{n}})$ by Lemma \ref{la4}.

Note that both results in (i)-(ii) are uniform  over ${\cal B}_{l_0} (s_0) $ follows by noticing that right hand side of equations (\ref{po21})-(\ref{po24}) only depend on $\beta_0$ through $s_0, S_0$.

{\bf Q.E.D.}\\

{\bf Proof of Theorem \ref{thm2}}. We want to simplify the numerator of the test statistic. We start with our formula for desparsified structured sparsity estimator.
\begin{eqnarray*}
\alpha' (\hat{b} - \beta_{0})& =& \alpha' (\hat{\beta} - \beta_{0}) - \alpha' \hat{\Theta} [ \frac{1}{n} \sum_{i=1}^n \dot{\rho} (y_i, X_i' \hat{\beta}) X_i] \\
& = & \alpha' (\hat{\beta} - \beta_{0})  - \alpha' \hat{\Theta} [ \frac{1}{n} \sum_{i=1}^n \dot{\rho} (y_i, X_i' \beta_0) X_i + \frac{1}{n} \sum_{i=1}^n \ddot{\rho} (y_i, \tilde{a}_i) X_i X_i' (\hat{\beta} - \beta_0)],
\end{eqnarray*}
where $\tilde{a}_i \in (X_i' \beta_0, X_i' \hat{\beta})$ and we use the mean value theorem. 

Add and subtract from the above $\hat{\Theta}[n^{-1} \sum_{i=1}^n \ddot{\rho} (y_i, X_i' \hat{\beta}) X_i X_i' (\hat{\beta} - \beta_0)]$ 
\begin{eqnarray*}
\alpha' (\hat{b} - \beta_{0})& =&  - \alpha' \hat{\Theta}  [ \frac{1}{n} \sum_{i=1}^n \dot{\rho} (y_i, X_i' \beta_0) X_i]  \\
& - & \alpha' \left(  \hat{\Theta} \frac{1}{n} \sum_{i=1}^n \ddot{\rho} (y_i, X_i' \hat{\beta}) X_i X_i' - I_p
\right) (\hat{\beta} - \beta_0) \\
& - & \alpha' \hat{\Theta} \left[ \frac{1}{n} \sum_{i=1}^n [ \ddot{\rho} (y_i, \tilde{a}_i) - \ddot{\rho}(y_i, X_i' \hat{\beta})] X_i X_i' \right] (\hat{\beta} - \beta_0).
\end{eqnarray*}

Now we form our test statistic in a way that it reflects these three terms on the right side above.
 To that effect define 
 \begin{equation}
 \hat{V}_{\alpha}^2:=  \alpha' \hat{\Theta} [ \frac{1}{n} \sum_{i=1}^n \dot{\rho} (y_i, X_i' \hat{\beta})^2 X_i X_i' ] \hat{\Theta}' \alpha\label{hsig}
 \end{equation}
 and 
 \begin{equation}
 V_{\alpha}^2:= \alpha' \Theta [ E X_i X_i' \dot{\rho} (y_i, X_i' \beta_0)^2] \Theta \alpha.\label{sig}
 \end{equation}
  Furthermore define
 \[ tt_1:= \frac{-  \alpha' \hat{\Theta} [ \frac{1}{n^{1/2}} \sum_{i=1}^n \dot{\rho} (y_i, X_i' \beta_0) X_i]}{\hat{V}_{\alpha}}.\]
 \[ tt_2:=-\frac{n^{1/2} \alpha'  \left(  \hat{\Theta} \frac{\sum_{i=1}^n \ddot{\rho} (y_i, X_i' \hat{\beta}) X_i X_i'}{n} - I_p
\right) (\hat{\beta} - \beta_0) }{\hat{V}_{\alpha}}.\]
\[ tt_3:=- \frac{n^{1/2} \alpha'  \hat{\Theta} \left[ \frac{1}{n} \sum_{i=1}^n [ \ddot{\rho} (y_i, \tilde{a}_i) - \ddot{\rho}(y_i, X_i' \hat{\beta})] X_i X_i' \right] (\hat{\beta} - \beta_0) }{\hat{V}_{\alpha}}.\]
 
 Clearly our test statistic is 
\[ \frac{n^{1/2} \alpha'  (\hat{b} - \beta_{0})}{\hat{V}_{\alpha}^2} = tt_1 + tt_2 + tt_3.\]

Our proof consists of several steps. We will show $tt_1 \stackrel{d}{\to} N(0,1)$, $tt_2 = o_p (1)$, $tt_3=o_p (1)$. So the proof will follow two steps to prove these three claims.

{\bf Step 1}.  We consider $tt_1$. To start we define the following infeasible test statistic.
\[ tt_1':= \frac{ -\alpha' \Theta \sum_{i=1}^n \dot{\rho} (y_i, X_i' \beta_0)/n^{1/2}}{V_{\alpha}}.\]
Step 1 has two sub parts, First we show that $tt_1' \stackrel{d}{\to} N(0,1)$, then $tt_1 - tt_1' = o_p (1)$.

Step 1a.  We want to show $tt_1' \stackrel{d}{\to} N(0,1)$ here. First note that $E X_i \dot{\rho} (y_i, X_i' \beta_0)=0$ since $\beta_0$ is the minimizer of the population objective function and the objective function is  differentiable.  So 
\[ E \frac{  \alpha' \Theta [\sum_{i=1}^n \dot{\rho} (y_i , X_i' \beta_0) X_i ]/n^{1/2}}{V_{\alpha}} = 0.\]
Next using $V_{\alpha}^2$ definition  (\ref{sig}) in the denominator and independence
\[ E \left[ 
\frac{\alpha' \Theta \sum_{i=1}^n \dot{\rho} (y_i, X_i' \beta_0) X_i /n^{1/2}}{\sqrt{ \alpha' \Theta [E X_i X_i' \dot{\rho} (y_i, X_i' \beta_0)^2 ] \Theta' \alpha}}
\right]^2=1.\]

Before the next condition we need the following results, since $\| \alpha \|=1$, by (\ref{h1}), and $\Theta$ being symmetric
\begin{equation}
\| \alpha' \Theta \|_1 = \| \Theta \alpha \|_1 = \| \sum_{j \in {\cal H}} \Theta_j \alpha_j \|_1 \le  \sum_{j \in {\cal H}} |\alpha_j| \| \Theta_j \|_1 = O (\sqrt{h \bar{s}}),\label{a55a}
\end{equation}
where we use 
\begin{equation}
\| \Theta_j \|_1 = \| \frac{\gamma_{\beta_0,j}}{\tau_j^2} \|_1 =
 O (\sqrt{\bar{s}}),\label{pt2.1}
 \end{equation} 
by (\ref{nw1})(\ref{nw1a}), Assumption \ref{as1}, \ref{as4}(ii) with Lemma B.6, and exactly the same analysis  in (B.48)-(B.55) of Caner and Kock (2018).

Note that nonzero entries of $\alpha' \Theta$ contained in $\bar{S}:= \cup_{j \in {\cal H}} S_j$ which has cardinality at most $h \bar{s} \wedge p$.
Then see that  by Assumption \ref{as1}, \ref{as4} with  (\ref{a55a})
 \begin{eqnarray*}
 E | \alpha' \Theta X_i \dot{\rho} (y_i, X_i' \beta_0)/n^{1/2}|^{r_x/2} & \le &  E [ | \| \alpha' \Theta \|_1^{r_x/2} \max_{l \in \bar{S}} | X_{i,l} \dot{\rho} (y_i, X_i' \beta_0)/n^{1/2}|^{r_x/2}] \\
 & \le & O (\frac{(h\bar{s})^{r_x/4}}{n^{r_x/4}}) (h\bar{s} \wedge p) \max_{l \in  \bar{S }}E | X_{i,l} \ddot{\rho} (y_i, X_i' \beta_0)|^{r_x/2} \\
 & = & O (\frac{(h\bar{s})^{(r_x/4)+1} \wedge(h \bar{s})^{r_x/4}  p}{n^{r_x/4}}).
 \end{eqnarray*}
 Next 
 \[ \sum_{i=1}^n E | \alpha' \Theta X_i \dot{\rho} (y_i, X_i' \beta_0)/n^{1/2}|^{r_x/2} = O ( \frac{(h\bar{s})^{(r_x/4)+1} \wedge(h \bar{s})^{r_x/4}  p}{n^{r_x/4-1}}) = o(1),\]
 by Assumption \ref{as7}. Then see that by using $\Theta:= \Sigma_{\beta_0}^{-1}$ definition
 
 \begin{eqnarray}
 V_{\alpha}^2 & = & 
 ( \alpha' \Theta E X_i X_i' \dot{\rho} (y_i, X_i' \beta_0)^2 \Theta \alpha) \nonumber  \\
 &\ge  & Eigmin ( E X_i X_i' \dot{\rho} (y_i, X_i' \beta_0)^2 ) \| \Theta' \alpha\|_2^2 \nonumber\\
 & \ge & Eigmin ( E X_i X_i' \dot{\rho} (y_i, X_i' \beta_0)^2 )  
Eigmin^2 (\Theta) \nonumber  \\
& = &  Eigmin ( E X_i X_i' \dot{\rho} (y_i, X_i' \beta_0)^2 )[Eigmax^2 (\Sigma_{\beta_0})]^{-1} \nonumber\\
& \ge & c/C^2 > 0,\label{a56a}
  \end{eqnarray}
by Assumption \ref{as9}, and $\| \alpha \|_2=1$. So since Lyapounov condition is satisfied CLT holds and 
\[ tt_1' \stackrel{d}{\to} N(0,1).\]

Step 1b. We want to show that $tt_1 - tt_1' = o_p (1)$. To do that, we want to show that the denominators of both are asymptotically equivalent, then show numerators are asymptotically equivalent, since $V_{\alpha}^2$ is bounded away from 0 as in (\ref{a56a}) as shown in Step 1a, this will suffice to prove two tests are asymptotically equivalent.

Specifically we want to prove first 
\begin{equation}
 | \hat{V}_{\alpha}^2 - V_{\alpha}^2| = o_p (1).\label{cov}
\end{equation}

To prove (\ref{cov}) we need the following three results and triangle inequality, with (\ref{hsig})-(\ref{sig})

\begin{equation}
 \left| \alpha' \hat{\Theta} \left[ \frac{1}{n} \sum_{i=1}^n X_i X_i'  \dot{\rho} (y_i, X_i' \hat{\beta})^2 \right]  \hat{\Theta}' \alpha - 
\alpha' \hat{\Theta} \left[ \frac{1}{n} \sum_{i=1}^n X_i X_i'  \dot{\rho} (y_i, X_i' \beta_0)^2 \right]  \hat{\Theta}' \alpha \right| = o_p (1).\label{cov1}
\end{equation}

\begin{equation}
 \left| \alpha' \hat{\Theta} \left[ \frac{1}{n} \sum_{i=1}^n X_i X_i'  \dot{\rho} (y_i, X_i' \beta_0)^2 \right]  \hat{\Theta}' \alpha - 
\alpha' \hat{\Theta} \left[ E X_i X_i'  \dot{\rho} (y_i, X_i' \beta_0)^2 \right]  \hat{\Theta}' \alpha \right| = o_p (1).\label{cov2}
\end{equation}

\begin{equation}
 \left| \alpha' \hat{\Theta} \left[ E  X_i X_i'  \dot{\rho} (y_i, X_i' \beta_0)^2 \right]  \hat{\Theta}' \alpha- 
\alpha' \Theta \left[ E X_i X_i'  \dot{\rho} (y_i, X_i' \beta_0)^2 \right]  \Theta \alpha \right| = o_p (1).\label{cov3}
\end{equation}

Before starting our proofs we  need the following result:

\begin{equation}
\| \alpha' \hat{\Theta} \|_1 = \| \hat{\Theta}' \alpha \|_1 = \| \sum_{j \in {\cal H}} \hat{\Theta}_j \alpha_j \|_1\le ( \sum_{j \in {\cal H}} | \alpha_j |)  \max_{j \in {\cal H}}
\| \hat{\Theta}_j \|_1,\label{a60a}
\end{equation}
where we use $\alpha$ definition in the second equality.

\begin{eqnarray}
\max_{j \in {\cal H}} \| \hat{\Theta}_j \|_1& \le & \max_{j \in {\cal H}} \| \hat{\Theta}_j - \Theta_j \|_1 + \max_{j \in {\cal H}} \| \Theta_j \|_1 \nonumber \\
& \le & \max_{j \in {\cal H}} \underline{\Omega} (\hat{\Theta}_j - \Theta_j ) + \max_{j \in {\cal H} } \| \Theta_j \|_1 \nonumber \\
& = & o_p (1) + O (\bar{s}^{1/2}) = O_p ( \bar{s}^{1/2}),\label{pt2.2}
\end{eqnarray}
where we use triangle inequality for the first inequality, and the Lemma \ref{la1}(ii) for the second inequality, and for the rates we use Lemma \ref{la10}, 
and (\ref{pt2.1}). So by (\ref{pt2.2})(\ref{h1}) in (\ref{a60a})
\begin{equation}
\| \hat{\Theta} \alpha \|_1 = O_p ( \sqrt{h \bar{s}}).\label{a61a}
\end{equation}

To make the proofs easier to understand, we define

\begin{equation}
\hat{A}_n:=\left[ \frac{1}{n} \sum_{i=1}^n X_i X_i'  \dot{\rho} (y_i, X_i' \hat{\beta})^2 \right]  \label{pt2.3a}
\end{equation}

\begin{equation}
A_n:=\left[ \frac{1}{n} \sum_{i=1}^n X_i X_i'  \dot{\rho} (y_i, X_i' \beta_0)^2 \right]\label{pt2.3b}
\end{equation}

\begin{equation}
A:= E X_i X_i'  \dot{\rho} (y_i, X_i' \beta_0)^2.\label{pt2.3c}
\end{equation}

Now consider (\ref{cov1}). 

\begin{eqnarray}
 \left| \alpha' \hat{\Theta} (\hat{A}_n - A_n)   \hat{\Theta}' \alpha \right| &\le& \max_{1 \le j \le p} 
 \| \alpha'  \hat{\Theta} \|_1   \| [\frac{1}{n} \sum_{i=1}^n  X_i X_i' (\dot{\rho} (y_i, X_i' \hat{\beta})^2 - \dot{\rho} (y_i, X_i' \beta_0)^2)] \hat{\Theta}' \alpha \|_{\infty} 
 \nonumber \\
 & \le & 
 \| \alpha' \hat{\Theta} \|_1]^2 \| \frac{1}{n} \sum_{i=1}^n  X_i X_i' (\dot{\rho} (y_i, X_i' \hat{\beta})^2 - \dot{\rho} (y_i, X_i' \beta_0)^2)  \|_{\infty},\label{pt2.3d}
  \end{eqnarray}
where we use Holders inequality for the first one, and then  Lemma \ref{la2}(iv). Now analyze the second term on the right side above, to simplify further
see that, as in p.1198 of van de Geer et al. (2014)
\begin{equation}
\dot{\rho} (y_i, X_i' \hat{\beta})^2 - \dot{\rho} (y_i, X_i' \beta_0)^2 = u_i X_i' (\hat{\beta} - \beta_0),\label{pt2.4}
\end{equation} 
where we use mean value theorem, with $u_i:= 2 \ddot{\rho} (y_i, \tilde{a}_i)$, $\tilde{a}_i \in (X_i' \beta_0, X_i' \hat{\beta})$ and we use Assumption \ref{as4} to have 
$\max_i | u_i | = O (1)$.
Now 
\[ \| \hat{A}_n - A_n \|_{\infty} \le \max_i | u_i |  \| \frac{1}{n} X_i X_i' X_i' (\hat{\beta} - \beta_0)\|_{\infty} = O (1)  \| \frac{1}{n} X_i X_i' X_i' (\hat{\beta} - \beta_0)\|_{\infty}.\]
Next, by Cauchy-Schwartz inequality
\[
\max_{1 \le k, l \le p} \left| \frac{1}{n} \sum_{i=1}^n X_{i,k} X_{i,l} X_i' (\hat{\beta} - \beta_0) 
\right| \le \sqrt{ \max_{1 \le k,l \le p} \frac{1}{n} \sum_{i=1}^n X_{i,k}^2 X_{i,l}^2} \| X (\hat{\beta} - \beta_0) \|_n \]
To get the rate above we use 
\begin{equation}
\max_{1 \le k \le p} \max_{1 \le l \le p} | \frac{1}{n} \sum_{i=1}^n X_{ik}^2 X_{il}^2 - E X_{ik}^2 X_{il}^2| = O_p ( \sqrt{\frac{lnp}{n}}),\label{pt2.4aa}
\end{equation}
by seeing $ln p^2 = 2 ln p$, and using (\ref{sa4})(\ref{sa5}). To use (\ref{sa4})(\ref{sa5}) we need  Assumption \ref{as7}(i)-Cauchy-Schwartz inequality and  that $ \max_{ 1 \le k \le p}  E |X_{ik}|^{r_x} \le C < \infty$, with $r_x > 8$. Using this moment condition and Cauchy-Schwartz inequality together  we also get 
\begin{equation}
\max_{1 \le k \le p} \max_{1 \le l \le p} E | X_{ik}^2 X_{il}^2| \le C < \infty.\label{pt2.4ab}
\end{equation}
Then combine (\ref{pt2.4aa})(\ref{pt2.4ab}) to have 
\[ \max_{1 \le k \le p} \max_{1 \le l \le p} \frac{1}{n} \sum_{i=1}^n X_{ik}^2 X_{il}^2 = O_p (1).\]
Next we use (\ref{oi}) with $\lambda$ rate, and (\ref{oi}) does not depend on Central Limit Theorem proofs here.

\noindent Then combine all above to have 
\begin{equation}
\| \hat{A}_n - A_n \|_{\infty} = O_p ( s_0  \sqrt{\frac{lnp}{n}}).\label{pt2.5}
\end{equation}
\noindent Now, use $\hat{A}_n, A_n$ definition in (\ref{pt2.3d}) with (\ref{a61a})(\ref{pt2.5})
\begin{equation}
\left| \alpha' \hat{\Theta}  (\hat{A}_n - A_n)   \hat{\Theta}' \alpha \right| = O_p (h \bar{s} s_0 \sqrt{\frac{lnp}{n}})=o_p (1),\label{pt2.6} 
\end{equation}
where  we use Assumption \ref{as7}. Now consider (\ref{cov2}).
\begin{equation}
| \alpha' \hat{\Theta} (A_n - A ) \hat{\Theta}' \alpha  | \le \| \hat{\Theta}' \alpha \|_1^2  \| A_n - A \|_{\infty},\label{pt2.7}
\end{equation}
as in (\ref{pt2.3d}). We see that by Assumption \ref{as4}, $\max_{a_0 \in X_i' \beta_0} \sup_{y_i} | \dot{\rho} (y_i, X_i' \beta_0)| = O(1)$ uniformly over ${\cal B}_{l_0} (s_0)$, and by (\ref{e3.3a})-(\ref{e3.4a})
\begin{equation}
\| A_n - A \|_{\infty} = O_p ( \sqrt{\frac{lnp}{n}}).\label{pt2.8}
\end{equation}

\noindent Now use (\ref{pt2.8}) and (\ref{a61a}) in (\ref{pt2.7}) to have 
\begin{equation}
 | \alpha' \hat{\Theta} (A_n - A ) \hat{\Theta}' \alpha | = O_p ( h \bar{s} \sqrt{\frac{lnp}{n}}) = o_p (1).\label{pt2.9}
\end{equation}
where we use Assumption \ref{as7}(iii). Now we consider (\ref{cov3}) term above.  We use Lemma \ref{la1}(ii)(iv), Lemma \ref{la11}, Lemma \ref{la12} with $v:= \Theta' \alpha, \hat{v}:=
\hat{\Theta}' \alpha$ to get the inequality in (\ref{pt2.10})
\begin{eqnarray}
| \alpha' \hat{\Theta} A \hat{\Theta}' \alpha - \alpha' \Theta A \Theta' \alpha | & \le & 
\left[ (\max_{j \in {\cal H}} \underline{\Omega} (\hat{\Theta}_j - \Theta_j )) \sum_{ j \in {\cal H}} | \alpha_j | 
\right]^2 \| A \|_{\infty} \nonumber \\
& + & 2 \| A \|_{\infty} \| \Theta' \alpha \|_1 [ \max_{j \in {\cal H}} \underline{\Omega} (\hat{\Theta}_j - \Theta_j)  \nonumber \\
& = & \left[ O_p \left( g_n  \bar{s}^{1/2} \sqrt{\frac{lnp}{n}} \max (\bar{s}, H_n^2 s_0^2)\right) 
\right]^2 O (h) O(1) \nonumber \\
& + & O (1) O ( h^{1/2} \bar{s}^{1/2})  O_p ( g_n \bar{s}^{1/2} \sqrt{\frac{lnp}{n}} \max (\bar{s}, H_n^2 s_0^2)) \nonumber \\
& = & O_p \left( (h^{1/2}  \bar{s} g_n \sqrt{\frac{lnp}{n}} \max (\bar{s}, H_n^2 s_0^2)) \right) = o_p (1) ,\label{pt2.10}
\end{eqnarray}
where the rate is derived by Lemma \ref{la10}, (\ref{h1}), $\| A \|_{\infty} = O(1)$ by uniformly bounded first order partial derivative by Assumption \ref{as4}, and Assumption \ref{as1} on bounds on covariates, $\| \Theta' \alpha \|_1 = O (\sqrt{h \bar{s}})$ by the proof in (\ref{pt2.2})(\ref{a61a}), and by Assumption \ref{as7} to get asymptotically negligible result. Combine (\ref{pt2.6})(\ref{pt2.9})(\ref{pt2.10}) to have the desired result (\ref{cov}). See that uniformity over $l_0$ ball ${\cal B}_{l_0} (s_0)$ result follows by observing on (\ref{pt2.6})(\ref{pt2.10}) results depend on $\beta_0$ through $s_0$ only.

Now we want to show that the numerators of $tt_1$, $tt_1'$ are asymptotically equivalent, uniformly over $j=1,\cdots, p$. Namely we want to prove 
\[ \frac{ \alpha' (\hat{\Theta} - \Theta)' \sum_{i=1}^n \dot{\rho} (y_i, X_i' \beta_0) X_i}{n^{1/2}} = o_p (1).\]

To show that 

\begin{eqnarray}
\left|  \frac{n^{1/2} \alpha'  (\hat{\Theta} - \Theta) \sum_{i=1}^n \dot{\rho} (y_i, X_i' \beta_0) X_i}{n}
\right| & \le & \|\alpha'  \underline{\Omega} (\hat{\Theta}- \Theta) \|_1 \nonumber \\
&\times &  n^{1/2} \| n^{-1} \sum_{i=1}^n \dot{\rho} (y_i, X_i' \beta_0) X_i \|_{\infty}\nonumber \\
& \le & [ \max_{ j \in {\cal H}} \underline{\Omega} ( \hat{\Theta}_j - \Theta_j)]
[ \sum_{j \in {\cal H} } | \alpha_j | ]\nonumber \\
&\times &  n^{1/2} \| n^{-1} \sum_{i=1}^n \dot{\rho} (y_i, X_i' \beta_0) X_i \|_{\infty},\label{pt2.11}
\end{eqnarray}
where we use Holders inequality and Lemma \ref{la12}  to get the result. Note that since $E \dot{\rho} (y_i, X_i' \beta_0)=0$ due to $\beta_0$ definition,
\[ \| \frac{1}{n} \sum_{i=1}^n  \dot{\rho} (y_i, X_i' \beta_0) X_i \|_{\infty} = O_p (\sqrt{\frac{lnp}{n}}),\]
by (\ref{sa2})(\ref{sa4})(\ref{sa5}) via Assumptions \ref{as1}, \ref{as2}, and \ref{as4} which is $\dot{\rho} (y_i, X_i' \beta_0)$ being uniformly bounded. Combine this with Lemma \ref{la10}, and (\ref{h1})
\begin{equation}
 \left|  \frac{n^{1/2} \alpha' (\hat{\Theta} - \Theta) \sum_{i=1}^n \dot{\rho} (y_i, X_i' \beta_0) X_i}{n}
\right| = O_p ( h^{1/2} g_n \bar{s}^{1/2} \frac{lnp}{n^{1/2}} \max (\bar{s}, H_n^2 s_0^2)) = o_p (1),\label{pt2.12}
\end{equation}
by Assumption \ref{as7}. See that (\ref{pt2.12}) is result is uniform over $l_0$ ball ${\cal B}_{l_0} (s_0)$ since the result depends on $\beta_0$ through $s_0$ only.

{\bf Step 2}. We want to show that $tt_2=o_p (1),$ and $tt_3=o_p(1)$.  The denominators of these tests are the same as $tt_1$ in Step 1. They converge to $V_{\alpha}^2$, and $V_{\alpha}^2$ is bounded away from zero. So the denominators of $tt_2, tt_3$ are bounded away from zero wpa1.  We need to show that numerators of $tt_2, tt_3$ converge to
zero in probability.  We start with the numerator of $tt_2$. We can write that as
\[ n^{1/2} \left| \left[\alpha' [\hat{\Theta}  \hat{\Sigma}_{\hat{\beta}} - I_p)\right] (\hat{\beta} - \beta_0) 
\right| \le  \max_{j \in {\cal H}} |n^{1/2} (\hat{\Theta}_j'  \hat{\Sigma}_{\hat{\beta}} - e_j') (\hat{\beta} - \beta_0)|
\sum_{ j \in {\cal H}} | \alpha_j|
,\]
by using $\hat{\Sigma}_{\hat{\beta}}:= \frac{X_{\hat{\beta}} X_{\hat{\beta}}'}{n}= \frac{\sum_{i=1}^n \ddot{\rho} (y_i, X_i' \hat{\beta}) X_i X_i'}{n},$ and 
$e_j$ is a $p \times 1 $ vector of zeroes except $j$ th term which is one.

Now 
\begin{eqnarray*}
\max_{j \in {\cal H}}
|n^{1/2} (\hat{\Theta}_j'  \hat{\Sigma}_{\hat{\beta}} - e_j') (\hat{\beta} - \beta_0)| & \le & n^{1/2}  \max_{j \in {\cal H}} \underline{\Omega}_* ( \hat{\Theta}_j' \hat{\Sigma}_{\hat{\beta}} - e_j')
\underline{\Omega} (\hat{\beta} - \beta_0) \\
& \le & (\max_{j \in {\cal H}} \frac{\lambda_{nw}}{\hat{\tau}_j^2}) n^{1/2} \underline{\Omega} (\hat{\beta} - \beta_0) \\
& = & n^{1/2} O_p ( \sqrt{\frac{lnph}{n}}) O_p ( s_0 \sqrt{\frac{lnp}{n}})\\
& = & O_p ( \frac{s_0 lnp}{n^{1/2}}) = o_p (1),
\end{eqnarray*}
where we use Lemma \ref{la1}(i) for the first inequality, and (\ref{that3}) for the second inequality, and for the rates use Lemma \ref{la7}, Lemma \ref{la9} and  Theorem \ref{thm1}
\begin{equation}
\lambda_{nw} = O (\sqrt{\frac{lnph}{n}}) = O ( \sqrt{\frac{lnp}{n}}),\label{nwt}
\end{equation}
 by $h<p$, we have $ln ph = ln p + ln h < 2 ln p$, and by Assumption \ref{as7}. 
 By (\ref{h1}) 
 \[ n^{1/2} \left| \left[\alpha' [\hat{\Theta}  \hat{\Sigma}_{\hat{\beta}} - I_p)\right] (\hat{\beta} - \beta_0)\right| = O_p (
 \frac{ h^{1/2} s_0 ln p}{n^{1/2}}) = o_p (1),\] 
 by Assumption \ref{as7}.
 So we show that the numerator of $tt_2 = o_p (1)$. Clearly by the last result $tt_2$ numerator is asymptotically negligible is uniform over ${\cal B}_{l_0} (s_0)$. This can be seen by seeing that the last result rate depends on $\beta_0$ only through $s_0$.

\noindent Now we analyze the numerator of $tt_3$. 
\begin{eqnarray}
n^{1/2} & \times & | \alpha' \hat{\Theta} [ \frac{1}{n} \sum_{i=1}^n (\ddot{\rho} (y_i, \tilde{a}_i) - \ddot{\rho} (y_i, X_i' \hat{\beta})) X_i X_i' (\hat{\beta} - \beta_0)] |  \le  
n^{1/2} \| \alpha'  \hat{\Theta} \|_1 \nonumber \\
& \times &  \| n^{-1} \sum_{i=1}^n (\ddot{\rho} (y_i, \tilde{a}_i) - \ddot{\rho} (y_i, X_i' \hat{\beta})) X_i X_i' (\hat{\beta} - \beta_0)] \|_{\infty} \nonumber \\
& \le & n^{1/2} \| \alpha' \hat{\Theta}\|_1 \nonumber \\
& \times &   \| n^{-1} \sum_{i=1}^n (\ddot{\rho} (y_i, \tilde{a}_i) - \ddot{\rho} (y_i, X_i' \hat{\beta})) X_i X_i' \|_{\infty} \| \hat{\beta} - \beta_0 \|_1 \nonumber \\
& \le & n^{1/2} \| \alpha' \hat{\Theta} \|_1 \nonumber \\
& \times &   \| n^{-1} \sum_{i=1}^n (\ddot{\rho} (y_i, \tilde{a}_i) - \ddot{\rho} (y_i, X_i' \hat{\beta})) X_i X_i' \|_{\infty} \underline{\Omega} (\hat{\beta} - \beta_0 ),\label{pt2.13}
\end{eqnarray}
where we use Holders inequality for the first and second inequality,  and Lemma \ref{la1}(ii) for the third inequality. In (\ref{pt2.13}) we consider the middle term on the right side.
\begin{eqnarray}
\| n^{-1} \sum_{i=1}^n (\ddot{\rho} (y_i, \tilde{a}_i) - \ddot{\rho} (y_i, X_i' \hat{\beta})) X_i X_i' \|_{\infty} & \le & 
\| n^{-1} \sum_{i=1}^n X_i X_i' \|_{\infty}  \max_{1 \le i \le n} | \ddot{\rho} (y_i, \tilde{a}_i) - \ddot{\rho} (y_i, X_i' \hat{\beta})| \nonumber \\
& \le & \| n^{-1} \sum_{i=1}^n X_i X_i' \|_{\infty}  \max_{1 \le i \le n} | X_i (\hat{\beta} - \beta_0)| \nonumber \\
& \le & \| n^{-1} \sum_{i=1}^n X_i X_i' \|_{\infty}  \max_{1 \le i \le n} \underline{\Omega}_* (X_i) \underline{\Omega} (\hat{\beta} - \beta_0) \nonumber \\
& \le & \| n^{-1} \sum_{i=1}^n X_i X_i' \|_{\infty}  \max_{1 \le i \le n} \| X_i \|_{\infty}  \underline{\Omega} (\hat{\beta} - \beta_0),\label{pt2.14} 
\end{eqnarray}
where we use Lipschitz condition for second order partial derivatives-Assumption \ref{as4} for the second inequality, Lemma \ref{la2}(i) for the third inequality, and Lemma \ref{la2}(iii) for the last inequality. Incorporate (\ref{pt2.14}) into (\ref{pt2.13})
{\small \begin{eqnarray}
n^{1/2} | \alpha' \hat{\Theta} [ \frac{1}{n} \sum_{i=1}^n (\ddot{\rho} (y_i, \tilde{a}_i) - \ddot{\rho} (y_i, X_i' \hat{\beta})) X_i X_i' (\hat{\beta} - \beta_0)] | & \le &
n^{1/2} \| \alpha' \hat{\Theta} \|_1  \| n^{-1} \sum_{i=1}^n X_i X_i' \|_{\infty}  \max_{1 \le i \le n} \| X_i \|_{\infty}  [\underline{\Omega} (\hat{\beta} - \beta_0)]^2 \nonumber \\
& = & n^{1/2} O_p (h^{1/2} \bar{s}^{1/2}) O_p (1) O_p ( K_n)  O_p ( s_0^2 \frac{lnp}{n})\nonumber \\
&  = &O_p ( h^{1/2} \bar{s}^{1/2} K_n s_0^2 \frac{lnp}{n^{1/2}} ) = o_p (1),\label{pt2.15}
\end{eqnarray}}
where the rate is by (\ref{a55a}), (\ref{e3.3a})(\ref{e3.4}), by Markov's inequality-Lemma A.3 of Caner and Kock (2018), and by (\ref{asymt1}). The asymptotic negligibility is by Assumption \ref{as7}.
Note that by (\ref{pt2.15}) numerator of $tt_3$ being asymptotically negligible is uniform over $l_0$ ball ${\cal B}_{l_0} (s_0)$.
{\bf Q.E.D.} \\

\newpage

\setcounter{equation}{0}\setcounter{lemma}{0}
\renewcommand{\theequation}{B.%
\arabic{equation}}\renewcommand{\thelemma}{B.\arabic{lemma}}%
\renewcommand{\baselinestretch}{1}\baselineskip=15pt

{\bf Part B: Nodewise Regression Proofs}:\\

We define the following events in the next four Lemmata that will help us in nodewise regression proofs. 
\[ E_4:= \{ \| X (\hat{\beta} - \beta_0) \|_n^2  \max_{j \in {\cal H}} \| \eta_{\beta_0,j} \|_{\infty}^2 \le C H_n^2 s_0^2 \lambda^2 (1+t_1)\}
,\]
with $t_1$ is defined as in (\ref{e3.4})(\ref{e3.4a}). Let $N_1>0$ be a sufficiently large positive constant. $H_n$ is defined in Assumption \ref{as5}.

\begin{lemma}\label{la5}

Under Assumptions \ref{as1}-\ref{as5}, $E_4$ holds with probability at least  
$1 - \frac{3}{p^{2c}} - \frac{1}{p^c} - \frac{7 C}{4(ln p)^2} - \frac{C}{N_1} =1 - o(1).$ So, since $t_1=o(1)$, uniformly over ${\cal B}_{l_0} (s_0)$

\begin{equation}
\max_{j \in {\cal H}} \| X (\hat{\beta} - \beta_0) \|_n^2 \| \eta_{\beta_0,j} \|_{\infty}^2 = O_p (H_n^2 s_0^2 \lambda^2).\label{r1e4}
\end{equation}

\end{lemma}

{\bf Proof of Lemma \ref{la5}}.

First by Assumption \ref{as5} we have 
\begin{equation}
\max_{j \in {\cal H}} \| \eta_{\beta_0,j} \|_{\infty}^2 = O_p ( H_n^2).\label{eta}
\end{equation} 
To see this by Markov's inequality and  by Lemma A.3 of Caner and Kock (2018)
\begin{equation}
P [ \max_{1 \le i \le n } \max_{ j \in {\cal H}} | \eta_{\beta_0,j,i}| > t] \le \frac{C h n }{t^{r/2}} = \frac{C}{N_1},\label{b.1a}
\end{equation}
with $t=N_1 h^{2/r} n^{2/r}$, and by  $H_n := O (h^{2/r} n^{2/r})$.

\noindent Then 
\begin{eqnarray*}
\| X (\hat{\beta} - \beta_0) \|_n^2 & = & (\hat{\beta} - \beta_0)' (\hat{\Sigma} - \Sigma) (\hat{\beta} - \beta_0)\nonumber \\
& + & (\hat{\beta} - \beta_0)' \Sigma (\hat{\beta} - \beta_0).
\end{eqnarray*} 

\noindent Analyze the first term on the right side above, by using (\ref{e3.3})(\ref{e3.3a}), (\ref{asymt1}), with $C$ being a positive constant
\begin{eqnarray}
|(\hat{\beta} - \beta_0)' (\hat{\Sigma} - \Sigma) (\hat{\beta} - \beta_0)|
& \le & [\underline{\Omega} (\hat{\beta} - \beta_0)]^2 \| \hat{\Sigma} - \Sigma \|_{\infty} \nonumber \\
& \le & C \lambda^2 s_0^2 t_1,\label{la5.1}
\end{eqnarray}
with probability at least $1- \frac{3}{p^{2c}} - \frac{1}{p^c} - \frac{ 7 C}{ 4(lnp)^2}$, since $\| \hat{\Sigma} - \Sigma \|_{\infty}$ proof is a subset of the proof of 
$\underline{\Omega} (\hat{\beta} - \beta_0)$. Next by  (\ref{e3.3}), Theorem \ref{thm1}-Remark 2 by (\ref{asymt1}), Assumption \ref{as1} with Cauchy-Schwartz inequality
\begin{equation}
(\hat{\beta} - \beta_0)' \Sigma (\hat{\beta} - \beta_0) \le [\underline{\Omega} (\hat{\beta} - \beta_0)]^2 \| \Sigma \|_{\infty} \le 
C \lambda^2 s_0^2,\label{la5.2}
\end{equation}
with probability at least $1 - \frac{3}{p^{2c}} - \frac{1}{p^c} - \frac{  7 C }{4  (lnp)^2}$. Combine (\ref{la5.1})(\ref{la5.2}) to have the result.

 Note that we have $t_1= O( \sqrt{lnp/n}) = o(1)$ established in Appendix before this result in (\ref{e3.4}), Assumption \ref{as2}. So the term with $t_1$ converges to zero faster, hence 
\begin{equation}
 \| X (\hat{\beta} - \beta_0) \|_n^2 = O_p (s_0^2 \lambda^2).\label{oi}
 \end{equation}
By (\ref{eta}), the result at (\ref{r1e4}) is obtained. The tail probability in $E_4$ is obtained by adding (\ref{b.1a}), and the probability attached to (\ref{la5.1}).  (\ref{la5.1}) probability implies (\ref{la5.2}). Uniformity can be seen since the end result depends on $\beta_0$ only through $s_0$.
{\bf Q.E.D.}


The empirical version of the compatibility condition is:
\begin{equation}
 \hat{\phi}^2 (L, S_j):= \min \{ | S_j | \| X_{\hat{\beta}, -j} \gamma_{\beta_0,S_j} - X_{\hat{\beta}, -j} \gamma_{\beta_0,S_j^c} \|_n^2:  \Omega (\gamma_{\beta_0,S_j}) =1, \Omega^{S^c} (\gamma_{\beta_0,S_j^c}) \le L \}
.\label{evl2}
\end{equation}

\begin{lemma}\label{la6}

 With Assumptions  \ref{as1} with $r_x > 8$, \ref{as2}-\ref{as5},\ref{as6}(ii), with probability approaching at least $1 - \frac{8}{p^{2c} }- \frac{2}{p^c} - \frac{2C}{N_1} - \frac{ 15  C}{ 4(ln p)^2} $ for each $j=1,\cdots, p$, for sufficiently large $n$,
\[ E_{5,j}= \left\{ \hat{\phi}^2 (L, S_j) \ge \phi^2 (L, S_j)/2 \right\}.\]

Note that the statement holds with probability approaching one, when $n \to \infty$, and since $N_1>0$ is a large constant. The results are uniform over $l_0$ ball ${\cal B}_{l_0} (s_0)$.
\end{lemma}

{\bf Proof of Lemma \ref{la6}}. First we start with definitions, set $\hat{\Sigma}_{\hat{\beta}, -j}:= X_{\hat{\beta}, -j}' X_{\hat{\beta}, -j}/n, \hat{\Sigma}_{\hat{\beta}}:= X_{\hat{\beta}}' X_{\hat{\beta}}/n$. Then their population counterparts are defined: $\Sigma_{\hat{\beta}, -j}:= E [ X_{\hat{\beta}, -j}' X_{\hat{\beta}, -j}/n], \Sigma_{\hat{\beta}}:=  E [X_{\hat{\beta}}' X_{\hat{\beta}}/n]$. Define also $\Sigma_{\beta_0}:= E X_{\beta_0}' X_{\beta_0}/n$, and $\Sigma_{\beta_0, -j}: E X_{\beta_0,-j}' X_{\beta_0, -j}/n$ where $X_{\beta_0,-j}: n \times p-1$ matrix where its a subset of  $X_{\beta_0}:n \times p$ matrix (where $X_{\beta_0,-j}$ is $j$ th column of $X_{\beta_0}$ is deleted). 
Use the analysis in (\ref{ev06}), 


\begin{equation}
\frac{  | \gamma_{\beta_0,j}' \hat{\Sigma}_{\hat{\beta},-j} \gamma_{\beta_0,j} |}{[\Omega (\gamma_{\beta_0,S_j})]^2} \ge 
\frac{| \gamma_{\beta_0,j}' \Sigma_{\beta_0,-j} \gamma_{\beta_0,j}|}{[\Omega (\gamma_{\beta_0,S_j})]^2} - (L+1)^2  \| \hat{\Sigma}_{\hat{\beta}, -j} - \Sigma_{\beta_0,-j} \|_{\infty}.\label{pla61}
\end{equation}

\noindent The key is bounding the second term on the right side of (\ref{pla61})

\begin{eqnarray}
\| \hat{\Sigma}_{\hat{\beta}, -j} - \Sigma_{\beta_0,-j} \|_{\infty} & \le &  \| \hat{\Sigma}_{\hat{\beta}} - \Sigma_{\beta_0} \|_{\infty}  \nonumber \\
& \le & \| \hat{\Sigma}_{\hat{\beta}} - \Sigma_{\hat{\beta}} \|_{\infty}  + \| \Sigma_{\hat{\beta}} - \Sigma_{\beta_0} \|_{\infty}.\label{pla62}
 \end{eqnarray}
First consider by definition of $\hat{\Sigma}_{\hat{\beta}}, \Sigma_{\hat{\beta}}$
\begin{eqnarray}
\| \hat{\Sigma}_{\hat{\beta}} - \Sigma_{\hat{\beta}} \|_{\infty} & := & \| \frac{1}{n} \sum_{i=1}^n (X_i X_i' - E X_i X_i') w_{\hat{\beta},i}^2 \|_{\infty} \nonumber \\
& \le & \| \frac{1}{n} \sum_{i=1}^n (X_i X_i' - E X_i X_i')\|_{\infty} [\max_{1 \le i \le n} w_{\hat{\beta},i}^2].\label{pla63}
\end{eqnarray}
Next by the definition of the estimated weight $w_{\hat{\beta},i}:= \sqrt{ \ddot{\rho} (y_i, X_i' \hat{\beta})}$, with probability $1 - 3/p^{2c} - 1/p^c - C/N_1 - 7C/4(lnp)^2$
\begin{equation}
\max_{1 \le i \le n} w_{\hat{\beta},i}^2 \le C < \infty,\label{b.8a}
\end{equation}
by Assumption \ref{as4}, Lemma \ref{la1}(i), and Theorem \ref{thm1} (in detail the proof can be seen by (\ref{pla66}) below and simple triangle inequality). So with probability at least $1 - 4/p^{2c}- 1/p^c-C/N_1-8C/4 (ln p)^2$ 
\begin{equation}
\| \hat{\Sigma}_{\hat{\beta}} - \Sigma_{\hat{\beta}} \|_{\infty} \le C t_1,\label{pla64}
\end{equation}
as in (\ref{e3.3a})(\ref{e3.4}). In (\ref{pla62}) consider the second term on the right side, by definition 
\begin{eqnarray}
\| \Sigma_{\hat{\beta}} - \Sigma_{\beta_0} \|_{\infty}&:= & \| \frac{1}{n} \sum_{i=1}^n E X_i X_i' (w_{\hat{\beta},i}^2 - w_{\beta_0,i}^2) \|_{\infty} \nonumber \\
& \le & \| \frac{1}{n} \sum_{i=1}^n E X_i X_i' \|_{\infty} [ \max_{1 \le i \le n} (w_{\hat{\beta},i}^2 - w_{\beta_0,i}^2)].\label{pla65}
\end{eqnarray}
In (\ref{pla65}) take the second term on the right side
\begin{eqnarray}
\max_{1 \le i \le n} (w_{\hat{\beta},i}^2 - w_{\beta_0,i}^2) & \le & 
 \max_{1 \le i \le n } | X_i' (\hat{\beta} - \beta_0)| \nonumber \\
& \le &  [\max_{1 \le i \le n} \underline{\Omega}_* (X_i) ] [ \underline{\Omega} (\hat{\beta} - \beta_0)] \nonumber \\
& \le &  [ \max_{1 \le i \le n} \| X_i \|_{\infty}] [ \underline{\Omega} (\hat{\beta} - \beta_0)] \nonumber \\
& \le &  K_n (\frac{9 \lambda s_0}{ \phi^2 (2,S_0) - 9 t_1 s_0}),\label{pla66}
\end{eqnarray}
where we use Assumption \ref{as4}(iii), Lipschitz continuity of the weights, and Theorem 1 for the second inequality, and Lemma \ref{la1}(i) for the third inequality, Lemma \ref{la1}(iii) for the fourth inequality, and Lemma A.3 in Caner and Kock (2018) and Theorem \ref{thm1} for the last inequality with probability at least $1 - \frac{3}{p^{2c}} - \frac{1}{p^c} - \frac{C}{N_1} - 
\frac{7 C }{4 (ln p)^2}$, where $N_1$ is a sufficiently large positive constant.  Combining all the results (\ref{pla64})(\ref{pla66})

\begin{equation}
\frac{  | \gamma_{\beta_0,j}' \hat{\Sigma}_{\hat{\beta},-j} \gamma_{\beta_0,j} |}{\Omega (\gamma_{\beta_0,S_j})^2} \ge 
\frac{| \gamma_{\beta_0,j}' \Sigma_{\beta_0,-j} \gamma_{\beta_0,j}|}{\Omega (\gamma_{\beta_0,S_j})^2} - (L+1)^2 \left[ C t_1+ \frac{ 9 C K_n \lambda s_0 }{\phi^2(2, S_0)
- 9 t_1 s_0} 
\right],\label{pla67}
\end{equation}
holding with probability at least $1 - \frac{8}{p^{2c} }- \frac{7}{p^c} - \frac{2C}{N_1} - \frac{ 15 C }{4 (ln p)^2}$.
 Then multiply each side of the inequality with cardinality $|S_j|$, 
 and then take a minimum of each side to have, and note that the equivalence of the compatibility definition just before this Lemma and (\ref{pla67}) based formula from Lemma 4.1 of van de Geer (2014)
 \begin{equation}
 \hat{\phi}^2 (L, S_j) = \{ \min_{\gamma_{\beta_0,j}} \frac{ | \gamma_{\beta_0,j}' \hat{\Sigma}_{\hat{\beta}, -j} \gamma_{\beta_0,j}|}{\Omega(\gamma_{\beta_0,  S_j})^2}: {\mbox subject \, to} \quad \Omega^{S^c} (\gamma_{\beta_0, S_j^c}) \le L \Omega (\gamma_{\beta_0, S_j}) \}.\label{eqcc1}
 \end{equation}
\noindent and
 \[ \phi^2 (L, S_j) = \{ \min_{\gamma_{\beta_0,j}} \frac{ | \gamma_{\beta_0,j}' \Sigma_{\beta_0, -j} \gamma_{\beta_0,j}|}{\Omega(\gamma_{\beta_0,  S_j})^2}: {\mbox subject \, to} \quad \Omega^{S^c} (\gamma_{\beta_0, S_j^c}) \le L \Omega (\gamma_{\beta_0, S_j}) \}.\]

 \noindent  to have  
 
\[ \hat{\phi}^2 (L, S_j) \ge \phi^2 (L, S_j) - (L+1)^2 | S_j |  \left[ C t_1+ \frac{ 9 C K_n \lambda s_0 }{\phi^2(2, S_0)
- 9 t_1 s_0} 
\right] .\]

 Now we simplify the expression above. We want to show that, for $j=1,\cdots, p$ 
 \[(L +1)^2  s_j [ C t_1 + \frac{9 C K_n \lambda s_0}{\phi^2 ( 2, S_0) - 9 t_1 s_0}]
 \le \phi^2 (L, S_j)/2.\]
 
 Note that for each $j=1,\cdots, p$ $\phi^2 (L, S_j) \ge c > 0$ by Assumption \ref{as5}.
 First start with the denominator term $t_1 s_0 = o(1)$ by Assumption \ref{as2}(ii), with $t_1= O ( \sqrt{lnp/n})$. So with sufficiently large $n$
 \begin{equation}
  \phi^2 (2, S_0) - 9 t_1 s_0 = \phi^2 (2, S_0) - o(1) \ge c - o(1) > 0,\label{pla6aa}
  \end{equation}
 by Assumption 1 with explanation in (\ref{42a}) showing that effective sparsity is tied to the inverse of the compatibility constant. Then in the numerator 
 by Assumption \ref{as6}(ii), since $|S_j|= s_j \le \bar{s}$, $\bar{s}s_0 K_n \sqrt{lnp/n} = o(1)$, also in the same way $\bar{s} t_1 = o(1)$. Then combining all above 
 \[(L +1)^2  \bar{s} [ C t_1 + \frac{9 C K_n \lambda s_0}{\phi^2 ( 2, S_0) - 9 t_1 s_0}] = o(1)
 \le \phi^2 (L, S_j)/2.\] 
 Uniformity over $l_0$ ball ${\cal B}_{l_0} (s_0)$ is clear since we use Assumption \ref{as5}. To see this point more clearly, the first term on the right side of (\ref{pla67}) is uniform over $l_0$ ball ${\cal B}_{l_0} (s_0)$ by Assumption \ref{as5}(i), then the second term on the right side of (\ref{pla67}) depends on $\beta_0$ only through $s_0$.
 {\bf Q.E.D.}

Define $E_5:= \{ \cap_{j \in H} E_{5,j} \}$. By (\ref{pla62})-(\ref{pla67}), 
\[ \left\{  \| \hat{\Sigma}_{\hat{\beta},-j} - \Sigma_{\beta_0, -j} \|_{\infty} \le 
\left[ C t_1 + \frac{K_n \lambda s_0}{\phi^2 (2, S_0) - 9 t_1 s_0}
\right]  \right\} \subseteq E_{5,j}.\]
By (\ref{pla62}), and then (\ref{pla65})(\ref{pla66})
$ \| \hat{\Sigma}_{\hat{\beta},-j} - \Sigma_{\beta_0, -j} \|_{\infty} \le \| \hat{\Sigma}_{\hat{\beta}} - \Sigma_{\beta_0} \|_{\infty}$, so 
\[ \left\{ \| \hat{\Sigma}_{\hat{\beta}} - \Sigma_{\beta_0} \|_{\infty} \le 
\left[ C t_1 + \frac{K_n \lambda s_0}{\phi^2 (2, S_0) - 9 t_1 s_0}
\right]  \right\} \subseteq E_{5,j}.\]
So 
\[ \left\{ \| \hat{\Sigma}_{\hat{\beta}} - \Sigma_{\beta_0} \|_{\infty} \le 
\left[ C t_1 + \frac{K_n \lambda s_0}{\phi^2 (2, S_0) - 9 t_1 s_0}
\right]  \right\} \subseteq \cap_{j \in {\cal H}}E_{5,j}:= E_5.\]
This implies 
\begin{eqnarray}
P ( \{E_5 \})^c& :=&  P ( \{ \cap_{j \in {\cal H }} E_{5,j} \}^c) \nonumber \\
& \le & P \left[ \| \hat{\Sigma}_{\hat{\beta}} - \Sigma_{\beta_0} \|_{\infty} \le 
\left[ C t_1 + \frac{K_n \lambda s_0}{\phi^2 (2, S_0) - 9 t_1 s_0}
\right]  \right] \nonumber \\
& \le & \frac{8}{p^{2c}} + \frac{2}{p^c} + \frac{2C}{N_1} + \frac{ 15 C }{4 (ln p)^2} \nonumber \\
& \to &  0.
\label{pla68} 
\end{eqnarray}

Now we want to prove that following events hold with probability approaching one. To that effect, note that $X_{\beta_0,j}$ is $W_{\beta_0} X_{j}$, where $W_{\beta_0}$ is a diagonal matrix described in the main text, $X_j$ is the $j$ th column of the $X$ matrix, $j=1,2,\cdots, p$.  Also define $X_{\beta_0, -j}$ which is $n \times p-1$ matrix, which is $X_{\beta_0}$ without $j$ th column. See that $| S_j | \cup | S_j^c | = p-1$.

Note that $X_{\beta_0, S_j}$ and $X_{\beta_0, S_j^c}$ are the regressors of $X_{\beta_0, -j}$  that correspond to allowed set $S_j$ (see Definition \ref{def2}) and its complement $S_j^c$ respectively. In other words, we can write $X_{\beta_0, S_j} \cup X_{\beta_0, S_j^c}  = X_{\beta_0, -j}$.
We start with a condition that provides our results. Since $S_j, S_j^c $ is a subset of $\{ 1, 2,\cdots, p-1\}$,  
\[ \max_{j \in {\cal H}} \| X_{\beta_0, S_j}' \eta_{\beta_0,j}/n \|_{\infty} \le \max_{j \in {\cal H}}\| X_{\beta_0,-j}' \eta_{\beta_0,j}/n \|_{\infty} \le \lambda_1 .\]
\[ \max_{j \in {\cal H}}
\| X_{\beta_0, S_j^c}' \eta_{\beta_0,j}/n \|_{\infty} \le \max_{j \in {\cal H}}
\| X_{\beta_0,-j}' \eta_{\beta_0,j}/n \|_{\infty} \le \lambda_2 .\]
 So if we prove that  with wpa1
\[ \max_{j \in {\cal H}} \| X_{\beta_0,-j}' \eta_{\beta_0,j}/n \|_{\infty} \le \lambda_1,\]
and similarly with wpa1
\[ \max_{j \in {\cal H}}
\| X_{\beta_0,-j}' \eta_{\beta_0,j}/n \|_{\infty} \le \lambda_2,\]
will imply that prove our events below hold with wpa1.

Events are:

\[ E_6:= \{  \max_{j \in {\cal H}} \| X_{\beta_0, S_j}' \eta_{\beta_0,j}/n \|_{\infty} \le \lambda_1\},\]
\[ E_7: = \{ \max_{j \in {\cal H}}  \| X_{\beta_0, S_j^c}' \eta_{\beta_0,j}/n \|_{\infty} \le \lambda_2\}.\]
Without losing any generality in our asymptotic results set $\lambda_{nw}: = 2 \lambda_2, \lambda_1:= \lambda_2$.
Specifically define  for $l=1,2$
\begin{equation}
\lambda_l := 2 K  [ \sqrt{\frac{lnph}{n}} + \frac{\sqrt{ E M_3^2} ln ph}{n}] + \sqrt{\frac{lnph}{n}}.\label{lam}
\end{equation}

\begin{lemma}\label{la7}

Under Assumptions \ref{as1},\ref{as2}, \ref{as4},\ref{as5} we have 
the events $E_6, E_7$ each of them holding with probability at least $1- \frac{1}{(ph)^c} - \frac{C }{ (ln ph)^2}= 1 - o(1)$, and 
$\lambda_l = O (\sqrt{\frac{lnph}{n}})=O (\sqrt{\frac{lnp}{n}})$, as well as $\lambda_{nw}= O (\sqrt{\frac{lnp}{n}})$ since $h < p$.

\end{lemma}

{\bf Proof of Lemma \ref{la7}}. 
We start with  for each $i=1,\cdots, n$ using iid nature of data in Assumption \ref{as1}, and zero mean  of $E X_{\beta_0,j}' \eta_{\beta_0,j}=0$ by (\ref{nw1a}) (since we use nodewise regression which can be proved though matrix partition/inversion formulas as in Caner and Kock (2018)). Denote $X_{\beta_0, -j,ik}$ as the $i$th row  and $k$ th column element  of $X_{\beta_0,-j}$ matrix which is of $n \times p-1$ dimension, and $\eta_{\beta_0,j,i}$ as the $i$ th element of the $n \times 1 $ vector $\eta_{\beta_0,j}$, for each $j \in {\cal H}$

\begin{eqnarray*}
 \max_{1 \le k \le p-1} E [ X_{\beta_0, -j,ik} \eta_{\beta_0,ij}]^2 & \le & [\max_{1 \le j \le p-1} E | X_{\beta_0, -j,ik}|^4]^{1/2}
 [ E [ | \eta_{\beta_0,j,i}|^4]]^{1/2} \nonumber \\
 & \le & C[ \max_{1 \le k \le p-1} E  | X_{\beta_0,-j,ik}|^4]^{1/2} \le C < \infty,
\end{eqnarray*}
where we use Cauchy-Schwartz inequality for the first inequality, and then Assumption \ref{as5} for the second inequality and then Assumption \ref{as1} and \ref{as4}(ii) (weights $w_{\beta_0,i}$ being uniformly bounded away from infinity) for the last inequality. Now we can apply (\ref{sa2})(\ref{sa4})(\ref{sa5})
to get the result with $\lambda_1$ definition, same for the $\lambda_2$ result since they are the same number, and conditions before lemma provides the result. Last result is by (\ref{nwt}).{\bf Q.E.D.}

We define 
\begin{equation}
E_8:= { \sup_{\beta \in {\cal B}_{l_0} (s_0)} \max_{1 \le i \le n} \left|  \frac{\ddot{\rho} (y_i, X_i' \beta_0)}{\ddot{\rho} (y_i, X_i' \hat{\beta})}
\right| \le 1 }.\label{e8}
\end{equation}

Next Lemma is used in the proof of nodewise regression oracle inequality. 

\begin{lemma}\label{la7a}
Under Assumptions \ref{as1}-\ref{as4}, and \ref{as6}(ii), with sufficiently large n
\[ P (E_8) \ge 1 - 3/p^{2c} - 1/p^c- C/N_1 - 7C/4(lnp)^2.\]
\end{lemma}

{\bf Proof of Lemma \ref{la7a}}.
\[ \max_{1 \le i \le n} \left| \frac{\ddot{\rho} (y_i, X_i' \beta_0)}{\ddot{\rho} (y_i, X_i' \hat{\beta})}
\right| 
\le \max_{1 \le i \le n} \left| \frac{\ddot{\rho} (y_i, X_i' \beta_0)}{\ddot{\rho} (y_i, X_i' \beta_0) - | \ddot{\rho} (y_i, X_i' \hat{\beta}) -
\ddot{\rho} (y_i, X_i' \beta_0)|}
\right|.\]

Then by (\ref{pla66}), via Lipschitz continuity in Assumption \ref{as4}(iii), with Lemma \ref{la1}
\[ | \ddot{\rho} (y_i, X_i' \hat{\beta}) -
\ddot{\rho} (y_i, X_i' \beta_0)| \le \frac{9 K_n \lambda s_0}{\phi^2 (2, S_0) - 9 t_1 s_0},\]
with probability at least $1 - 3/p^{2c} - 1/p^c- C/N_1 - 7C/4(lnp)^2$.
With sufficiently large $n$, with (\ref{pla6aa}) and Assumption \ref{as6}(ii)
\[  \frac{9 K_n \lambda s_0}{\phi^2 (2, S_0) - 9 t_1 s_0} \to 0.\]
We have the desired result uniformly over ${\cal B}_{l_0} (s_0)$ by combining the last three equations.{\bf Q.E.D.}

We provide two inequalities here before the following lemma. The first is called the triangle property, and the second one is related to compatibility condition. They are independently provided. Compatibility condition is provided by van de Geer (2014). Note that our norm, $\underline{\Omega}(.)$ is decomposable by definition in (\ref{unorm}), and  hence weakly decomposable (since it is generated from cones: Section 6.9, van de Geer (2016)).
Then by definition of the decomposable norm in section 2 which is in (\ref{unorm}):
\begin{equation}
 \underline{\Omega} (\gamma_{\beta_0,j}) :=  \Omega (\gamma_{\beta_0,S_j}) + \Omega^{S^c} (\gamma_{\beta_0,S_j^c}).\label{nw12a}
 \end{equation}
The same definition above holds for $\hat{\gamma}_{\hat{\beta},j}$ as well.
\begin{equation}
\underline{\Omega} (\hat{\gamma}_{\hat{\beta}_j}):= \Omega (\hat{\gamma}_{\hat{\beta}, S_j}) + \Omega^{S^c} (\hat{\gamma}_{\hat{\beta}, S_j^c}),\label{nw12b}
\end{equation}
Hence $S_j$ is an allowed set. Then we have 
\begin{equation}
\underline{\Omega} (\gamma_{\beta_0,j}) - \underline{\Omega} (\hat{\gamma}_{\hat{\beta},j}) \le 
\Omega (\hat{\gamma}_{\hat{\beta}, S_j} - \gamma_{\beta_0, S_j}) + \Omega^{S^c} (\gamma_{\beta_0, S_j^c}) 
- \Omega^{S^c} ( \hat{\gamma}_{\hat{\beta}_{S_j^c}}).\label{nw12} 
\end{equation}  

To get (\ref{nw12}) above, subtract (\ref{nw12b}) from (\ref{nw12a}) 

\begin{eqnarray*}
\underline{\Omega} (\gamma_{\beta_0, j}) - \underline{\Omega} (\hat{\gamma}_{\hat{\beta},j}) & = & 
\Omega (\gamma_{\beta_0,S_j}) - \Omega (\hat{\gamma}_{\hat{\beta}, S_j}) \\
& + & \Omega^{S^c} (\gamma_{\beta_0, S_j^c}) - \Omega^{S^c} ( \hat{\gamma}_{\hat{\beta}, S_j^c}) \\
& \le & \Omega (\hat{\gamma}_{\hat{\beta}, S_j} - \gamma_{\beta_0, S_j}) + \Omega^{S^c} (\gamma_{\beta_0, S_j^c}) - \Omega^{S^c} ( \hat{\gamma}_{\hat{\beta}, S_j^c}),
\end{eqnarray*}
by reverse triangle inequality  to get the inequality above. This is $\underline{\Omega}$ (a decomposable norm) version of triangle property in section 6.4 of van de Geer (2016).  Our triangle property (\ref{nw12}), is new and we use a weaker norm $\underline{\Omega}$ on the left side, rather than $\Omega$ which exists in van de Geer (2016).  This new result is necessary for proof of the next lemma, existing triangle property results are not helpful due to their usage of stronger norm $\Omega$.

Next we provide an inequality related to compatibility condition. Lemma 4.1 of van de Geer (2014), simply modifying the proof for compatibility condition, shows that if the following
cone condition holds (see (\ref{eqcc1}))
\begin{equation}
\Omega^{S_j^c} (\hat{\gamma}_{\hat{\beta}, S_j^c}  - \gamma_{\beta_0, S_j^c}) \le L 
\Omega (\hat{\gamma}_{\hat{\beta}, S_j} - \gamma_{\beta_0, S_j}).\label{cc1}
\end{equation}
then the following inequality holds:
\begin{equation}
\Omega (\hat{\gamma}_{\hat{\beta}, S_j} - \gamma_{\beta_0, S_j}) \le 
\sqrt{|S_j|} \frac{ \| X_{\hat{\beta}, -j} ( \hat{\gamma}_{\hat{\beta}, j } - \gamma_{\beta_0, j}) \|_n }{\hat{\phi}_{\Omega} (L,S_j)}.\label{nw13}
\end{equation}

We provide one of the main results in our paper. This next result may be useful in other contexts as well.

\begin{lemma}\label{la8}

(i). For each $j \in {\cal H}$, under Assumptions \ref{as1} with $r_x>8$ and \ref{as2}-\ref{as5}, \ref{as6}(ii)
\[ \underline{\Omega} ( \hat{\gamma}_{\hat{\beta},j} - \gamma_{\beta_0, j}) \le 
\frac{16 \lambda_{nw} |S_j|}{\phi_{\Omega}^2 (L, S_j)}
+ 32 \Omega^{S^c} (\gamma_{\beta_0,S_j^c}) + 8 C H_n^2 \frac{\lambda^2}{
\lambda_{nw}} s_0^2 (1+t_1),\]
with probability at least $ 1 - \frac{14}{p^{2c}} - \frac{4}{p^c} - 
\frac{4C}{N_1} - \frac{C}{ln (ph)^2} - 
\frac{ 29 C }{ 4(ln p)^2} - \frac{1}{(ph)^c}= 1 - o(1)$.

(ii). Adding Assumption \ref{as6}(i) to Assumptions above in (i) and with  partial sparsity condition $ \sup_{\beta_0 \in {\cal B}_{l_0} (s_0)} \max_{j \in {\cal H}} \Omega^{S^c} (\gamma_{\beta_0, S_j^c}) = o(d_n)=o(1)$, we have 
\[ \max_{j \in{\cal H}} \underline{\Omega} ( \hat{\gamma}_{\hat{\beta},j} - \gamma_{\beta_0, j})
= O_p \left( \sqrt{\frac{lnp}{n}} max (\bar{s}, H_n^2 s_0^2) \right) = o_p (1).\]

The  result in (ii) is uniform over $l_0$ ball ${\cal B}_{l_0} (s_0)$. The rate  $d_n$ is defined in (\ref{eqdn}).

\end{lemma}

{\bf Proof of Lemma \ref{la8}}. 

(i). We start the proof under Events $E_4, E_5, E_6, E_7, E_8$ then at the very end of proof of (i) here, we relax these using Lemmata \ref{la5}-\ref{la7a} above.

The proof will be in two parts. 
Our proof extends Theorem 6.1 of van de Geer (2016) considerably. Theorem 6.1 of van de Geer (2016) has linear least-squares loss with a strong $\Omega$ norm. Our proof uses GLM loss with weaker $\underline{\Omega}$. To do so, we benefit from triangle property for $\underline{\Omega}$ above, in (\ref{nw12}). Also, we prove all high-level conditions in our lemma rather than assuming them.

{\bf Part 1.} Given (\ref{nw3}) we can use Lemma 6.1 of van de Geer (2016), two-point inequality, which is derived from a simple algebraic minimization 
\begin{equation}
(\gamma_{\beta_0,j} - \hat{\gamma}_{\hat{\beta},j})' X_{\hat{\beta},-j}' (X_{\hat{\beta},j} - X_{\hat{\beta},-j} \hat{\gamma}_{\hat{\beta},j})/n
\le \lambda_{nw} \underline{\Omega} (\gamma_{\beta_0,j}) - \lambda_{nw} \underline{\Omega} (\hat{\gamma}_{\hat{\beta},j}).\label{nw4}
\end{equation}

\noindent We need to simplify (\ref{nw4}) so that we can use in the following proofs. Impose (\ref{nw2}) in (\ref{nw4})
\begin{equation}
\frac{ (\hat{\gamma}_{\hat{\beta},j} - \gamma_{\beta_0,j})' X_{\hat{\beta},-j}' X_{\hat{\beta},-j}  (\hat{\gamma}_{\hat{\beta},j} - \gamma_{\beta_0,j})}{n}
\le \frac{ (W_{\hat{\beta}} W_{\beta_0}^{-1} \eta_{\beta_0,j})' X_{\hat{\beta}, -j}  (\hat{\gamma}_{\hat{\beta},j} - \gamma_{\beta_0,j})}{n}
+ \lambda_{nw} \underline{\Omega} (\gamma_{\beta_0,j}) - \lambda_{nw} \underline{\Omega} (\hat{\gamma}_{\hat{\beta},j}).\label{nw5}
\end{equation}

We add and subtract $(\eta_{\beta_0,j}' X_{\beta_0,-j}/n) (\hat{\gamma}_{\hat{\beta},j} - \gamma_{\beta_0,j})$ from the right side of (\ref{nw5}) above
\begin{eqnarray}
\frac{ (\hat{\gamma}_{\hat{\beta},j} - \gamma_{\beta_0,j})' X_{\hat{\beta},-j}' X_{\hat{\beta},-j}  (\hat{\gamma}_{\hat{\beta},j} - \gamma_{\beta_0,j})}{n}
&\le &  \frac{ \{ [W_{\hat{\beta}} W_{\beta_0}^{-1} \eta_{\beta_0,j}]' X_{\hat{\beta}, -j} - (\eta_{\beta_0,j}' X_{\beta_0,j})\}  (\hat{\gamma}_{\hat{\beta},j} - \gamma_{\beta_0,j})}{n} \nonumber \\
& + & \frac{ (\eta_{\beta_0,j}' X_{\beta_0,j})  (\hat{\gamma}_{\hat{\beta},j} - \gamma_{\beta_0,j})}{n} \nonumber \\
&+& [\lambda_{nw} \underline{\Omega} (\gamma_{\beta_0,j}) - \lambda_{nw} \underline{\Omega} (\hat{\gamma}_{\hat{\beta},j})].\label{nw6}
\end{eqnarray}
Then apply Cauchy-Schwartz inequality to the first term on the right side of (\ref{nw6}), and using the definition $ X_{\hat{\beta}, -j}:= W_{\hat{\beta}} 
W_{\beta_0}^{-1} X_{\beta_0,-j}$ we have 
\begin{equation}
\frac{ \{ [W_{\hat{\beta}} W_{\beta_0}^{-1} \eta_{\beta_0,j}]' X_{\hat{\beta}, -j} - (\eta_{\beta_0,j}' X_{\beta_0,j})\}  (\hat{\gamma}_{\hat{\beta},j} - \gamma_{\beta_0,j})}{n} \le 
[\| (W_{\hat{\beta}}^2 W_{\beta_0}^{-2} - I_n) \eta_{\beta_0,j} \|_n 
\| X_{\beta_0,-j} ( \hat{\gamma}_{\hat{\beta},j}
- \gamma_{\beta_0,j} \|_n].\label{nw7}
\end{equation}
Consider 
\begin{eqnarray}
 \| (W_{\hat{\beta}}^2 W_{\beta_0}^{-2} - I_n) \eta_{\beta_0,j} \|_n^2 &\le& \frac{1}{n}  \sum_{i=1}^n \left( \frac{\hat{w}_{i,\hat{\beta}}^2 - w_{i, \beta_0}^2}{w_{i,\beta_0}^2}
\right)^2  \max_{ j \in {\cal H}}
\| \eta_{\beta_0,j}\|_{\infty}^2 \nonumber \\
& \le & \| X (\hat{\beta} - \beta_0) \|_n^2  \max_{j \in {\cal H}} \| \eta_{\beta_0,j}\|_{\infty}^2 \nonumber \\
& \le & C H_n^2 \lambda^2 s_0^2 (1+ t_1) ,\label{nw8}
\end{eqnarray}
where we use Assumption \ref{as4}, Lipschitz condition in the second inequality, and the rest is the event definition, $E_4 =\{ \max_{j \in {\cal H}} (\| X (\hat{\beta} - \beta_0) \|_n^2 )\| \eta_{\beta_0,j}\|_{\infty}^2 \le C H_n^2 s_0^2 \lambda^2 (1+t_1) \}
$.  
Set  $a= \| X_{\beta_0, -j} (\hat{\gamma}_{\hat{\beta},j} - \gamma_{\beta_0,j}) \|_n$, and $ b= \| (\hat{W}_{\hat{\beta}}^2  W_{\beta_0}^{-2}
- I_n ) \eta_{\beta_0,j} \|_n$. Then by $ab \le a^2/2+ b^2/2$ with (\ref{nw7})
\begin{eqnarray}
\| (W_{\hat{\beta}}^2 W_{\beta_0}^{-2} - I_n) \eta_{\beta_0,j} \|_n & \times &  \| X_{\beta_0,-j} ( \hat{\gamma}_{\hat{\beta},j}
- \gamma_{\beta_0,j}) \|_n\nonumber \\
&\le & 
\| (W_{\hat{\beta}}^2 W_{\beta_0}^{-2} - I_n) \eta_{\beta_0,j} \|_n^2/2+ \| X_{\beta_0,-j} ( \hat{\gamma}_{\hat{\beta},j}
- \gamma_{\beta_0,j} \|_n^2/2.\label{nw9}
\end{eqnarray}
Use (\ref{nw7})-(\ref{nw9}) in (\ref{nw6})
\begin{eqnarray}
\| X_{\hat{\beta}, -j} (\hat{\gamma}_{\hat{\beta},j} - \gamma_{\beta_0,j}) \|_n^2 & \le & 
\frac{C H_n^2 \lambda_0^2 s_0^2(1+t_1)}{2} + \frac{ \| X_{\beta_0,-j} (\hat{\gamma}_{\hat{\beta},j} - \gamma_{\beta_0,j}) \|_n^2}{2}
\nonumber \\
& + & (\frac{\eta_{\beta_0,j}' X_{\beta_0,-j}}{n})(\hat{\gamma}_{\hat{\beta},j} - \gamma_{\beta_0,j}) \nonumber \\
& + & \lambda_{nw}  \underline{\Omega} (\gamma_{\beta_0,j}) - \lambda_{nw} \underline{\Omega} (\hat{\gamma}_{\hat{\beta},j}).\label{nw10} 
\end{eqnarray}
Then under event $E_8$ since  $\| X_{\beta_0,-j} ( \hat{\gamma}_{\hat{\beta},j} - \gamma_{\beta_0,j}) \|_n^2 \le  \| X_{\hat{\beta},j} (\hat{\gamma}_{\hat{\beta},j}
- \gamma_{\beta_0,j}) \|_n^2$, by definition of squared weights, and $X_{\beta_0,-j}:= W_{\beta_0} W_{\hat{\beta}}^{-1}  X_{\hat{\beta},-j}$
\begin{eqnarray}
\frac{1}{2}  \| X_{\hat{\beta}, -j} ( \hat{\gamma}_{\hat{\beta},j} - \gamma_{\beta_0,j} )\|_n^2 &+& \lambda_{nw} \underline{\Omega} (\hat{\gamma}_{\hat{\beta},j})
\le (\frac{\eta_{\beta_0,j}' X_{\beta_0,-j}}{n})(\hat{\gamma}_{\hat{\beta},j} - \gamma_{\beta_0,j}) \nonumber \\
& + & \lambda_{nw} \underline{\Omega} (\gamma_{\beta_0,j}) + \frac{C H_n^2 \lambda^2 s_0^2 (1+t_1)}{2}.\label{nw11} 
\end{eqnarray}

{\bf Part 2.} This part extends the proof of Theorem 6.1 in van de Geer (2016) and fills in some blanks in that proof. First, we  extend that proof  to GLM from least-squares, and then  we allow for a different norm-decomposable one- $\underline{\Omega}$. The rates at the end of the proof are different for GLM than least-squares loss. In other words, there is a fundamental difference of our result compared with least squares loss.

We provide two conditions, and we show the bound under these two and then merge the conditions and bounds. 
We start with a trivial one.  

Part 2a. Assume that 
\begin{equation}
\frac{ \lambda_{nw} \underline{\Omega} (\hat{\gamma}_{\hat{\beta},j}  - \gamma_{\beta_0,j})}{8}
+ \| X_{\hat{\beta},-j} (\hat{\gamma}_{\hat{\beta},j}  - \gamma_{\beta_0,j}) \|_n^2 \le  C H_n^2 \lambda^2 s_0^2 (1+ t_1) + 4 \lambda_{nw} \Omega^{S^c} (\gamma_{\beta_0, S_j^c}).\label{p3c1} 
\end{equation}

This provides the following upper bound:
\begin{equation}
 \underline{\Omega} (\hat{\gamma}_{\hat{\beta},j}  - \gamma_{\beta_0,j})
\le  8  C H_n^2 \frac{\lambda^2}{\lambda_{nw}} s_0^2 (1+t_1)+ 32 \Omega^{S^c} (\gamma_{\beta_0, S_j^c}).\label{p3c3} \end{equation}

Part 2b. Now assume that we are under 
\begin{equation}
\frac{ \lambda_{nw}  \underline{\Omega} (\hat{\gamma}_{\hat{\beta},j}  - \gamma_{\beta_0,j})}{8}
+ \| X_{\hat{\beta},-j} (\hat{\gamma}_{\hat{\beta},j}  - \gamma_{\beta_0,j}) \|_n^2 \ge  C H_n^2 \lambda^2 s_0^2 (1+ t_1)  + 4 \lambda_{nw} \Omega^{S^c}  (\gamma_{\beta_0, S_j^c}).\label{p3c12} 
\end{equation}

Decompose $X_{\beta_0,-j}:= (X_{\beta_0, S_j}, X_{\beta_0, S_j^c})$ (where we do not put $-j$ index of these two terms, $X_{\beta_0, S_j}, X_{\beta_0, S_j^c}$ not to complicate the notation), similarly $\hat{\gamma}_{\hat{\beta},j}:= (\hat{\gamma}_{\hat{\beta}, S_j}, \hat{\gamma}_{\hat{\beta}, S_j^c})$, $\gamma_{\beta_0,j}:=
(\gamma_{\beta_0, S_j}, \gamma_{\beta_0, S_j^c})$.
Now we go back to (\ref{nw11}) and consider the first term on the right-side. Let $\Omega_*^{S^c}$ be the dual norm of $\Omega^{S^c}$.
\begin{eqnarray}
| (\hat{\gamma}_{\hat{\beta},j} - \gamma_{\beta_0,j})' X_{\beta_0,-j}' \eta_{\beta_0,j}/n| 
& = & | (\hat{\gamma}_{\hat{\beta}, S_j} - \gamma_{\beta_0, S_j})' X_{\beta_0, S_j}' \eta_{\beta_0,j} + 
(\hat{\gamma}_{\hat{\beta}, S_j^c} - \gamma_{\beta_0, S_j^c})' X_{\beta_0, S_j^c}' \eta_{\beta_0,j}|/n \nonumber \\
& \le & 
\Omega (\hat{\gamma}_{\hat{\beta},S_j} - \gamma_{\beta_0,S_j}) \Omega_{*}( X_{\beta_0,S_j}' \eta_{\beta_0,j}/n)
\nonumber \\
& + & 
\Omega^{S^c}  (\hat{\gamma}_{\hat{\beta},S_j} - \gamma_{\beta_0,S_j}) \Omega_{*}^{S^c} ( X_{\beta_0,S_j^c}' \eta_{\beta_0,j}/n) \nonumber \\
& \le & 
\Omega (\hat{\gamma}_{\hat{\beta},S_j} - \gamma_{\beta_0,S_j}) \| X_{\beta_0,S_j}' \eta_{\beta_0,j}/n\|_{\infty}  \nonumber \\
& + & 
\Omega^{S^c}  (\hat{\gamma}_{\hat{\beta},S_j^c} - \gamma_{\beta_0,S_j^c}) \| X_{\beta_0,S_j^c}' \eta_{\beta_0,j}/n\|_{\infty} \nonumber \\
& \le & 
\Omega (\hat{\gamma}_{\hat{\beta},S_j} - \gamma_{\beta_0,S_j})  (\lambda_{nw}/2) + \Omega^{S^c}  (\hat{\gamma}_{\hat{\beta},S_j^c} - \gamma_{\beta_0,S_j^c})
(\lambda_{nw}/2),\label{p31}
\end{eqnarray}
where we use Lemma \ref{la1}(i) and triangle inequality for the first inequality,  and then Lemma \ref{la1}(iii) for the second inequality, and the last inequality is by our events $E_6, E_7$. Now apply (\ref{p31}) in (\ref{nw11})
\begin{eqnarray}
\frac{ \| X_{\hat{\beta}, -j } (\hat{\gamma}_{\hat{\beta},j} - \gamma_{\beta_0,j}) \|_n^2}{2} & \le & 
\frac{\lambda_{nw}}{2} \Omega (\hat{\gamma}_{\hat{\beta}, S_j } - \gamma_{\beta_0, S_j}) + \frac{\lambda_{nw}}{2} \Omega^{S^c} ( \hat{\gamma}_{\hat{\beta},{S_j^c}}  - \gamma_{\beta_0, S_j^c}) 
\nonumber \\
& + & \lambda_{nw} \underline{\Omega} (\gamma_{\beta_0,j}) - \lambda_{nw} \underline{\Omega} (\hat{\gamma}_{\hat{\beta}, j}) + \frac{C H_n^2 \lambda^2 s_0^2 (1+ t_1) }{2}.\label{nw15}
\end{eqnarray}
Next, by a definition of $\underline{\Omega}(.)$ in (\ref{nw12a}), we can rewrite the third term on the right side of (\ref{nw15}) to get the first inequality below, and then apply triangle inequality for the second term on the right side of (\ref{nw15}) to get the second inequality  below
\begin{eqnarray}
\frac{ \| X_{\hat{\beta}, -j } (\hat{\gamma}_{\hat{\beta},j} - \gamma_{\beta_0,j}) \|_n^2}{2} & \le & 
\frac{\lambda_{nw}}{2}  \Omega (\hat{\gamma}_{\hat{\beta}, S_j } - \gamma_{\beta_0, S_j}) + \frac{\lambda_{nw}}{2} \Omega^{S^c} ( \hat{\gamma}_{\hat{\beta},{S_j^c}} 
- \gamma_{\beta_0, S_j^c}) \nonumber \\
&+& \lambda_{nw} \Omega (\gamma_{\beta_0, S_j}) + \lambda_{nw} \Omega^{S^c} (\gamma_{\beta_0, S_j^c}) \nonumber \\
& - & \lambda_{nw} \underline{\Omega} (\hat{\gamma}_{\hat{\beta},j}) + \frac{C H_n^2 \lambda^2 s_0^2 (1+ t_1) }{2} \nonumber \\
& \le & 
\frac{\lambda_{nw}}{2}  \Omega (\hat{\gamma}_{\hat{\beta}, S_j } - \gamma_{\beta_0, S_j}) + \frac{\lambda_{nw}}{2} \Omega^{S^c} ( \hat{\gamma}_{\hat{\beta},{S_j^c}} )
+ \frac{3 \lambda_{nw}}{2} \Omega^{S^c} (\gamma_{\beta_0, S_j^c})  \nonumber \\
& + & [ \lambda_{nw} \Omega (\gamma_{\beta_0, S_j}) - \lambda_{nw} \underline{\Omega} (\hat{\gamma}_{\hat{\beta},j}) ]+ \frac{C H_n^2 \lambda^2 s_0^2 (1 + t_1) }{2}.\label{nw16}
\end{eqnarray}

\noindent Start with the square bracketed terms on the right-side of (\ref{nw16}). Add and subtract $\lambda_{nw} \Omega^{S^c} (\gamma_{\beta_0, S_j^c})$
\begin{equation}
\lambda_{nw} \Omega (\gamma_{\beta_0, S_j}) - \lambda_{nw} \underline{\Omega} (\hat{\gamma}_{\hat{\beta}, j}) =
[ \lambda _{nw}\underline{\Omega} (\gamma_{\beta_0,j}) - \lambda_{nw} \underline{\Omega} (\hat{\gamma}_{\hat{\beta},j})] - \lambda_{nw} \Omega^{S^c} ( \gamma_{\beta_0, S_j^c}),\label{nw17} 
\end{equation}
where we also use the definition $\underline{\Omega} (\gamma_{\beta_0,j}) := \Omega (\gamma_{\beta_0, S_j}) + \Omega^{S^c} ( \gamma_{\beta_0, S_j^c})$. 
Now apply triangle property which is (\ref{nw12}) for the square bracketed term in (\ref{nw17}) above
\begin{equation}
\lambda_{nw} \underline{\Omega} (\gamma_{\beta_0, j}) - \lambda_{nw} \underline{\Omega} (\hat{\gamma}_{\hat{\beta},j}) 
\le \lambda_{nw} \Omega (\hat{\gamma}_{\hat{\beta}, S_j} - \gamma_{\beta_0, S_j}) + \lambda_{nw} \Omega^{S^c} (\gamma_{\beta_0, S_j^c}) 
- \lambda_{nw} \Omega^{S^c} ( \hat{\gamma}_{\hat{\beta}, S_j^c}).\label{nw18}
\end{equation}
Now apply (\ref{nw18}) to square bracketed term  on the right side of (\ref{nw17}) 
\begin{equation}
\lambda_{nw} \Omega (\gamma_{\beta_0, S_j}) - \lambda_{nw} \underline{\Omega} (\hat{\gamma}_{\hat{\beta}, j}) \le 
\lambda_{nw} \Omega (\hat{\gamma}_{\hat{\beta}, S_j} - \gamma_{\beta_0, S_j}) 
-\lambda_{nw} \Omega^{S^c} (\hat{\gamma}_{\hat{\beta}, S_j^c}).\label{nw19}
\end{equation}
Now use (\ref{nw19}) in the square bracketed term on the right side of (\ref{nw16}) and add and subtract $\lambda_{nw} \Omega^{S^c} (\gamma_{\beta_0, S_j^c})$ to get the equality below
\begin{eqnarray}
\frac{ \| X_{\hat{\beta}, -j } (\hat{\gamma}_{\hat{\beta},j} - \gamma_{\beta_0,j}) \|_n^2}{2} & \le & 
\frac{3\lambda_{nw}}{2}   \Omega (\hat{\gamma}_{\hat{\beta}, S_j} - \gamma_{\beta_0, S_j}) \nonumber \\
& + & \frac{\lambda_{nw}}{2} \Omega^{S^c} ( \hat{\gamma}_{\hat{\beta}, S_j^c}) + \frac{3\lambda_{nw}}{2} \Omega^{S^c} (\gamma_{\beta_0, S_j^c}) \nonumber \\
& - &  \lambda_{nw} \Omega^{S^c} (\hat{\gamma}_{\hat{\beta}, S_j^c}) 
+  \frac{C H_n^2 \lambda^2 s_0^2 (1+ t_1)}{2} \nonumber \\
& =  & \frac{3\lambda_{nw}}{2} \Omega (\hat{\gamma}_{\hat{\beta}, S_j} - \gamma_{\beta_0, S_j}) + \frac{3 \lambda_{nw}}{2}  \Omega^{S^c} (\gamma_{\beta_0, S_j^c}) \nonumber \\
& - & [\frac{\lambda_{nw}}{2}  \Omega^{S^c} (\hat{\gamma}_{\hat{\beta}, S_j^c}) + \frac{\lambda_{nw}}{2}  \Omega^{S^c} (\gamma_{\beta_0, S_j^c})] +\frac{\lambda_{nw}}{2} \Omega^{S^c}
(\gamma_{\beta_0, S_j^c}) \nonumber \\
& +& \frac{C H_n^2 \lambda^2 s_0^2 +(1+t_1)}{2}.\label{nw20}
\end{eqnarray}
Apply triangle inequality to the third  term (square bracketed term) on the right side of (\ref{nw20}) and simplifying

\begin{eqnarray}
\frac{ \| X_{\hat{\beta}, -j } (\hat{\gamma}_{\hat{\beta},j} - \gamma_{\beta_0,j}) \|_n^2}{2} & \le & 
\frac{3 \lambda_{nw}}{2}  \Omega (\hat{\gamma}_{\hat{\beta}, S_j} - \gamma_{\beta_0, S_j}) + 2  \lambda_{nw} \Omega^{S^c} (\gamma_{\beta_0, S_j^c}) \nonumber \\
& - & \frac{\lambda_{nw}}{2}  \Omega^{S^c} ( \hat{\gamma}_{\hat{\beta}, S_j^c} - \gamma_{\beta_0, S_j^c}) +
[\frac{C H_n^2 \lambda^2 s_0^2 (1+t_1) }{2}] 
.\label{nw21}
\end{eqnarray} 
This extends (6.9) of van de Geer (2016), up to constants,  but with added square bracketed term on the last right side term above due to GLM rather than least squares. Now multiply both sides of (\ref{nw21}) above by 2 and use on the left side of (\ref{p3c12}), and via simple cancellation of terms on the left and the right below in inequality (\ref{nw22})

\begin{equation}
 \frac{\lambda_{nw}\underline{\Omega} (\hat{\gamma}_{\hat{\beta}, j} - \gamma_{\beta_0,j})}{8} + 3 \lambda_{nw} 
\Omega (\hat{\gamma}_{\hat{\beta}, S_j} - \gamma_{\beta_0, S_j}) -  \lambda_{nw} \Omega^{S^c} (\hat{\gamma}_{\hat{\beta}, S_j^c}
- \gamma_{\beta_0, S_j^c}) \ge 0.\label{nw22}
\end{equation}
Use the definition of the norm 
\begin{equation}
\underline{\Omega} (\hat{\gamma}_{\hat{\beta}, j} - \gamma_{\beta_0,j}) := \Omega (\hat{\gamma}_{\hat{\beta}, S_j} - \gamma_{\beta_0,S_j})
+ \Omega^{S^c} (\hat{\gamma}_{\hat{\beta}, S_j^c} - \gamma_{\beta_0,S_j^c} ),\label{b38a}
\end{equation}
which implies by (\ref{nw22})
\begin{equation}
\frac{25 \lambda_{nw}}{8} \Omega (\hat{\gamma}_{\hat{\beta}, S_j} - \gamma_{\beta_0,S_j})
\ge \frac{7 \lambda_{nw}}{8}\Omega^{S_j^c} (\hat{\gamma}_{\hat{\beta}, S_j^c} - \gamma_{\beta_0,S_j^c} ).\label{nw23}
\end{equation}
This clearly shows that cone condition, (\ref{cc1}),  is satisfied in compatibility condition, with 
$ L = 25/7.$
Next consider 
(\ref{nw21}) first by adding $\frac{\lambda_{nw}}{2} \Omega (\hat{\gamma}_{\hat{\beta}, S_j} - \gamma_{\beta_0, S_j})$ to both sides for the first inequality below and use the compatibility condition 
in (\ref{nw13}), 
and the inequality $ab \le a^2/4+b^2$, with $a:= \| X_{\hat{\beta}, -j}  (\hat{\gamma}_{\hat{\beta},j} - \gamma_{\beta_0,j}) \|_n$, $b:= 2 \lambda_{nw} \sqrt{ | S_j |}/\hat{\phi} (L, S_j)$
for the second and third inequalities below
\begin{eqnarray}
\| X_{\hat{\beta}, -j} (\hat{\gamma}_{\hat{\beta},j }- \gamma_{\beta_0,  j}) \|_n^2/2 &+& 
\frac{\lambda_{nw}}{2} \Omega^{S^c} (\hat{\gamma}_{\hat{\beta}_{S_j^c}} - \gamma_{\beta_0, S_j^c} )
 +  \frac{\lambda_{nw}}{2} \Omega (\hat{\gamma}_{\hat{\beta}, S_j} - \gamma_{\beta_0, S_j}) \nonumber \\
& \le & 2 \lambda_{nw}  \Omega (\hat{\gamma}_{\hat{\beta}, S_j} - \gamma_{\beta_0, S_j}) 
\nonumber \\
& + & 2 \lambda_{nw}  \Omega^{S^c}  (\gamma_{\beta_0, S_j^c}) + \frac{C H_n^2 \lambda^2 s_0^2 (1 + t_1)  }{2} \nonumber \\
& \le &   2 \lambda_{nw} \sqrt{ | S_j |}  \frac{ \| X_{\hat{\beta}, -j} (\hat{\gamma}_{\hat{\beta},j} - \gamma_{\beta_0,  j})\|_n }{\hat{\phi} (L,S_j)} \nonumber \\
& + & 2 \lambda_{nw}  \Omega^{S^c} (\gamma_{\beta_0, S_j^c}) + \frac{C H_n^2 \lambda^2 s_0^2 (1 + t_1)  }{2} \nonumber \\
& \le & \frac{   4 \lambda_{nw}^2  |S_j | }{\hat{\phi}^2 ( L, S_j)}
+ \frac{\| X_{\hat{\beta}, -j} (\hat{\gamma}_{\hat{\beta},j} - \gamma_{\beta_0,  j}) \|_n^2 }{4}  \nonumber \\
& + & 2 \lambda_{nw}  \Omega^{S^c} (\gamma_{\beta_0, S_j^c}) + \frac{C H_n^2 \lambda^2 s_0^2 (1 + t_1) }{2}.\label{nw24}
\end{eqnarray}

Now form the norm bound by observing that on the left side of (\ref{nw24}) the second and third items add up to $\frac{\lambda}{2} \underline{\Omega} (\hat{\gamma}_{\hat{\beta}, j} - \gamma_{\beta_0,j})$ by (\ref{b38a}), and multiply each side by $2/\lambda_{nw}$, and also the first term on the left side of (\ref{nw24}) is larger than the second  term on the right side of (\ref{nw24}), so bound is preserved without them,

\begin{equation}
\underline{\Omega} (\hat{\gamma}_{\hat{\beta}, j} - \gamma_{\beta_0, j}) \le 
\frac{  8 \lambda_{nw} |S_j | }{\hat{\phi}_{\Omega}^2 ( L, S_j)} + 
4  \Omega^{S^c} (\gamma_{\beta_0, S_j^c}) + C H_n^2 \frac{\lambda^2}{\lambda_{nw}} s_0^2 (1+t_1) .\label{nw25b}
\end{equation}
(\ref{nw25b}) provides the upper bound for the condition in part 2b here, (\ref{p3c12}). Now combine the upper bounds in parts 2a and part 2b, by taking max of those in  (\ref{p3c3})(\ref{nw25b}) we have 
\begin{equation}
\underline{\Omega} (\hat{\gamma}_{\hat{\beta}, j} - \gamma_{\beta_0, j}) \le 
\frac{   8 \lambda_{nw}  |S_j | }{\hat{\phi}_{\Omega}^2 ( L, S_j)} + 
 32 \Omega^{S^c} (\gamma_{\beta_0, S_j^c})  + 8C H_n^2 \frac{\lambda^2}{\lambda_{nw}} s_0^2 (1+t_1).\label{nw25c}
\end{equation}

Next add Lemmata \ref{la5}-\ref{la7a}, and (\ref{pla68}) to have the desired result in (i).

(ii). Given the condition $\sup_{\beta \in {\cal B}_{l_o} (s_0)} \max_{ j \in {\cal H}} \Omega (\gamma_{\beta_0, S_j^c}) = o(d_n)=o(1)$ and adding Assumption \ref{as6}(i) we have the desired result. Note that Assumption \ref{as6}(i)-(ii) are stronger than Assumption \ref{as2}(ii). See that by Lemma \ref{la7}(ii) we have $\lambda_{nw}= O_p (\sqrt{\frac{lnp}{n}})$.
Also, we see that the asymptotic result is uniform over the $l_0$ ball ${\cal B}_{l_0} (s_0)$, since the asymptotic upper bound depends on $\beta_0$ through $s_0$. This condition is partial sparsity since we impose this only on the $h$ rows that we are interested in, not all rows of the precision matrix.  
{\bf Q.E.D.}

\begin{lemma}\label{la9}
Under Assumptions \ref{as1}-\ref{as2}(i), \ref{as3}-\ref{as5}, \ref{as6}(i)-(ii), and $\sup_{\beta_0 \in {\cal B}_{l_0} (s_0)}
\max_{ j \in {\cal H}} \Omega^{S^c} (\gamma_{\beta_0, S_j^c})=o(d_n)=o(1)$,  we have 
\[ \max_{j \in {\cal H}} |\hat{\tau}_j^2 - \tau_j^2 | = O_p (\sqrt{\bar{s}} \sqrt{\frac{lnp}{n}} \max( \bar{s}, H_n^2 s_0^2) ) = o_p (1).\]
and $min_{1 \le j \le p} \tau_j^2 \ge c > 0$, and $\max_{1 \le j \le p} \tau_j^2 \le C < \infty$, for universal positive constants, $c, C$. This result is uniform over $l_0$ ball ${\cal B}_{l_0} (s_0)$.

\end{lemma}

{\bf Proof of Lemma \ref{la9}}. Start with definitions, for each $j=1,\cdots,p$,
\begin{equation}
\hat{\tau}_j^2:= X_{\hat{\beta},j}' (X_{\hat{\beta},j} - X_{\hat{\beta}, -j} \hat{\gamma}_{\hat{\beta},j})/n.\label{ta1}
\end{equation}
and its population version, for $i=1,\cdots,n$, $\eta_{\beta_0, j,i}$ represents the $i$ th element of $n \times 1$ vector $\eta_{\beta_0,j}$.
\begin{equation}
\tau_j^2:= E \eta_{\beta_0, j,i}^2.\label{ta2}
\end{equation}
We also use the transformation
\begin{equation}
X_{\hat{\beta}, -j} = W_{\hat{\beta}} W_{\beta_0}^{-1} X_{\beta_0,-j},\label{ta3}
\end{equation}
with similar equation holding for $X_{\hat{\beta},j}$, and with (\ref{nw1})
\begin{equation}
X_{\hat{\beta}, j} = W_{\hat{\beta}} W_{\beta_0}^{-1} X_{\beta_0,j} =
W_{\hat{\beta}} W_{\beta_0}^{-1} (X_{\beta_0,-j} \gamma_{\beta_0,j} + \eta_{\beta_0,j}).\label{ta4}
\end{equation}
Use definition of $\hat{\tau}_j^2$
 \begin{eqnarray}
\hat{\tau}_j^2 - \tau_j^2 & = & X_{\beta_0,j}' ( X_{\beta_0, j} - X_{\beta_0,-j} \hat{\gamma}_{\hat{\beta},j})/n
- \tau_j^2 \nonumber \\
& + & X_{\beta_0,j}' (W_{\hat{\beta}}^2 W_{\beta_0}^{-2} - I_n) ( X_{\beta_0, j} - X_{\beta_0,-j} \hat{\gamma}_{\hat{\beta},j})/n.\label{ta5}
\end{eqnarray}
To get the right side of (\ref{ta5}) above, we impose (\ref{ta3})-(\ref{ta4})(first equality) in $X_{\hat{\beta},j}$ in (\ref{ta1}) then add and subtract 
$X_{\beta_0,j}' (X_{\beta_0,j} - X_{\beta_0,-j} \hat{\gamma}_{\hat{\beta},j})/n$.

Analyze the first term above by (\ref{nw1})(\ref{ta4})
\begin{eqnarray}
X_{\beta_{0,j}}' ( X_{\beta_{0, j}} - X_{\beta_{0,-j}} \hat{\gamma}_{\hat{\beta},j})/n
- \tau_j^2 & = & X_{\beta_{0,j}}' ( \eta_{\beta_{0, j}} - X_{\beta_{0,-j}}' (\hat{\gamma}_{\hat{\beta},j} - \gamma_{\beta_0,j}))/n - \tau_j^2 \nonumber \\
& = & X_{\beta_{0j}}' \eta_{\beta_{0,j}}/n - (X_{\beta_{0,j}}' X_{\beta_{0,-j}}/n) (\hat{\gamma}_{\hat{\beta},j} - \gamma_{\beta_0,j}) - \tau_j^2 \nonumber \\
& = & \frac{\gamma_{\beta_{0,j}} X_{\beta_{0,-j}}' \eta_{\beta_{0,j}}}{n} + \left[ \frac{\eta_{\beta_{0,j}}' \eta_{\beta_{0,j}}}{n} - \tau_j^2
\right] \nonumber \\
& - & \frac{ X_{\beta_{0,j}}' X_{\beta_{0,-j}}}{n} (\hat{\gamma}_{\hat{\beta},j} - \gamma_{\beta_0,j})
.\label{ta6}
\end{eqnarray}
By Assumptions \ref{as1},\ref{as5}
we have via (\ref{sa2})(\ref{sa4})(\ref{sa5}) 
\begin{equation}
\max_{j \in {\cal H}} \left[ \frac{\eta_{\beta_{0,j}}' \eta_{\beta_{0,j}}}{n} - \tau_j^2
\right] =O_p ( \sqrt{\frac{lnh}{n}}).\label{ta7}
\end{equation}
Then by p.159 and (B.48) of Caner and Kock (2018) via Assumption \ref{as5},  $\max_{j \in {\cal H}} \| \gamma_{\beta_0,j} \|_1 = O (\bar{s}^{1/2})$ ,  and Lemma \ref{la7} here
\begin{equation}
\max_{j \in {\cal H}} |\frac{\gamma_{\beta_{0,j}} X_{\beta_{0,-j}}' \eta_{\beta_{0,j}}}{n} | \le \max_{j \in {\cal H}}
\| \gamma_{\beta_{0,j}} \|_1 \max_{j \in {\cal H}}
\| \frac{X_{\beta_{0,-j}}' \eta_{\beta_{0,j}}}{n} \|_{\infty}
= O(\sqrt{\bar{s}}) O_p ( \sqrt{\frac{lnp}{n}}).\label{ta8}
\end{equation}

\noindent Next, using the expression of $X_{\beta_0,j}$ in (\ref{nw1})  in the third term on the right side of (\ref{ta6}) 
\begin{eqnarray}
\frac{ X_{\beta_{0,j}}' X_{\beta_{0,-j}}}{n} (\hat{\gamma}_{\hat{\beta},j} - \gamma_{\beta_0,j})
& \le & \max_{j \in {\cal H}}
| \gamma_{\beta_{0,j}}' \frac{X_{\beta_{0,-j}}' X_{\beta_{0,-j}}}{n} (\hat{\gamma}_{\hat{\beta},j} - \gamma_{\beta_{0,j}})| \nonumber 
\\
& + & \max_{j \in {\cal H}}
| \frac{\eta_{\beta_{0.j}}' X_{\beta_{0,-j}}'}{n}  (\hat{\gamma}_{\hat{\beta},j} - \gamma_{\beta_{0,j}})|.\label{ta9}
\end{eqnarray}
In (\ref{ta9}) above analyze
\begin{eqnarray}
\max_{j \in {\cal H}} | \gamma_{\beta_{0,j}}' \frac{X_{\beta_{0,-j}}' X_{\beta_{0,-j}}}{n} (\hat{\gamma}_{\hat{\beta},j} - \gamma_{\beta_{0,j}})| 
& \le & \max_{j \in {\cal H}}
\| \gamma_{\beta_{0,j}} \|_1  
\max_{j \in {\cal H}} \| \frac{X_{\beta_{0,-j}}' X_{\beta_{0,-j}}}{n} (\hat{\gamma}_{\hat{\beta},j} - \gamma_{\beta_{0,j}})\|_{\infty}
\nonumber \\
& \le  & \max_{j \in {\cal H}}
\| \gamma_{\beta_{0,j}} \|_1 \max_{j \in {\cal H}}
\| \frac{X_{\beta_{0,-j}}' X_{\beta_{0,-j}}}{n} \|_{\infty}
\max_{j \in {\cal H}}
\|(\hat{\gamma}_{\hat{\beta},j} - \gamma_{\beta_{0,j}})\|_1 \nonumber \\
& \le & \max_{j \in {\cal H}} \| \gamma_{\beta_{0,j}} \|_1 \max_{j \in {\cal H}}
\left[\| \frac{X_{\beta_{0,-j}}' X_{\beta_{0,-j}}}{n} \|_{\infty} \right]
\max_{j \in {\cal H}}
\underline{\Omega} ((\hat{\gamma}_{\hat{\beta},j} - \gamma_{\beta_{0,j}})) \nonumber \\
& = & O (\sqrt{\bar{s}}) O_p (1) O_p ( \sqrt{\frac{lnp}{n}} max (\bar{s}, H_n^2 s_0^2)),\label{ta10}
\end{eqnarray}
by Holders inequality for the first inequality above, then again the inequality Lemma \ref{la1}(iv) for the second inequality,  and Lemma \ref{la1}(ii) for the third inequality, and the rates are from (B.48) of Caner and Kock (2018), and by uniformly bounded weights and Assumptions \ref{as1},\ref{as2} to use (\ref{sa2})(\ref{sa4})(\ref{sa5}) and Lemma \ref{la8}(ii). In (\ref{ta9}) we consider 
\begin{eqnarray}
\max_{j \in {\cal H}}
| \frac{\eta_{\beta_{0.j}}' X_{\beta_{0,-j}}'}{n}  (\hat{\gamma}_{\hat{\beta},j} - \gamma_{\beta_{0,j}})| & \le & 
\max_{j \in {\cal H}}
\underline{\Omega}^* (\frac{\eta_{\beta_{0.j}}' X_{\beta_{0,-j}}'}{n} ) 
\max_{j \in {\cal H}}
\underline{\Omega} (\hat{\gamma}_{\hat{\beta},j} - \gamma_{\beta_{0,j}})
\nonumber \\
& \le & \max_{j \in {\cal H}}
\| (\frac{\eta_{\beta_{0.j}}' X_{\beta_{0,-j}}'}{n} \|_{\infty} 
\max_{j \in {\cal H}}
\underline{\Omega} (\hat{\gamma}_{\hat{\beta},j} - \gamma_{\beta_{0,j}}) \nonumber \\
& = & O_p (\sqrt{\frac{lnp}{n}}) O_p ( \sqrt{\frac{lnp}{n}} max (\bar{s}, H_n^2 s_0^2)),\label{ta11}
\end{eqnarray}
where we use dual norm inequality in Lemma \ref{la1}(i) for the first inequality, and then Lemma \ref{la1} (iii) for the second inequality and the rates are 
by Lemma \ref{la7} and Lemma \ref{la8}. Combine (\ref{ta7})-(\ref{ta11}) in (\ref{ta6}) and by the slowest rate in (\ref{ta10}), and Assumption \ref{as6}(i)
\begin{equation}
X_{\beta_{0,j}}' ( X_{\beta_{0, j}} - X_{\beta_{0,-j}} \hat{\gamma}_{\hat{\beta},j})/n
- \tau_j^2  = O_p (\sqrt{\bar{s}} \sqrt{\frac{lnp}{n}} max (\bar{s}, H_n^2 s_0^2)).\label{ta12}
\end{equation}

We now consider the second term on the right side of (\ref{ta5}). To do that we need the following results, first for  vectors of $n$ dimension, $x,y$, and a diagonal matrix of $A$
\begin{equation}
|x' A y| \le \| x\|_{\infty} \| A y \|_1 = \| x \|_{\infty} \sum_{i=1}^n | A_{ii} y_i | \le  \| x \|_{\infty} \left[ \sum_{i=1}^n |A_{ii}|\right] \| y \|_{\infty}.\label{ineq1}
\end{equation}
Then see that  by (\ref{nw1})
\[ X_{\beta_0,j} - X_{\beta_0,-j} \hat{\gamma}_{\hat{\beta},j} = \eta_{\beta_0,j} - X_{\beta_0,j} ( \hat{\gamma}_{\hat{\beta},j} - \gamma_{\beta_0,j}).\]
Next
\begin{eqnarray}
\max_{j \in {\cal H}} 
\| \eta_{\beta_0,j} - X_{\beta_0,j}' (\hat{\gamma}_{\hat{\beta},j} - \gamma_{\beta_0,j} ) \|_{\infty}  & \le &  
\max_{j \in {\cal H}} \| \eta_{\beta_0,j} \|_{\infty} + \max_{j \in {\cal H}}
\| X_{\beta_0,-j}' (\hat{\gamma}_{\hat{\beta},j} - \gamma_{\beta_0,j}) \|_{\infty}  \nonumber \\
& \le & \max_{j \in {\cal H}}
\| \eta_{\beta_0,j} \|_{\infty} + \max_{j \in {\cal H}}
 \underline{\Omega}_* ( X_{\beta_0,-j} )  \max_{j \in {\cal H}}
 \underline{\Omega} (\hat{\gamma}_{\hat{\beta},j} - \gamma_{\beta_0,j}) \nonumber \\
& \le & \max_{j \in {\cal H}}
\| \eta_{\beta_0,j} \|_{\infty} +  \max_{j \in {\cal H}}
\| X_{\beta_0,-j}  \|_{\infty} 
\max_{j \in {\cal H}} \underline{\Omega} (\hat{\gamma}_{\hat{\beta},j} - \gamma_{\beta_0,j}) \nonumber \\
& = & O_p (H_n) + O_p (H_n) o_p (1) = O_p (H_n),\label{ta13}
\end{eqnarray}
where we use triangle inequality for the first inequality, and dual norm inequality in Lemma \ref{la1} (i) for the second inequality, and Lemma \ref{la1}(iii) for the last inequality, and the rates are obtained by (\ref{eta}), and Markov's inequality-Lemma A.3 Caner and Kock (2018), with Assumptions \ref{as1}, \ref{as4} with weights bounded away from zero and infinity, Lemma \ref{la8}. Now we analyze the second term 
on the right side of (\ref{ta5}).
{\small
\begin{eqnarray}
\frac{X_{\beta_0,j}' (W_{\hat{\beta}}^2 W_{\beta_0}^{-2} - I_n) ( X_{\beta_0, j} - X_{\beta_0,-j} \hat{\gamma}_{\hat{\beta},j})}{n} & \le & 
\max_{j \in {\cal H}} \| X_{\beta_0,j} \|_{\infty}  \left[  \frac{1}{n} \sum_{i=1}^n \frac{|w_{\hat{\beta},i}^2 - w_{\beta_0,i}^2|}{w_{\beta_0,i}^2}
\right ] \nonumber \\
& \times & \left[ \max_{ j \in {\cal H}}  \| \eta_{\beta_0,j} \|_{\infty} +  \max_{1 \le j \le p} \| X_{\beta_0,-j}  \|_{\infty}  
\max_{j \in {\cal H}}
\underline{\Omega} (\hat{\gamma}_{\hat{\beta},j} - \gamma_{\beta_0,j}) \right] \nonumber \\
& \le  & max_{ j \in {\cal H}} \| X_{\beta_0,j} \|_{\infty}  \frac{C_p^2 \| X (\hat{\beta} - \beta_0) \|_n}{n^{1/2}} \nonumber \\
& \times & \left[ \max_{ j \in {\cal H}}  \| \eta_{\beta_0,j} \|_{\infty} +  \max_{ j \in {\cal H}} \| X_{\beta_0,-j}  \|_{\infty}  
\max_{j \in {\cal H}}
\underline{\Omega} (\hat{\gamma}_{\hat{\beta},j} - \gamma_{\beta_0,j}) \right] 
\nonumber \\
& = & O_p ( H_n) O_p ( s_0 \sqrt{\frac{lnp}{n}} \frac{1}{\sqrt{n}}) O_p (H_n),\label{ta14}
\end{eqnarray}
}
\noindent where we use (\ref{ineq1}) for the first inequality and (\ref{ta13}) for the second inequality 
with (\ref{weight}), and by Assumption \ref{as4}(iii)
\[ | w_{\hat{\beta},i}^2 - w_{\beta_0,i}^2 | \le | X_i' (\hat{\beta} - \beta_0) |.\]
Then by Assumption \ref{as3}, and (\ref{weight}) at $\beta_0$
\[ \frac{1}{w_{\beta_0,i}^2} = \frac{1}{\ddot{\rho} (y_i, X_i' \beta_0)} \le C_p^2.\]
So 
\[ \frac{1}{n} \sum_{i=1}^n \frac{|w_{\hat{\beta},i}^2 - w_{\beta_0, i}^2|}{w_{\beta_0,i}^2} \le 
\frac{C_p^2}{n^{1/2}} \| X (\hat{\beta}- \beta_0)\|_n.\]

 Next the rates in (\ref{ta14}) are by Assumption \ref{as1} with Markov's inequality, Lemma A.3 of Caner and Kock (2018), and we use (\ref{oi})
and (\ref{ta13}). Comparing the rates in (\ref{ta12}) with (\ref{ta14}) clearly slowest is (\ref{ta12}), and also we have $\tau_j^2 \le C < \infty$ by $\tau_j^2$ definition and Assumption \ref{as5}. Also the minimum eigenvalue condition in Assumption \ref{as5} with $\tau_j^2$ definition provides that $\min_{1 \le j \le p} \tau_j^2 \ge c > 0$. To see this last point, by p.157 of Caner and Kock (2018) we can have, for all $j =1, \cdots, p$  
\[ \tau_j^2= \frac{1}{\Sigma_{\beta_0,j,j}} \ge \inf_{\beta_0 \in {\cal B}_{l_0} (s_0)} 1/Eigmax(\Sigma_{\beta_0})  = \inf_{\beta_0 \in {\cal B}_{l_0} (s_0)} Eigmin(\Sigma_{\beta_0}) \ge c > 0.\]
Uniformity follows through since the bounds depend on $\beta_0$ through $s_0$.

{\bf Q.E.D.}

{\bf Proof of Lemma \ref{la10}}.
Note that $\hat{C}_j$ is a $p \times 1$ vector with $1$ in $j$ th cell, and $-\hat{\gamma}_{\hat{\beta},j}$  as the remaining $p-1$ part. For the population quantities $C_j$ ($p \times 1$ vector) is defined as $1$ in $j$ th cell, and the rest of the vector is $-\gamma_{\beta_0,j}$.
Now 
\begin{eqnarray}
\max_{ j \in {\cal H}} \underline{\Omega} (\hat{\Theta}_j - \Theta_j) & \le & \max_{ j \in {\cal H} } \underline{\Omega} ( \frac{1}{\hat{\tau}_j^2 } - \frac{1}{\tau_j^2}) + 
\max_{ j \in {\cal H}  } \underline{\Omega} ( \frac{\hat{\gamma}_{\hat{\beta},j}}{\hat{\tau}_j^2 } - \frac{\gamma_{\beta_0,j}}{\tau_j^2}) \nonumber \\
& \le & \max_{j \in {\cal H} } \left|  \frac{1}{\hat{\tau}_j^2 } - \frac{1}{\tau_j^2} \right| + \max_{ j \in {\cal H} } \underline{\Omega} (\frac{\hat{\gamma}_{\hat{\beta},j} - \gamma_{\beta_0,j}}{\hat{\tau}_j^2}) \nonumber \\
&+& \max_{ j \in {\cal H} } \underline{\Omega} ( \frac{\gamma_{\beta_0,j}}{\hat{\tau}_j^2} - \frac{\gamma_{\beta_0,j}}{\tau_j^2}) \nonumber \\
& \le &  \max_{ j \in {\cal H}  } \left|  \frac{1}{\hat{\tau}_j^2 } - \frac{1}{\tau_j^2} \right| + \max_{ j \in {\cal H} } | \frac{1}{\hat{\tau}_j^2}| \max_{j \in {\cal H} } 
\underline{\Omega} ( \hat{\gamma}_{\hat{\beta},j} - \gamma_{\beta_0,j}) \nonumber \\
& + & \max_{ j \in {\cal H} } | \frac{1}{\hat{\tau}_j^2} - \frac{1}{\tau_j^2} | \max_{ j \in {\cal H} } \underline{\Omega} (\gamma_{\beta_0,j}).\label{ta15}
 \end{eqnarray}
 Consider the following in (\ref{ta15}) 
 \begin{eqnarray}
 \max_{ j \in {\cal H} } \left| \frac{1}{\hat{\tau}_j^2} - \frac{1}{\tau_j^2} 
 \right| & = & \max_{ j \in {\cal H} } \left|  \frac{\hat{\tau}_j^2 - \tau_j^2}{\hat{\tau}_j^2 \tau_j^2}
 \right| \nonumber \\
 & = & O_p (\sqrt{\bar{s}} \max(\bar{s}, H_n^2 s_0^2) \sqrt{\frac{lnp}{n}}),\label{ta16}
 \end{eqnarray}
where we use Lemma \ref{la9}. Then in (\ref{ta15}) consider 
\begin{equation}
\max_{ j \in {\cal H} } | \frac{1}{\hat{\tau}_j^2}| \max_{1 \le j \le p} 
\underline{\Omega} ( \hat{\gamma}_{\hat{\beta},j} - \gamma_{\beta_0,j})  = O_p (1) O_p (\max(\bar{s}, H_n^2 s_0^2) \sqrt{\frac{lnp}{n}}),\label{ta17}
\end{equation}
by Lemma \ref{la8} and Lemma \ref{la9} Last, in (\ref{ta15}) 
\begin{equation}
 \max_{ j \in {\cal H}} | \frac{1}{\hat{\tau}_j^2} - \frac{1}{\tau_j^2} | \max_{ j \in {\cal H} } \underline{\Omega} (\gamma_{\beta_0,j})
= O_p ( \sqrt{\bar{s}} \max(\bar{s}, H_n^2 s_0^2) \sqrt{\frac{lnp}{n}}) O (g_n),\label{ta18}
\end{equation}
where we use Lemma \ref{la9} above and Assumption \ref{as6}(iii),  $ \max_{ j \in {\cal H}} \underline{\Omega} (\gamma_{\beta_0,j}) = O ( g_n)$. Function $g_n$ is known and depends on the norm. For the case of $\underline{\Omega} = l_1(,)$, we have $g_n = \bar{s}^{1/2}$. This last rate is the slowest rate in (\ref{ta15}) terms.
Note that also the result is uniform over $l_0$ ball ${\cal B}_{l_0} (s_0)$ since the upper bound depends on $\beta_0$ only through $s_0$.
{\bf Q.E.D.}

We use a lemma in van de Geer et al. (2014), a simple application of Holder's inequality.  Let $\hat{v} \in R^p, v \in R^p$ and 
$A$ is symmetric $p \times p$ matrix.

\begin{lemma}\label{la11}
\[| \hat{v}' A \hat{v} - v' A v | \le[  \| (\hat{v} - v ) \|_1]^2  \| A \|_{\infty}  + 2 \| A v \|_{\infty}  \| (\hat{v} - v)\|_1 .\]
\end{lemma}


Next lemma will be helpful in central limit theorem type results.

\begin{lemma}\label{la12}
\[ \| \hat{\Theta}' \alpha - \Theta' \alpha \|_1 \le \left[ \max_{j \in {\cal H}} \underline{\Omega} (\hat{\Theta}_j - \Theta_j)\right] \left[ \sum_{j \in {\cal H}} |\alpha_j | \right].\]

\end{lemma}

{\bf Proof of Lemma \ref{la12}}. 
By using $\alpha$ definition
\begin{eqnarray*}
\| (\hat{\Theta}' - \Theta') \alpha \|_1 & = & \| \sum_{j \in {\cal H}} (\hat{\Theta}_j - \Theta_j) \alpha_j \|_1 \\
& \le & \sum_{j \in {\cal H}} \| \hat{\Theta}_j - \Theta_j \|_1 |\alpha_j | \\
& \le & [ \max_{j \in{\cal H}} \| \hat{\Theta}_j - \Theta_j \|_1 ] [ \sum_{j \in {\cal H}} | \alpha_j |] \\
& \le & [ \max_{ j \in {\cal H}} \underline{\Omega} (\hat{\Theta}_j - \Theta_j )] \sum_{j \in {\cal H}} | \alpha_j |,
\end{eqnarray*}
last inequality is by Lemma \ref{la1}(ii).{\bf Q.E.D.}\\

{\bf \large References}:\\

Belloni, A., Chernozhukov, V., Hansen, C. 2014. Inference on treatment effects after selection amongst high dimensional controls. Review of Economic Studies, 81, 608-650.

Belloni, A., Chernozhukov, V., Wei, Y. 2017. Post-selection inference for generalized linear models with many controls. Journal of Business and Economic Statistics, 34, 606-619.

Belloni, A., Chernozhukov, V., Hansen, C, Newey, W. 2017. Simultaneous confidence intervals for high dimensional linear models with many endogenous variables. arXiv:1712.08102.

Caner, M., Kock, A.B. 2018. Asymptotically honest confidence regions for high dimensional parameters by the desparsified conservative lasso. Journal of Econometrics, 203, 143-168.

Caner, M. Kock, A.B. 2019. High dimensional linear GMM. arXiv. 1811.08779.v2.

Chernozhukov, V. Chetverikov, D. Kato, K. 2017. Central limit theorems and bootstrap in high dimensions. Annals of Probability, 45, 2309-2352.

Chernozhukov, V., Goldman, M., Semenova, V., Taddy, M. 2018. Orthogonal machine learning for demand estimation: High dimensional causal inference  in dynamic panels. arXiv:1712.099988.

Chiang, H., Sasaki, Y. 2019. Causal inference by quantile regression kink design. Journal of Econometrics, 210, 405-433.

Jankova, J., van de Geer, S. 2016. Confidence regions for high-dimensional generalized linear models under sparsity. arXiv:1610.01353v1.

Kock, A.B. 2016. Oracle Inequalities, Variable Selection and Uniform Inference in High-Dimensional Correlated Random Effects Panel Data Models. Journal of Econometrics, 195, 71-85.

Kock, A.B., Tang, H. 2019. Inference in high dimensional dynamic panels. Econometric Theory, 35, 295-359.

Lounici, K., Pontil, M. van de Geer, S., Tsybakov, A. 2011. Oracle inequalities and optimal inference under group sparsity.
Annals of Statistics, 39, 2164-2204.

Meier, L., van de Geer, S., Buhlmann, P. 2008. The group lasso for logistic regression. Journal of  Royal Statistical Society Series B, 70, 53-71.

Meier, L. 2020. Fitting User-Specified Models with Group Lasso Penalty. "grplasso-R".
website: https://cran.r-project.org/web/packages/grplasso/grplasso.pdf

Mitra, R., Zhang, C.H. 2016. The benefit of group sparsity in group inference with de-biased scaled group lasso. Electronic Journal of  Statistics, 10, 1829-1873.

Ning, Y., Liu, H. 2017. A general theory of hypotheses tests and confidence regions for sparse high dimensional models. Annals of Statistics, 45, 158-195.

Shi, C., Song, R., Chen, Z., Li, R. 2019. Linear hypotheses tests for high dimensional generalized linear models. Annals of Statistics, 47, 2671- 2703.

Stucky, B., van de Geer, S. 2018. Asymptotic confidence regions for high-dimensional structured sparsity. IEEE Transactions on Signal Processing, 66, 2178-2189.

van de Geer, S. 2014. Weakly decomposable regularization penalties and structured sparsity. Scandinavian Journal of Statistics, 41, 72-86. 

van de Geer, S. 2016. Estimation and testing under sparsity. Springer Verlag, Berlin.

van de Geer, S. Buhlmann, P., Ritov, Y., Dezeure, R. 2014. On asymptotically optimal confidence regions and tests for high dimensional models. Annals of Statistics, 42, 1166-1202.

Wainwright. M.J. 2019. High-Dimensional Statistics: A non-asymptotic viewpoint. Cambridge University Press.

Xia, L., Nan, B., Li, Y. 2020. A revisit to debiased lasso for GLM. arXiv:2006.12778.

Yuan, M., Lin, Y. 2006. Model selection and estimation in regression with grouped variables. Journal of  Royal Statistical Society Series B, 68,49-67.

\end{document}